\documentclass[journal]{IEEEtran}

\usepackage{amsmath,amssymb}
\usepackage{colortbl}
\usepackage{diagbox}
\usepackage{subeqnarray}
\usepackage{cases}
\usepackage{setspace}
\usepackage{tabularx}
\usepackage{booktabs}
\usepackage{multicol}
\usepackage{dcolumn}
\usepackage{graphicx}
\usepackage{subfigure}
\usepackage{float}
\usepackage{verbatim}
\usepackage{epsfig}
\usepackage{amsmath}
\usepackage{amssymb}
\usepackage{array}
\usepackage{epstopdf}
\usepackage{color}
\usepackage{xcolor}
\usepackage{cite}
\usepackage{psfrag}
\usepackage{cite}
\usepackage{multirow}
\usepackage{color}
\usepackage{bm} 
\usepackage{adjustbox}
\usepackage{overpic}
\usepackage{rotating}
\usepackage{multicol}

\graphicspath{{figures/}}
\usepackage[bookmarks,colorlinks,linkcolor=red, anchorcolor=blue, citecolor=green]{hyperref}
\makeatletter
\newcommand*\bigcdot{\mathpalette\bigcdot@{.5}}
\newcommand*\bigcdot@[2]{\mathbin{\vcenter{\hbox{\scalebox{#2}{$\m@th#1\bullet$}}}}}
\makeatother

\begin{document}
	\title{Learning Dual-Pixel Alignment for Defocus Deblurring}
	
	\author{Yu~Li,
		Dongwei~Ren, \emph{Member, IEEE},
  Yaling~Yi,
		Qince~Li,
		Wangmeng~Zuo, \emph{Senior Member, IEEE}
		
		\thanks{Y. Li, Y. Yi, D. Ren, Q. Li and W. Zuo are with the School of Computer Science and Technology, Harbin Institute of Technology, Harbin 150001, China (e-mail: liyuhit@outlook.com, csylyi@outlook.com, rendongweihit@gmail.com, qinceli@hit.edu.cn, cswmzuo@hit.edu.cn).}
		
	}
	
	\markboth{IEEE Transactions on Image Processing,~Vol.~xx, No.~xx, Month~Year}%
	{Shell \MakeLowercase{\textit{et al.}}: Bare Demo of IEEEtran.cls for IEEE Journals}
	\maketitle

	\begin{abstract}
		It is a challenging task to recover sharp image from a single defocus blurry image in real-world applications.  
		On many modern cameras, dual-pixel (DP) sensors create two-image views, based on which stereo information can be exploited to benefit defocus deblurring. 
        {{
        Despite the impressive results achieved by existing DP defocus deblurring methods, the misalignment between DP image views is still not studied, leaving room for improving DP defocus deblurring.}}
		%
		In this work, we propose a Dual-Pixel Alignment Network (DPANet) for defocus deblurring. 
		Generally, DPANet is an encoder-decoder with skip-connections, where two branches with shared parameters in the encoder are employed to extract and align deep features from left and right views, and one decoder is adopted to fuse aligned features for predicting the sharp image. 
		Due to that DP views suffer from different blur amounts, it is not trivial to align left and right views.  
		To this end, we propose novel encoder alignment module (EAM) and decoder alignment module (DAM).
		In particular, a correlation layer is suggested in EAM to measure the disparity between DP views, whose deep features can then be accordingly aligned using deformable convolutions. 
		DAM can further enhance the alignment of skip-connected features from encoder and deep features in decoder.  
		By introducing several EAMs and DAMs, DP views in DPANet can be well aligned for better predicting latent sharp image. 
		Experimental results on real-world datasets show that our DPANet is notably superior to state-of-the-art deblurring methods in reducing defocus blur while recovering visually plausible sharp structures and textures.  
		
	\end{abstract}
	
	
	\section{Introduction}

	\IEEEPARstart{D}{efocus} blur is a common issue when capturing scene context out of camera's depth of field (DoF). 
	In many real-world applications \cite{guo2021exploring,kong2020foveabox, sun2021supervised}, defocus blur yields undesired loss of image details. 
	It is a challenging task to recover sharp image from a single defocus blurry image due to the spatially variant blurs. 
	In pioneering works~\cite{Shi_2015_CVPR,yi2016lbp,d2016non,karaali2017edge,Lee_2019_CVPR,Park_2017_CVPR}, researchers mainly focus on estimating a defocus blur map. 
	Then non-blind deblurring methods \cite{krishnan2009fast,fish1995blind} can be adopted to predict the sharp image.
	However, these methods perform very limited in reducing defocus blur for real-world images, since the defocus blur map cannot be accurately estimated and the estimation error would be amplified by non-blind deblurring, yielding visible artifacts. 

	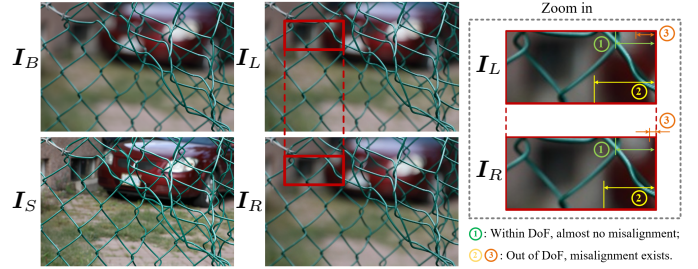
\begin{figure}[!t]
	\begin{overpic}[width=0.5\textwidth]{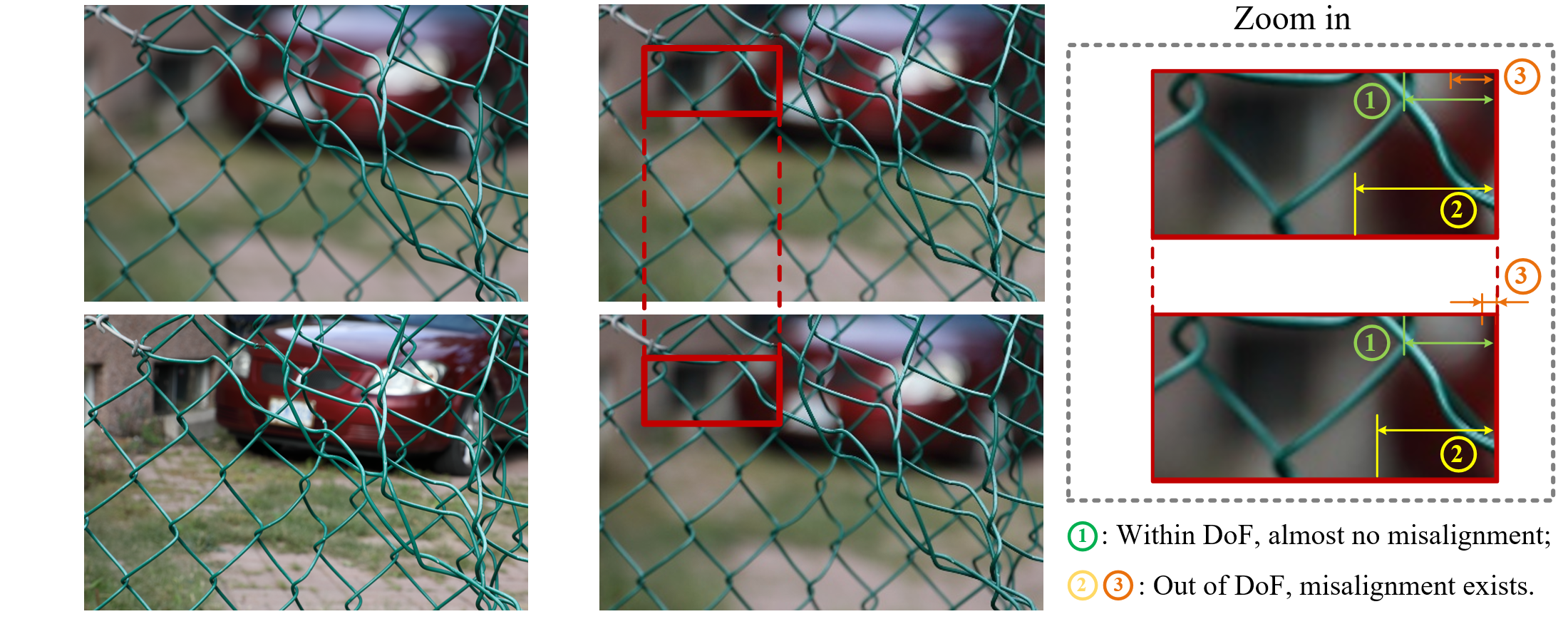}
		\put(33.8,30){\small $\bm{I}_{L}$}
        \put(33.8,9){\small $\bm{I}_{R}$}
        \put(1,30){\small $\bm{I}_{B}$}
        \put(1,9){\small $\bm{I}_{S}$}
        \put(69,29){\small $\bm{I}_{L}$}
        \put(69,13.5){\small $\bm{I}_{R}$}
	\end{overpic}
	\caption{An example of dual-pixel view images from DPDD dataset. 
For each case, there are four images, i.e., blurry image $\bm{I}_B$, left-view blurry image $\bm{I}_L$, right-view blurry image $\bm{I}_R$ and all-in-focus sharp image $\bm{I}_S$.
When objects are out of DoF, misalignments exist between left and right views.}
	\label{fig:misalignment}
\end{figure}
	
	Recently, dual-pixel (DP) sensors \cite{jang2015sensor,sliwinski2013simple,Herrmann_2020_CVPR,Abuolaim_2018_ECCV} have been widely equipped on modern cameras of smartphone and digital single-lens reflex. 
	One image captured by DP sensors can be split into two-image views, whose stereo information is beneficial to several computer vision tasks \cite{punnappurath2019reflection,garg2019learning,zhang20202,wu2020single,deng2020image,zhou2019davanet} as well as defocus deblurring \cite{pan2021dual,abuolaim2020defocus,abuolaim2020learning}. 
	In \cite{abuolaim2020defocus}, Abuolaim \emph{et al.} established a real-world dual-pixel defocus deblurring (DPDD) dataset using a Cannon camera, based on which a basic encoder-decoder is trained to learn the mapping from DP views to the latent sharp image. 
	In DDDNet \cite{pan2021dual}, DP defocus deblurring and depth estimation are simultaneously modeled, in which depth estimation and defocus deblurring are suggested to benefit each other.   
	In RDPD \cite{abuolaim2020learning}, a realistic DP defocus model and a recurrent convolution network are proposed to tackle DP defocus deblurring. 
 {{
    BaMBNet \cite{BaMBNet} firstly generates a circle of confusion map, based on which different image regions are tackled by different network branches.
    Restormer \cite{Restormer} proposes an efficient transformer to capture long-range pixel interaction, while still remaining applicable to large images.
}}
	Although these DP defocus deblurring methods have notably outperformed single image defocus deblurring methods \cite{Shi_2015_CVPR,yi2016lbp,d2016non,karaali2017edge,Lee_2019_CVPR,Park_2017_CVPR,BaMBNet,Restormer}, {{
    they neglect to align the disparity between left and right views in the regions out of camera's depth of field (DoF).
    }}

\begin{figure*}[t]
		\begin{overpic}[width=1\textwidth]{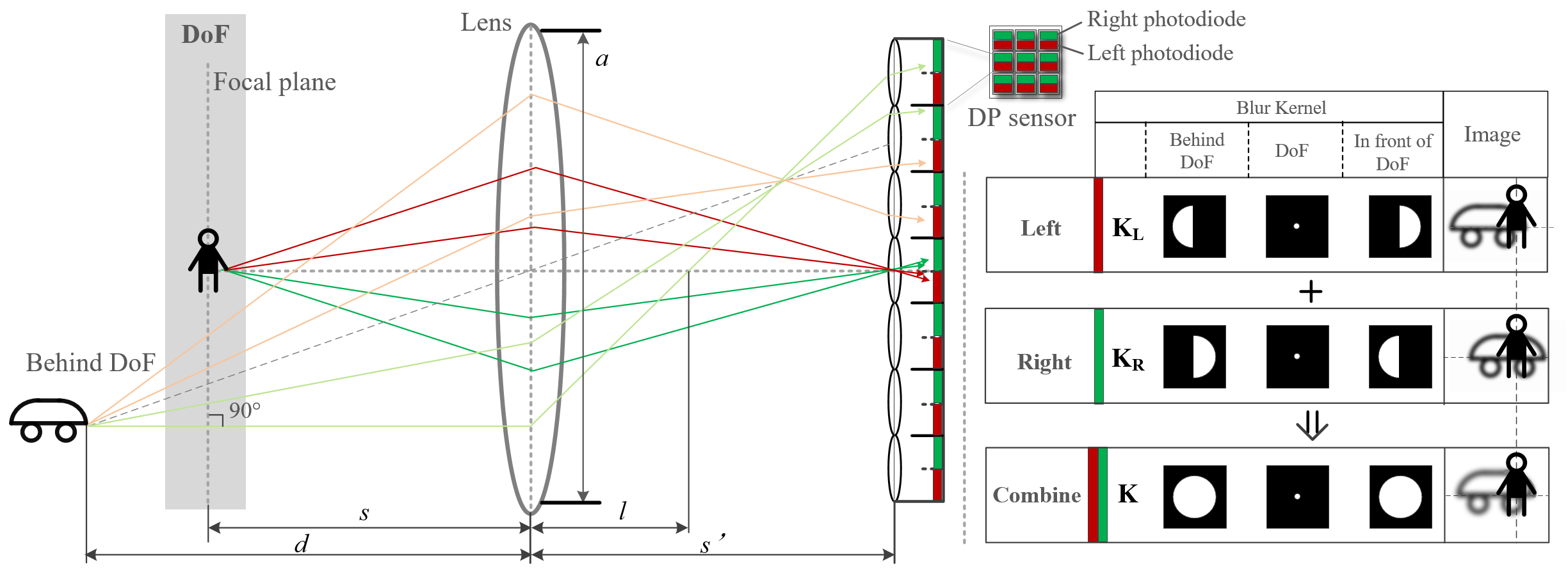}
	\end{overpic}
\caption{
  The formation of DP image pairs. The special design of DP cameras ensures that half of the light rays will be captured only by left photodiodes and the other half will be captured by right photodiodes. This imaging process makes the blurry/non-blurry visual effects 
 binded up with the misalignment/alignment in DP image pairs. For example, the person (within DoF) has a clear combined image and no misalignment bewteen DP image pairs, while the car (out of DoF) has a blurry combined image and misaligned DP image pairs.
        }
		\label{fig:dpformation}
	\end{figure*}

	As shown in Figs. \ref{fig:misalignment} and \ref{fig:dpformation}, there exists misalignment in the regions that are out of a camera's depth of field (DoF), and neither of DP views is aligned with the sharp image. 
	It is a natural strategy to align DP views for relieving the adverse effects of misalignment. 
	However, misalignment occurs in the regions out of DoF, where light rays have a relative shift, resulting in different amounts of defocus blur in left and right views and thus making it very difficult to align DP views.  
	A more critical issue is that neither of DP views is aligned with the sharp image, making it infeasible to perform alignment as a pre-processing step. 
	In this work, we resort to a dedicated network for simultaneous learning of DP alignment and defocus deblurring.

	To this end, we propose a Dual-Pixel Alignment Network (DPANet) for defocus deblurring.
	Generally, DPANet is an encoder-decoder with skip-connections \cite{ronneberger2015u, mao2016image}, where two branches in the encoder are employed to extract and align deep features of left and right views, and one decoder is adopted to fuse aligned features for predicting the latent sharp image. 
	As shown in Fig. \ref{fig:DPANet}, we propose the novel encoder alignment module (EAM) and decoder alignment module (DAM) for aligning DP views and better predicting the sharp image.  
	In the encoder, several EAMs are incorporated for gradually aligning features of left and right views. 
	In particular, a correlation layer is suggested in EAM to measure the disparity between DP views, whose deep features can then be accordingly aligned using deformable convolutions., as shown in Fig. \ref{fig:EAMDAM}(a). 
	%
	%
	Since deep features of DP views can be well aligned, we suggest to share parameters for two branches in encoder, which not only can reduce network parameters but also brings performance gains.

	In the decoder, deep features from DP views are fused to predict the sharp image, where skip-connections are adopted to better exploit shallow and deep features. 
	Since skip-connected features from the encoder may still have misalignment with deep features in the decoder, we propose to incorporate DAM in the decoder. 
	As shown in Fig. \ref{fig:EAMDAM}(b), DAM has a similar structure with EAM, where deformable convolutions are employed to further enhance the alignment of skip-connected features from the encoder and deep features in the decoder. 
	By introducing EAMs and DAMs, DP views can be well aligned in DPANet for better predicting the latent sharp image.

	Experiments are conducted on the real-wold defocus deblurring DPDD dataset \cite{abuolaim2020defocus}.
	Both quantitative and qualitative results show that our DPANet is notably superior to the state-of-the-art methods including single image defocus deblurring methods \cite{Shi_2015_CVPR, karaali2017edge, Lee_2019_CVPR}, motion deblurring methods \cite{zamir2021multi, zhang2019deep} and DP defocus deblurring methods \cite{abuolaim2020defocus,abuolaim2020learning, pan2021dual,BaMBNet,Restormer}. 
	%
	%
	Moreover, when handling real-world defocus blurry images captured by the Google PIXEL camera~\cite{abuolaim2020defocus}, our DPANet is superior to state-of-the-art methods in reducing defocus blur while recovering visually plausible sharp structures and textures.  
	
	Our contributions are three-fold:
	\begin{itemize}
		\item A dedicated network, \emph{i.e.}, DPANet, is proposed for simultaneous learning of DP alignment and defocus deblurring, in which DP views with different blur amounts can be well aligned for better predicting the sharp image. 
		\item 
		Two novel alignment modules, \emph{i.e.}, EAM and DAM, are proposed for aligning deep features of DP views in the encoder and decoder.  
		\item 
		Experiments on benchmark datasets have been conducted to verify the effectiveness of our DPANet in reducing defocus blur while recovering visually favorable structures and textures. 
		
	\end{itemize}
	
	The remainder of this paper is organized as follows: we briefly review relevant works of single image defocus deblurring and dual-pixel defocus deblurring methods in Section \ref{sec:related}. 
	In Section \ref{sec:method}, the proposed DPANet is presented in details along with two novel alignment modules. 
	In Section \ref{sec:experiment}, experiments are conducted to verify the effectiveness of our DPANet in comparison with state-of-the-art methods.
	Finally, Section \ref{sec:conclusion} ends this paper with concluding remarks.
	
	\section{Related Work}\label{sec:related}
In this section, we briefly survey relevant works including single image defocus deblurring and dual-pixel defocus deblurring methods. 
	
	\subsection{Single Image Defocus deblurring}
	\label{subsec:defocus}
	In the era of deep learning, single image motion deblurring is widely studied in terms of learning a mapping from the blurry image to the latent clean image \cite{kupyn2019deblurgan,9546678,zamir2021multi,zhang2019deep}. 
	As for single image defocus deblurring, considerable attention has been mainly paid to estimating a defocus map \cite{d2016non,karaali2017edge,zeng2018local,Lee_2019_CVPR,zhang2021joint,Park_2017_CVPR,liu2016defocus,Shi_2015_CVPR,yi2016lbp}.
	A widely adopted strategy for defocus map estimation is to calculate a sparse blur map along the edges in the image, which is then propagated across the input image to obtain the full defocus blur map.
	Karaali \emph{et al.}~\cite{karaali2017edge} proposed to smooth the sparse blur map based on pixel connectivity and then propagated it using a fast guided filter.
	Shi \emph{et al.}~\cite{Shi_2015_CVPR} detected just noticeable blur via sparse representation and image decomposition.
	Another strategy for estimating defocus blur maps resorts to machine learning techniques. 
	In \cite{d2016non}, regression tree fields are trained to label each pixel by the scale of defocus point spread function.
	In \cite{Lee_2019_CVPR}, Lee \emph{et al.} presented a deep learning method to estimate defocus maps using a domain adaption strategy.
	To achieve deblurring results, these methods should cooperate with non-blind deblurring methods~\cite{krishnan2009fast,fish1995blind} to deblur the input blurry image given the estimated defocus map.
	In this way, these methods not only lead to excessive computational cost but also their deblurring performance is very limited due to inevitable defocus map estimation error.  
	Removing defocus blur from a single image is rather challenging, and recent studies suggest that dual-pixel image pairs can be exploited to achieve notable deblurring performance gains than single image defocus deblurring methods~\cite{d2016non,karaali2017edge,Park_2017_CVPR,Lee_2019_CVPR}. 
	In this work, the proposed DPANet provides a better way to exploit DP views to benefit defocus deblurring.

	\subsection{Dual-Pixel Defocus Deblurring}
	\label{subsec:dp}
	Dual-pixel image pairs are captured by DP sensors equipped on many modern smartphones and digital single-lens reflex cameras. 
	In pioneering works, DP views are usually adopted to aid the auto-focusing~\cite{jang2015sensor,sliwinski2013simple,Herrmann_2020_CVPR}. 
	In each pixel position, a DP sensor has two photodiodes which split one pixel in half.
	Recently, researchers have applied DP views to other computer vision tasks, such as reflection removal~\cite{punnappurath2019reflection} and depth estimation~\cite{garg2019learning,zhang20202,punnappurath2020modeling}, in which stereo information can be exploited from DP views to benefit these tasks.
		
	DP views have also been suggested to benefit defocus deblurring \cite{abuolaim2020defocus,abuolaim2020learning,pan2021dual}. 
	In \cite{abuolaim2020defocus}, Abuolaim \emph{et al.} first presented a dual-pixel defocus dataset termed as DPDD, which is composed of images including 500 blurry images and corresponding sharp images.
	On DPDD, each blurry image comes along with two associated DP sub-aperture views, resulting in 500 DP views. 
	In \cite{abuolaim2020defocus}, the authors also adopted a plain encoder-decoder network with the concatenation of DP image pairs as input to recover latent sharp image.
	In DDDNet \cite{pan2021dual}, the authors simultaneously modeled depth estimation and defocus deblurring in order to achieve mutual promotion between the two tasks.	
	In \cite{abuolaim2020learning}, Abuolaim \emph{et al.} proposed a network named RDPD with convolutional LSTM units, which is designed to reduce defocus blur in both singe image and multiple blurry frames.
    {{BaMBNet \cite{BaMBNet} utilizes disparity between DP images to generate a circle of confusion map. After that 
    different image regions are assigned to different branches guided by the predicted map.
    Restormer \cite{Restormer} takes advantage of the long-range dependency by visual transformers and while effectively avoiding excessive computational overhead. Restormer can be applied to various image restoration tasks including dual-pixel defocus deblurring.
    
    }}
	Despite that existing methods \cite{abuolaim2020defocus,abuolaim2020learning,pan2021dual} have achieved {impressive} results, their performance is still limited by {{under-exploiting the close relationship between disparity aligning and defocus deblurring.
 }}
	In this work, we propose two novel DP alignment modules for relieving adverse effects of misalignment between DP views, and the proposed DPANet is superior to existing methods quantitatively and qualitatively.

	\section{Proposed Method}
	\label{sec:method}
{
 Before presenting our method, some key notations in this paper should be explained. 
In DPDD dataset, there are four images, i.e., blurry image $\bm{I}_B$, left-view blurry image $\bm{I}_L$, right-view blurry image $\bm{I}_R$ and all-in-focus sharp image $\bm{I}_S$. 
In the encoder of our proposed network, feature extractor $\mathcal{F}_E^i$ and alignment module $\mathcal{A}_E^i$ are incorporated into $i$-th encoder block, while $\mathcal{F}_D^i$ and $\mathcal{A}_D^i$ are deployed in $i$-th decoder block. 
Accordingly, $\bm{E}_L^i$ and $\bm{E}_R^i$ denote deep features from $i$-th encoder block for left-view and right-view respectively, while $\bm{D}^i$ is deep feature from $i$-th decoder block. 
The other scalars, e.g., depth $d$ and focal length $l$, are defined when they first appear. }

\subsection{Problem Formulation}
	\label{subsec:formulation}
{{
A thin-lens model is presented in Fig. \ref{fig:dpformation} to illustrate how DP image pairs are generated in DP cameras. The lens thickness is ignored to simplify the formulation.
Denoting the focal length, focus distance and aperture diameter by $l$, $s$ and $a$, respectively, the distance from lens to DP sensor is:
\begin{equation}
s' = \dfrac{s\times l}{s-l}.
\end{equation}

Considering an object with depth $d$ from the lens, its image will be blurry when it is out of DoF. In this case, a point on the object becomes a circle termed as circle of confusion in its image. The radius $r$ of the circle of confusion, which is referred to as the blur kernel, can be calculated as:
\begin{equation}\label{eq:kernelsize}
r = \dfrac{a}{2} \times \dfrac{s'}{s} \times \dfrac{d-s}{d}= \dfrac{a\times l}{2(s-l)} \times (1-\dfrac{s}{d}).
\end{equation}
	}}

 	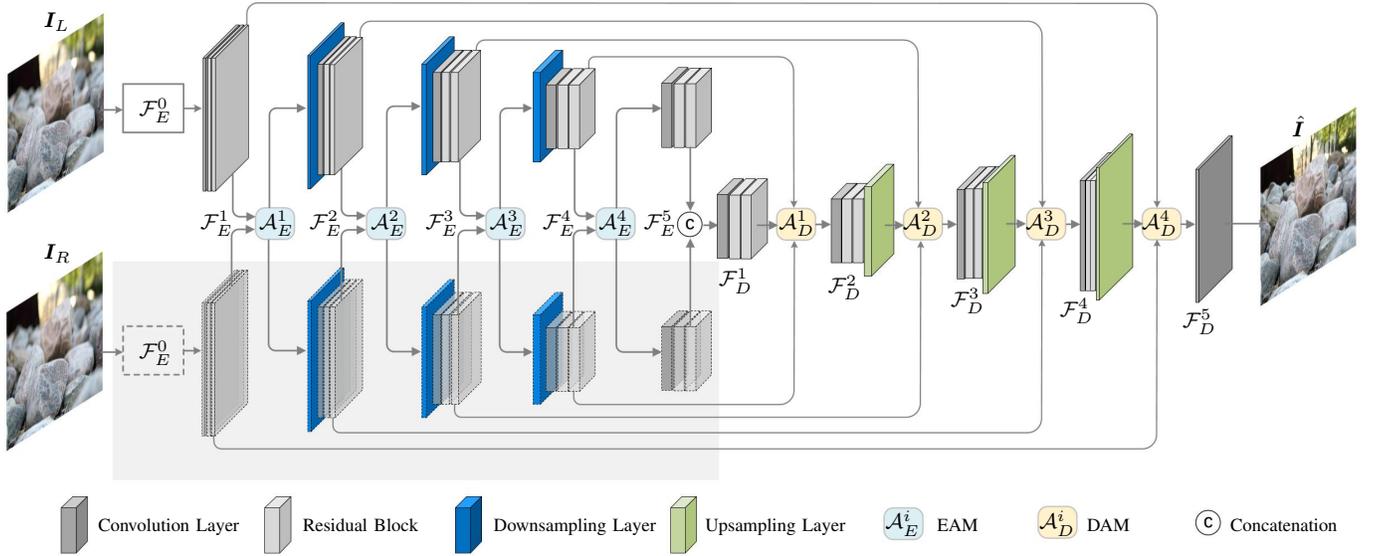
\begin{figure*}[htbp]
		\begin{overpic}[width=1\textwidth]{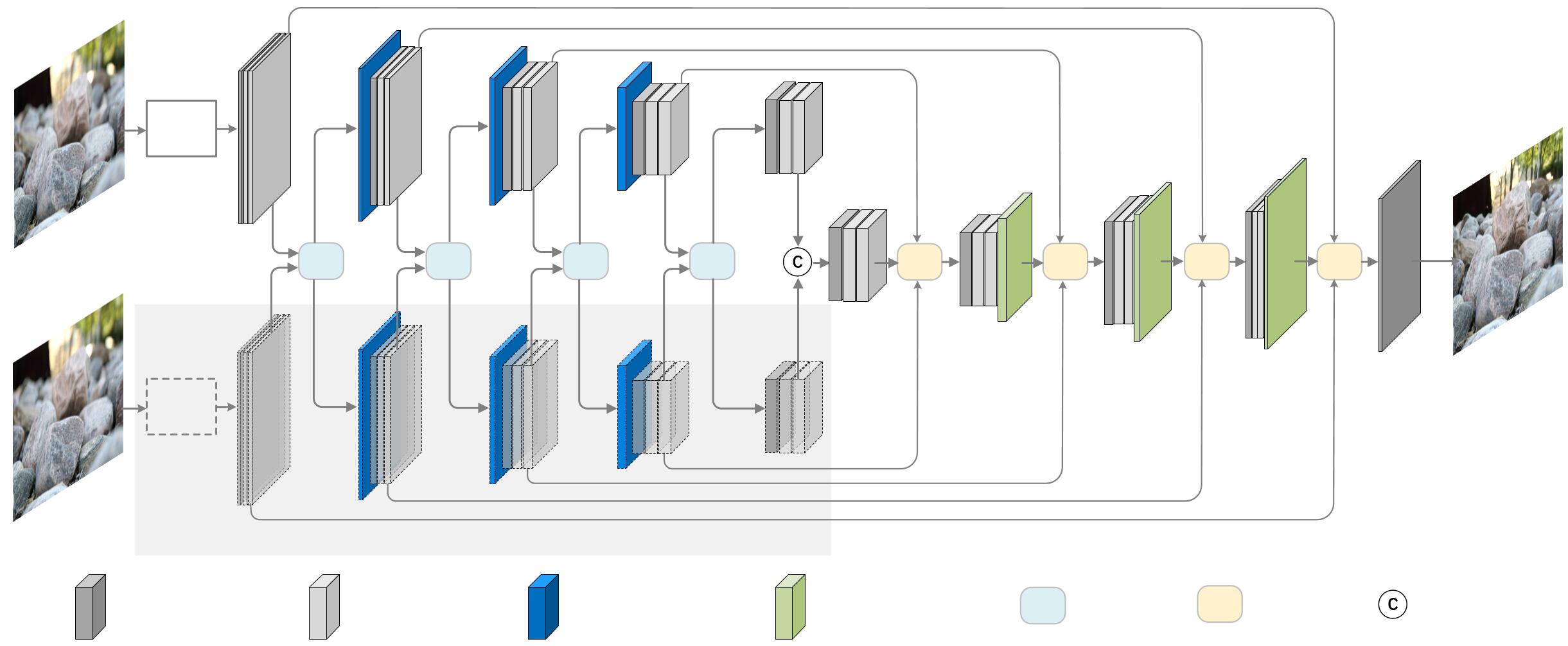}
			\put(3.5,39){\footnotesize $\bm{I}_L$}
			\put(3.5,22){\footnotesize $\bm{I}_R$}
			\put(10.5,32.5){\footnotesize $\mathcal{F}_E^0$}
			\put(10.5,14.7){\footnotesize $\mathcal{F}_E^0$}
			\put(15,24.2){\footnotesize $\mathcal{F}_E^1$}
			\put(19.3,24.2){\footnotesize $\mathcal{A}_E^1$}
			\put(23,24.2){\footnotesize $\mathcal{F}_E^2$}
			\put(27.5,24.2){\footnotesize $\mathcal{A}_E^2$}
			\put(31.5,24.2){\footnotesize $\mathcal{F}_E^3$}
			\put(36.2,24.2){\footnotesize $\mathcal{A}_E^3$}
			\put(40.1,24.2){\footnotesize $\mathcal{F}_E^4$}
			\put(44.2,24.2){\footnotesize $\mathcal{A}_E^4$}
			\put(47.5,24.2){\footnotesize $\mathcal{F}_E^5$}
			\put(53,20){\footnotesize $\mathcal{F}_D^1$}
			\put(61,19.5){\footnotesize $\mathcal{F}_D^2$}
			\put(70,18.5){\footnotesize $\mathcal{F}_D^3$}
			\put(78,17.8){\footnotesize $\mathcal{F}_D^4$}
			\put(87,17){\footnotesize $\mathcal{F}_D^5$}
			\put(57.4,24.2){\footnotesize $\mathcal{A}_D^1$}
			\put(66.5,24.2){\footnotesize $\mathcal{A}_D^2$}
			\put(75.5,24.2){\footnotesize $\mathcal{A}_D^3$}
			\put(84,24.2){\footnotesize $\mathcal{A}_D^4$}		
			\put(95,31){\footnotesize $\hat{\bm{I}}$}
			\put(65.2,2.2){\footnotesize $\mathcal{A}_E^i$}
			\put(76.5,2.2){\footnotesize $\mathcal{A}_D^i$}
			
			\put(7.5,2){\scriptsize Convolution Layer}
			\put(22.5,2){\scriptsize Residual Block}
			\put(36.5,2){\scriptsize Downsampling Layer}
			\put(52,2){\scriptsize Upsampling Layer}
			\put(69,2){\scriptsize EAM}
			\put(80,2){\scriptsize DAM}
			\put(90.5,2){\scriptsize Concatenation}
		\end{overpic}
		\caption{Architecture of the proposed DPANet for DP defocus deblurring, where two branches in the encoder are employed to extract features from DP views $\bm{I}_L$ and $\bm{I}_R$, and one decoder is adopted to predict the latent sharp image $\hat{\bm{I}}$. 
		To relieve adverse effect of misalignment between DP views, we propose EAM $\mathcal{A}_E$ and DAM $\mathcal{A}_D$ to progressively align features in the encoder and decoder, whose structures can be found in Fig. \ref{fig:EAMDAM}. 
		The network parameters are shared in two encoder branches.}
		\label{fig:DPANet}
	\end{figure*}
 
{{In DP sensors, two photodiodes are deployed at each pixel location.
As shown in Fig. \ref{fig:dpformation}, light rays from different halves of the lens will be cast onto different sides of the photodiodes. 
In this way, left and right photodiodes can record the corresponding light rays independently. Now considering a point that is out of DoF, its left-view image will be a semicircle in the right part and its right-view image will be in the left part (see the illustration in the right side of Fig. \ref{fig:dpformation}). This imaging process directly leads to the misalignment between left and right view image pairs.
\begin{itemize}
    \item 
(i) When the object is within DoF, i.e., $d=s$, both blur kernels $\bm{K}_L$ and $\bm{K}_R$ in left-view and right-view shrink to a $\delta$ kernel, making objects within DoF be sharp. 
\item
(ii) When the object is behind DoF, i.e., $d>s$, the blur kernel size $r$ will gets larger along with $d$ gets larger. 
Left-view blur kernel $\bm{K}_L$ and right-view blur kernel $\bm{K}_R$ have the opposite offsets. 
\item
(iii) When the object is in front of DoF, i.e., $d<s$, the value of $r$ will be negative. In this case, blur kernels $\bm{K}_L$ and $\bm{K}_R$ will be flipped and the kernel size will be $r$.
Thus, a smaller $d$ leads to larger blur kernels and larger misalignment between DP image pairs.
\end{itemize}
}}

\subsection{Motivation}
	\label{subsec:motivation}
	For a single image $\bm{I}_B$ with defocus blur, it is very challenging to recover latent sharp image, since defocus blur is spatially variant. 
	%
 {{
 When using DP sensors, the defocus blurry image $\bm{I}_B$ and corresponding DP blurry image pairs, \emph{i.e.}, left view $\bm{I}_L$ and right view $\bm{I}_R$, can be simultaneously obtained. 
 }}
	For a training dataset containing triplets $\mathcal{D}=\{\bm{I}_L^n, \bm{I}_R^n, \bm{I}_S^n\}_{n=1}^{N}$, where $\bm{I}_{S}$ is the ground-truth sharp image and $N$ is the number of training samples, DP defocus deblurring network can be trained to learn a mapping from DP views to the latent sharp image.

	%
  {{
  Compared with recovering the sharp image from a single blurry image, DP image pairs contains stereo information that indicates how the combined blurry image is generated, such that being beneficial to image deblurring.
  }}
	However, in the regions out of DoF, there exists misalignment between $\bm{I}_L$ and $\bm{I}_R$, and it is worse that neither of them is aligned with $\bm{I}_S$. 
	As shown in Fig. \ref{fig:misalignment}, spatial misalignment of DP views introduces extra difficulty for learning defocus deblurring networks. 
	A natural strategy is to align $\bm{I}_L$ and $\bm{I}_R$, but has two critical issues.  
	First, spatial misalignment occurs in the regions out of DoF, where $\bm{I}_L$ and $\bm{I}_R$ have different blur amounts, and cannot be well handled by off-line alignment methods \cite{ilg2017flownet,dosovitskiy2015flownet}. 
	Second, a more critical issue is that neither of $\bm{I}_L$ and $\bm{I}_R$ is aligned with $\bm{I}_S$, making it infeasible to perform alignment as a pre-processing step. 
	In this work, we propose a Dual-Pixel Alignment Network (DPANet), in which deep features of DP views $\bm{I}_L$ and $\bm{I}_R$ are gradually aligned for better predicting the latent sharp image $\bm{I}_S$.

	\begin{figure}[!t]
	\begin{overpic}[width=0.5\textwidth]{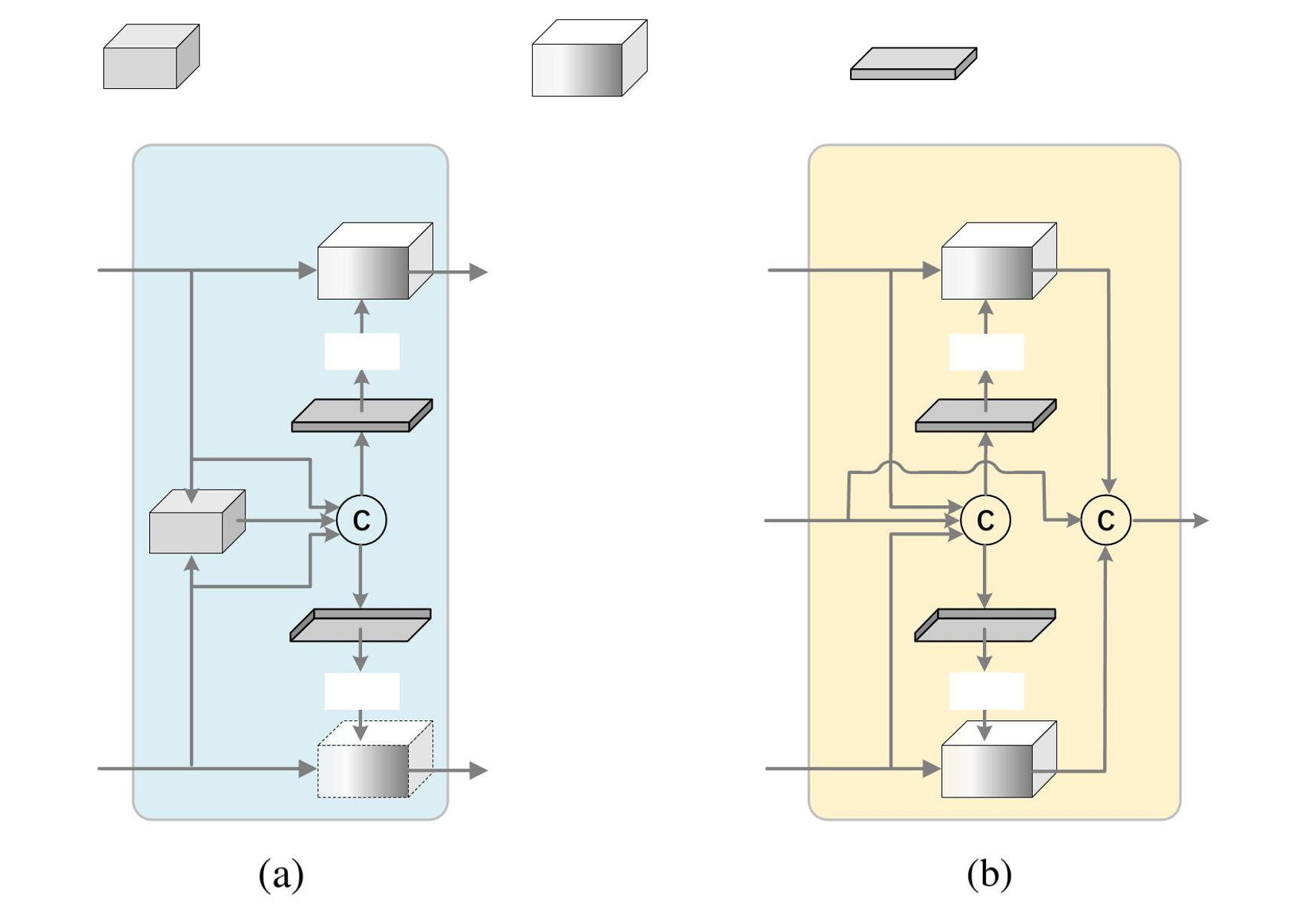}
		\put(0.5,48){\scriptsize $\widetilde{\bm{E}}_{L}^{i}$}
		\put(0.5,10.5){\scriptsize $\widetilde{\bm{E}}_{R}^{i}$}
		\put(11,54){\footnotesize $\mathcal{A}_E^i$}
		\put(62,54){\footnotesize $\mathcal{A}_D^j$}
		\put(52,48){\scriptsize $\bm{E}_{L}^{M-j}$}
		\put(52,10.5){\scriptsize $\bm{E}_{R}^{M-j}$}
		\put(38,47.5){\scriptsize $\bm{E}_{L}^{i}$}
		\put(38,9.5){\scriptsize $\bm{E}_{R}^{i}$}
		\put(52,29){\scriptsize $\bm{D}^{j-1}$}
		\put(91.5,28){\scriptsize $\widetilde{\bm{D}}^{j}$}
		\put(24.5,41.5){\tiny{${\Delta \bm{P}}_L^i$}}
		\put(24.5,15.5){\tiny{${\Delta \bm{P}}_R^i$}}
		\put(71.5,41.5){\tiny{${\Delta \bm{P}}_L^j$}}
		\put(71.5,15.5){\tiny{${\Delta \bm{P}}_R^j$}}
		\put(18,26){\scriptsize $\bm{V}^i$}
		
		\put(16,63.5){\scriptsize Correlation Layer}
		\put(51,63.5){\scriptsize DCNv2}
        \put(75,63.5){\scriptsize {{Conv layer}}}
		\put(8.3,62.8){\scriptsize Corr}
		\put(11.8,27.5){\scriptsize Corr}
		\put(41,62.5){\scriptsize DCN}
		\put(24.5,47){\scriptsize DCN}
		\put(24.5,9){\scriptsize DCN}
		\put(72,9){\scriptsize DCN}
		\put(72,47){\scriptsize DCN}

	\end{overpic}
	\caption{Structure of alignment modules. 
		(a) $i$-th EAM in the encoder aligns deep features from DP views. 
		(b) $j$-th DAM aligns skip-connected features from the encoder and deep features in the decoder.}
	\label{fig:EAMDAM}
	\end{figure}

\begin{figure}[t]
        \centering
		\begin{overpic}[width=0.49\textwidth]{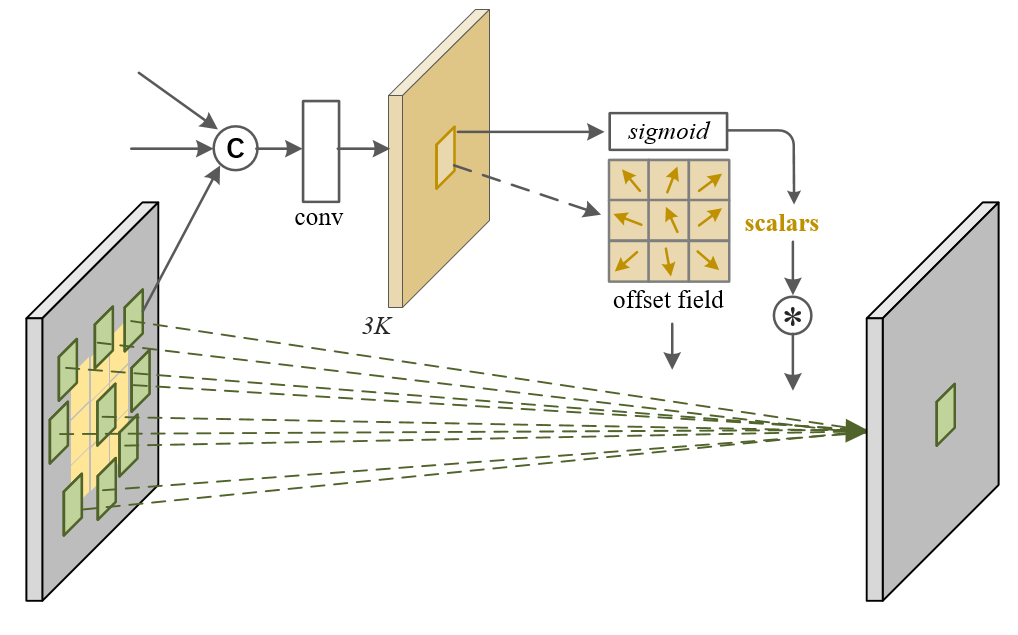}
        \put(9,47){\scriptsize $\bm{V}^i$}
        \put(9,55){\scriptsize $\widetilde{\bm{E}}_{R}^{i}$}
		\put(10,7){\scriptsize $\widetilde{\bm{E}}_{L}^{i}$}
        \put(59,26){\scriptsize $\Delta \bm{p}^k$}
        \put(70,26){\scriptsize $\Delta \bm{m}^k$}
        \put(93,7){\scriptsize $\bm{E}_{L}^{i}$}
		\end{overpic}
		\caption{{{
         Illustration of our adapted DCNv2. 
        Original DCNv2 additionally generates 2D offsets and modulation scalars by a side branch.
        The sampling locations are then rearranged by the offsets and the sampled values are rescaled by the modulation scalars. 
        To adapt to the context of DP aligning, we further feed the side branch with the feature of opposite view (e.g., the opposite view to the left is the right one) and also a cost volume. These additional inputs serve as strong hints guiding the network to align DP views.
        }}}
		\label{fig:dcnv2}
        \end{figure}
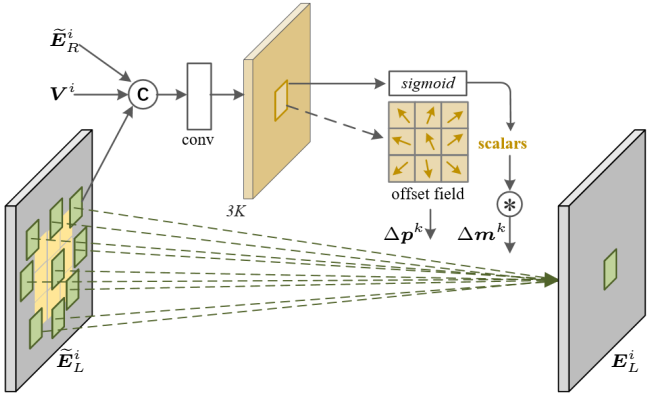
 
	\subsection{Dual-Pixel Alignment Network}
	\label{subsec:Architecture}
	
	As shown in Fig. \ref{fig:DPANet}, DPANet is generally an encoder-decoder with skip-connections, where two branches in encoder are employed to extract and align deep features from DP views $\bm{I}_L$ and $\bm{I}_R$, and one decoder aims to fuse aligned DP features for predicting the latent sharp image $\hat{\bm{I}}$. 
	To avoid adverse effects of misalignment, we propose encoder alignment module (EAM) and decoder alignment module (DAM), which are incorporated in the encoder and decoder of DPANet, respectively. 
	By respectively deploying EAM and DAM several times in DPANet, deep features extracted from $\bm{I}_L$ and $\bm{I}_R$ can be gradually aligned to the sharp image ${\bm{I}_S}$. 
	As shown in Fig. \ref{fig:alignexample}, DP features $\bm{E}_L^1$ and $\bm{E}_R^1$ after the first EAM have been largely aligned to ${\bm{I}_S}$, while misalignment of initial features $\bm{E}_L^0$ and $\bm{E}_R^0$ is consistent with input blurry DP views  $\bm{L}_L$ and $\bm{I}_R$.   
	%
	%
	Since deep features of DP views can be well aligned in our DPANet, we suggest to share network parameters of two branches in encoder, which not only can reduce network parameters but also brings deblurring performance gains. 
	In the following, we present structure details of the encoder and decoder in DPANet.

	\subsubsection{Encoder}
	Let $\mathcal{F}_E$ and $\mathcal{A}_E$ be the basic feature extractor and EAM in the encoder, respectively.
	Generally, the encoder of DPANet consists of $M$ feature extractors $\{\mathcal{F}_E^i\}_{i=1}^{M}$, between which $M-1$ alignment modules $\{\mathcal{A}_E^i\}_{i=1}^{M-1}$ are incorporated to align deep features from DP views.  
	Benefiting from $M-1$ EAMs, deep features from two views can be well aligned, and finally can be fed to decoder for predicting latent sharp image.

	Specifically, for the $i$-th block in the encoder, the features $\bm{E}_{L}^{i-1}$ from the previous block are first processed by the feature extractor,
	where the parameters of feature extractors $\mathcal{F}_E^i$ for two branches are shared. 
	Then an EAM  $\mathcal{A}_E^i$ is adopted to align the features $\widetilde{\bm{E}}_{L}^{i}$ and $\widetilde{\bm{E}}_{R}^{i}$,
    \begin{equation}
		\{\bm{E}_{L}^{i}, \bm{E}_{R}^{i}\}=\mathcal{A}_E^i(\mathcal{F}_E^i(\bm{E}_{L}^{i-1}), \mathcal{F}_E^i(\bm{E}_{R}^{i-1})),
	\end{equation}
	by which the features $\bm{E}_{L}^{i} $ and $\bm{E}_{R}^{i}$ have been aligned. 
	By introducing $M-1$ times alignment modules in the encoder, deep features from DP views can be progressively aligned. 
	Finally the block $\mathcal{F}_E^M$ is deployed to concatenate the well aligned features $\{\bm{E}_{L}^{M}, \bm{E}_{R}^{M}\}$, making it easier to predict the sharp image in the decoder.
	As for the initial features $\bm{E}_{L}^{0}$ and $\bm{E}_{R}^{0}$, we introduce Pyramid Feature Extraction Module~\cite{Wang_2019_CVPR_Workshops} to act as the initial feature extractor $\mathcal{F}_E^0$, \emph{i.e.}, $\bm{E}_{L}^{0}=\mathcal{F}_E^0(\bm{I}_L)$ and $\bm{E}_{R}^{0}=\mathcal{F}_E^0(\bm{I}_R)$. 
	The parameters of $\mathcal{F}_E^0$ for two branches are also shared.
	
	As for feature extractors $\{\mathcal{F}_E^i\}_{i=1}^{M}$, they share the similar structure, \emph{i.e.}, one convolutional layer and two Residual Blocks (ResBlocks). 
	Except $\mathcal{F}_E^1$ and $\mathcal{F}_E^M$, $2\times$ maxpooling is first introduced for downsampling the spatial feature size, while increasing $2\times$ feature channels. 
	The structure details of EAM can be found in Sec. \ref{subsec:Alignment}.

	\subsubsection{Decoder}
	The concatenation of aligned features $\{\bm{E}_{L}^{M}, \bm{E}_{R}^{M}\}$ from encoder acts as the initial feature of decoder $\bm{D}^0$. 
	Generally, the decoder also consists of $M$ feature fusion blocks $\{\mathcal{F}_D^j\}_{j=1}^{M}$, between which $M-1$ alignment modules $\{\mathcal{A}_D^j\}_{j=1}^{M-1} $ are incorporated. 
	Considering that skip-connected features from encoder may have misalignment with decoder features, especially for skip-connected features from the first several encoder blocks, $M-1$ alignment modules $\{\mathcal{A}_D^j\}_{j=1}^{M-1} $ can further enhance the alignment of deep features of the encoder and decoder.
	
	Specifically, for the $j$-th block in the decoder, skip-connected features $\bm{E}_{L}^{M-j}$ and $\bm{E}_{R}^{M-j}$ from its corresponding encoder block should be aligned to deep features in the decoder   
	\begin{equation}
	\widetilde{\bm{D}}^{j}=\mathcal{A}_D^j(\bm{D}^{j-1},\bm{E}_{L}^{M-j}, \bm{E}_{R}^{M-j}). 
	\end{equation}
	The aligned features from the encoder are concatenated with features in the decoder, and are then fused by $\mathcal{F}_D^j$
	\begin{equation}
		\bm{D}^{j}=\mathcal{F}_D^j(\widetilde{\bm{D}}_{}^{j}).
	\end{equation}

	As for the first $M-1$ feature fusion blocks $\{\mathcal{F}_D^j\}_{j=1}^{M-1}$, they share the similar structure, \emph{i.e.}, one convolutional layer and two ResBlocks.
 {{Except $\mathcal{F}_D^1$, the Nearest Neighbor Interpolation is performed to upsample the features before they are fed into DAM.}}
	As for the final block $\mathcal{F}_D^M$, it is a simple convolutional layer to map fused features $\bm{D}^{M-1}$ to the 3-channel RGB latent sharp image $\hat{\bm{I}}$.

	\subsubsection{Learning Objective}
	For training our DPANet, pixel-wise loss function can be readily adopted, such as mean square error (MSE) and $\ell_1$-norm loss. 
	Since $\ell_1$-norm loss is usually suggested to better preserve texture details, we adopt the Charbonnier loss~\cite{charbonnier1994two} as the relaxed $\ell_1$-norm loss,
	\begin{equation}
		\label{loss}
		{\rm \mathcal{L}_{deblur}}=\sum_{n=1}^{N}\sqrt{\parallel \bm{I}_S^n - \hat{\bm{I}}^n \parallel^{2} + \varepsilon^2},
	\end{equation}
	where $\varepsilon$ is a small positive value, and we empirically set it as $1\times 10^{-3}$ in all the experiments.
	
	\subsection{Dual-Pixel Alignment Modules}
	\label{subsec:Alignment}
	
	EAM and DAM are the key modules in the encoder and decoder for progressively aligning deep features from DP views. 
	In the following, we give the details of EAM and DAM.  
	
	\subsubsection{Encoder Alignment Module $\mathcal{A}_E$}
	The basic idea of EAM is to adopt deformable convolution to align deep features from DP views, in which the key issue is how to estimate spatial offsets from left and right views to the sharp image. 
	A natural strategy is to estimate offsets from the concatenation of deep features from DP views, by which the correlation between DP views cannot be fully exploited. 
	In this work, we propose a correlation-based EAM, whose structure details can be found in Fig. \ref{fig:EAMDAM}(a). 
	
	Specifically, an EAM consists of a correlation layer and a pair of modulated deformable convolutions (DCNv2~\cite{Zhu_2019_CVPR}). 
	%
	%
 {{Considering the heterogeneous misalignment amount across the image, the flexibility of sampling loactions in DCNv2 is a desirable property. In this regard, we devise a variant of DCNv2 by feeding the side branch with a concatenation of encoded DP features and a cost volume $\bm{V}$ (Fig. \ref{fig:dcnv2}).

 Formally, for the $i$-th EAM, we first compute a cost volume $\bm{V^i}$ to capture the association between features $\widetilde{\bm{E}}_L^{i}$ and $\widetilde{\bm{E}}_R^{i}$,
 \begin{equation}
    \label{correlation}
    \bm{V}^i (\bm{x}_1, \bm{x}_2) = \frac{1}{C} (\widetilde{\bm{E}}_L^{i}(\bm{x}_1))^\mathsf{T} \widetilde{\bm{E}}_R^{i}(\bm{x}_2) ,
\end{equation}
where $C$ is the cardinality of $\widetilde{\bm{E}}_L^{i}$ and $\widetilde{\bm{E}}_L^{i}$, $\bm{x}_1$ and $\bm{x}_2$ stand for element indices. Similar to~\cite{sun2018pwc}, we compute the cost volume $\bm{V}$ within a limited range of $d$ pixels, \emph{i.e.}, $|\bm{x}_1 - \bm{x}_2|_\infty \leq d$. Here, a small value of $d$ suffices, since the disparity of DP views is generally small in amount.
The offsets and modulation scalars of left view ${\Delta \bm{P}}_L^i$ and right view ${\Delta \bm{P}}_R^i$ are then estimated as follows, 
	\begin{equation}
		\label{offset_e}
		\begin{aligned}
			{\Delta \bm{P}}_L^i = f^i_L([\widetilde{\bm{E}}_L^{i},\bm{V}^i,\widetilde{\bm{E}}_R^{i}]),\\ 
			{\Delta \bm{P}}_R^i = f^i_R([\widetilde{\bm{E}}_L^{i},\bm{V}^i,\widetilde{\bm{E}}_R^{i}]), 
		\end{aligned}
	\end{equation}
where $f_L$ and $f_R$ represent the side branch convolutions for left and right views, respectively. $[.\,,\,.]$ denotes the concatenation operation. The outputs $\Delta \bm{ P}_L$ and $\Delta\bm{ P}_R$ have $3K$ channels. Taking the left view as an example, the coordinate offsets $\{\Delta \bm{p}_L^k\}_{k=1}^K$ are obtained from the first $2K$ channels of $\Delta\bm{ P}_L$, and the scalars $\{\Delta m_L^k\}_{k=1}^K$ are obtained from the last $K$ channels of $\Delta\bm{ P}_L$. With these offsets and modulation scalars, the output of the $i$-th EAM are obtained as,}}
 	\begin{equation}
		\label{align}
		\begin{aligned}
			\bm{E}_L^{i}(\bm{p}_0) = \sum_{k=1}^K w_L^k\cdot\widetilde{\bm{E}}_L^{i}(\bm{p}_0 + \bm{p}_L^k + \Delta \bm{p}_L^k) \cdot \Delta m_L^k,\\
			\bm{E}_R^{i}(\bm{p}_0) = \sum_{k=1}^K w_R^k\cdot\widetilde{\bm{E}}_R^{i}(\bm{p}_0 + \bm{p}_R^k + \Delta \bm{p}_R^k) \cdot \Delta m_R^k,
		\end{aligned}
	\end{equation}
{{where $p_0$ is the center location of a convolution window, $K$ is the size of the convolution window, $w^k$ and $\bm{p}^k$ denote the $k$-th convolution weight and pre-specified offset, respectively. For $3\times3$ deformable convolution, $K = 9$ and $\bm{p}^k \in \{(-1,-1),(-1,0),...,(1,1)\}$.}}
	%
	
	\subsubsection{Decoder Alignment Module $\mathcal{A}_D$}
	The structure details of DAM are shown in Fig.~\ref{fig:EAMDAM}(b).%
	~Since we adopt skip-connections in our DPANet, skip-connected features from the encoder are likely to still have mild misalignment with deep features of the decoder. 
	Thus, we propose DAM for aligning deep features in the decoder. 
	Generally, DAM also consists of two deformable convolutions. 
	Since skip-connected features have been largely aligned in encoder, we suggest to estimate offsets directly from the concatenation of deep features. 
	The offset fields here are expected to convey the misalignment of deep features from the encoder and decoder. 

 \begin{figure*}[t]
  \centering
	\begin{overpic}[width=0.8\textwidth]{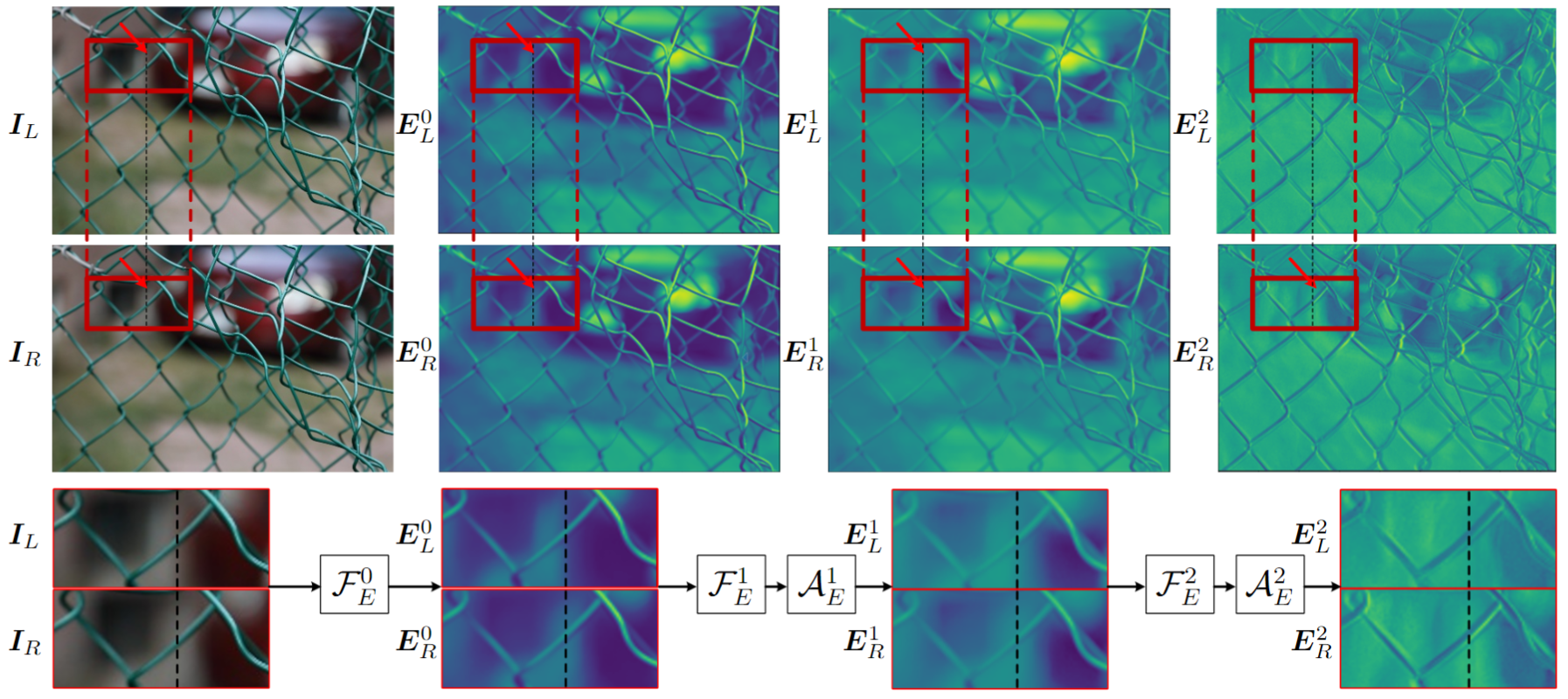}
	\end{overpic}
	\caption{{{An example of the aligning process.}}}
	\label{fig:alignexample}
\end{figure*}

	For the $j$-th DAM, skip-connected features $\bm{E}_L^{M-j}$ and $\bm{E}_R^{M-j}$ from the encoder can be aligned to deep features in the decoder. 
    {{
    Similar to the adapted DCNv2 in EAM, the offsets and modulation scalars of the $j$-th DAM are estimated as}}
	\begin{equation}
		\label{offset_e}
		\begin{aligned}
			{\Delta \bm{P}}_L^j = f^j_L([\bm{E}_L^{M-j},\bm{D}^{j-1},\bm{E}_R^{M-j}]),\\ 
			{\Delta \bm{P}}_R^j = f^j_R([\bm{E}_L^{M-j},\bm{D}^{j-1},\bm{E}_R^{M-j}]).
		\end{aligned}
	\end{equation}
	Also $\Delta\bm{ P}^j_L$ and $\Delta\bm{P}^j_L$ have $3K$ channels, where the first $2K$ channels are extracted for coordinate offsets and the last $K$ channels are used to obtain modulation scalars.  
	Then the aligned features $\widetilde{\bm{D}}_L^{j}$ and $\widetilde{\bm{D}}_R^{j}$ can be generated by
	\begin{equation}
		\label{align}
		\begin{aligned}
			\widetilde{\bm{D}}_L^{j}(\bm{p}_0) = \sum_{k=1}^K w_L^k\cdot\bm{E}^{M-j}(\bm{p}_0 + \bm{p}_L^k + \Delta \bm{p}_L^k) \cdot \Delta m_L^k,\\
			\widetilde{\bm{D}}_R^{j}(\bm{p}_0) = \sum_{k=1}^K w_R^k\cdot\bm{E}^{M-j}(\bm{p}_0 + \bm{p}_R^k + \Delta \bm{p}_R^k) \cdot \Delta m_R^k,
		\end{aligned}
	\end{equation}
	where features $\widetilde{\bm{D}}_{L}^{j}$ and $\widetilde{\bm{D}}_{R}^{j}$ have been well aligned.
	%
	Finally, the fused features of the $j$-th decoder block can be obtained by
	\begin{equation}
		\begin{aligned}
			\bm{D}^{j} = \mathcal{F}_D^{j}(\widetilde{\bm{D}}^{j})=\mathcal{F}_D^{j}([\widetilde{\bm{D}}_{L}^{j}, \bm{D}^{j-1},\widetilde{\bm{D}}_{R}^{j}]).
		\end{aligned}
	\end{equation}

	\begin{table*}[!t]
		\centering
		\caption{
			Quantitative comparison of competing methods on the DPDD test set. 
			We take state-of-the-art single image defocus deblurring methods, motion deblurring methods and DP defocus deblurring methods into comparison. 
			The best results are highlighted in bold.}
		\small
		\label{tab:comparison}
		\begin{tabular}{c|ccc|cc|cccc|c}
			\hline
			
			\hline
			\multicolumn{1}{c}{\multirow {3}{*}{Method}}&\multicolumn{3}{|c|}{Single Image Defocus Deblurring}& \multicolumn{2}{c|}{Motion Deblurring} & \multicolumn{5}{c}{DP Defocus Deblurring}\\
			\cline{2-11}
			{}&  JNB & EBDB& DMENet &DMPHN & MPRNet & DPDNet & RDPD+&BaMBNet & Restormer &DPANet \\
			{}&\cite{Shi_2015_CVPR}&\cite{karaali2017edge}&\cite{Lee_2019_CVPR}&\cite{zhang2019deep}&\cite{zamir2021multi}&\cite{abuolaim2020defocus}&\cite{abuolaim2020learning}&\cite{BaMBNet}&\cite{Restormer}& \\
			\hline
			PSNR $\uparrow$ & 22.13 & 23.19 & 23.32 & 25.33 & 25.68 & 25.13 & 25.39 & 26.40 & 26.66 & \textbf{26.80} \\
			SSIM $\uparrow$ & 0.676 & 0.713 & 0.715 & 0.781 & 0.789 & 0.786 & 0.772 & 0.821 & 0.833& \textbf{0.835} \\
			MAE $\downarrow$ & 0.056 & 0.051 & 0.051 & 0.040 & 0.039 & 0.041 & 0.040 & 0.036 & \textbf{0.035}& \textbf{0.035} \\
			LPIPS $\downarrow$ & 0.480 & 0.420 & 0.411 & 0.220 & 0.229 & 0.226 & 0.256 & 0.174 & \textbf{0.155}& 0.159 \\ 	
			\hline
			
			\hline
		\end{tabular}
	\end{table*}
	
	To summarize, we propose EAM and DAM for aligning deep features of DP views in the encoder and decoder, respectively. 
	By introducing EAM and DAM several times in DPANet, adverse misalignment effects with different blur amounts in left and right views can be well relieved, making our DPANet succeed in reducing defocus blur while recovering more visually plausible sharp structures and textures.

\subsection{Discussion}
{Considering misalignment between left and right views, dual-pixel defocus deblurring actually involves two problems, i.e., alignment of DP views and defocus deblurring. 
In previous methods \cite{Abuolaim_2018_ECCV,abuolaim2020defocus}, a deep network, e.g., UNet in DPDNet, is deployed to force intermediate deep features aligned to all-in-focus image when learning defocus deblurring, which actually brings extra training difficulty. 
Our DPANet is a unified framework to jointly tackle dual-pixel alignment and deblurring. 
Considering that it is not trivial to give theory analysis as that in deep learning approaches, we visualize the intermediate features of left and right views to observe how the alignment modules work, as shown in Fig. \ref{fig:alignexample}. 
Generally, offsets $\Delta \bm{P}_L$ and $\Delta \bm{P}_R$ in alignment modules can guide left and right views to be gradually aligned. 
In $\bm{I}_L$ and $\bm{I}_R$, the gaps between car and black dotted lines are obviously different due to the misalignment in defocus blurry regions.  
By passing through the first encoder block $\mathcal{F}_E^0$, the gaps in left-view and right-view features $\bm{E}_L^0$ and $\bm{E}_R^0$ are almost same with those in input $\bm{I}_L$ and $\bm{I}_R$, because $\mathcal{F}_E^0$ is only a feature extractor without introducing our alignment module. 
Then by passing through $\mathcal{F}_E^1$ and $\mathcal{F}_E^2$ with alignment modules $\mathcal{A}_E^1$ and $\mathcal{A}_E^2$, the misalignment between left and right views can be gradually relieved, by which the gaps between car and black dotted lines in left-view and right-view features are becoming the same. 
Benefiting from alignment modules, defocus deblurring is tackled based on aligned DP features, and thus deblurring performance can be boosted. }
	
	\section{Experiments}
	\label{sec:experiment}
	
	In this section, we first present the implementation details of DPANet, then show experimental results in comparison with state-of-the-art methods, and finally analyze the proposed DPANet in ablation study. 
	The source code of DPANet has been publicly available at \url{https://github.com/liyucs/DPANet}, from where more experimental results can be found.

	\subsection{Implementation Details}
	\label{subsec: Implementation Details}
	
	As for the network architecture, the number of encoder and decoder blocks is set as $M = 5$, and the pixel range for computing cost volume $\bm{V}$ is set as $d=9$. 
 {{
 For the deformable convolution, the initial values of $\Delta \bm{p}_L^k$ and $\Delta m_L^k$ are set as $\bm{0}$ and 0.5, respectively. 
 }}
	Following \cite{abuolaim2020defocus}, DPANet is trained on the DPDD dataset \cite{abuolaim2020defocus} captured by Cannon camera, on which $N=350$ training triplets are available, 76 images are used for testing while the remaining images for validation. 
	Training images are cropped into patches with size $512\times512$, and the batch size is set as 4. 
	During training, the parameters of DPANet are initialized using the strategy proposed by He \emph{et al.} \cite{he2015delving}, and are optimized using the Adam \cite{kingma2014adam} algorithm, where the learning rate is initialized as $2\times10^{-5}$, and is halved after every 60 epochs. 
	The training procedure ends with 150 epochs. 
	All the experiments are conducted in PyTorch~\cite{paszke2019pytorch} environment running on a PC with one NVIDIA TITAN RTX GPU.

	\begin{figure*}[!htbp]
		\tiny
		\small
		\scriptsize
		\centering
		\begin{tabular}{ccccc}
			\footnotesize
			\hspace{-0.4cm}
			\begin{adjustbox}{valign=t}
				\begin{tabular}{cccccccc}
					\includegraphics[width=0.12\textwidth]{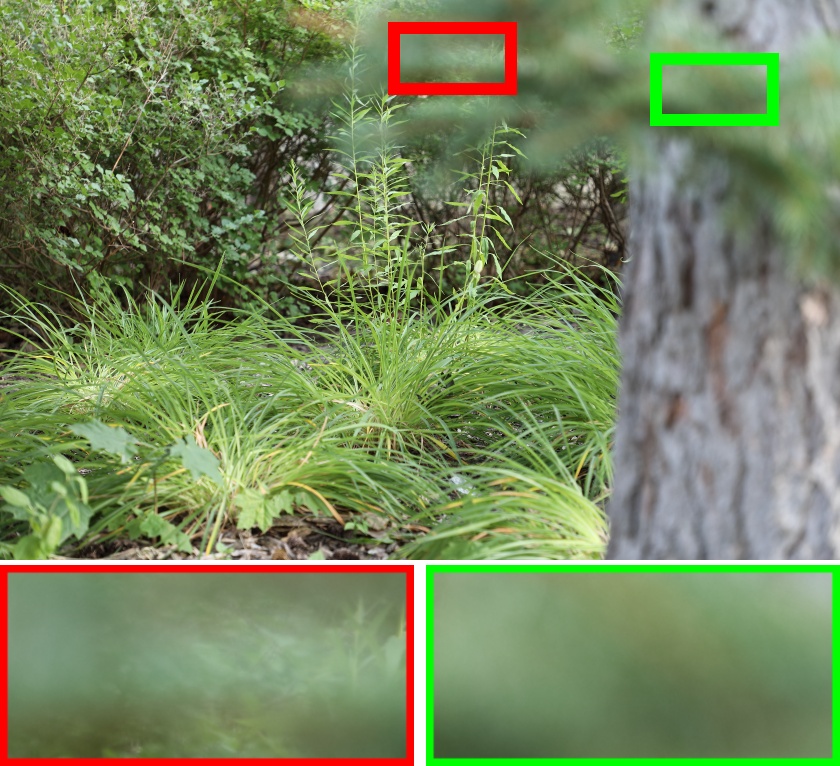}&\hspace{-4.2mm}
					\includegraphics[width=0.12\textwidth]{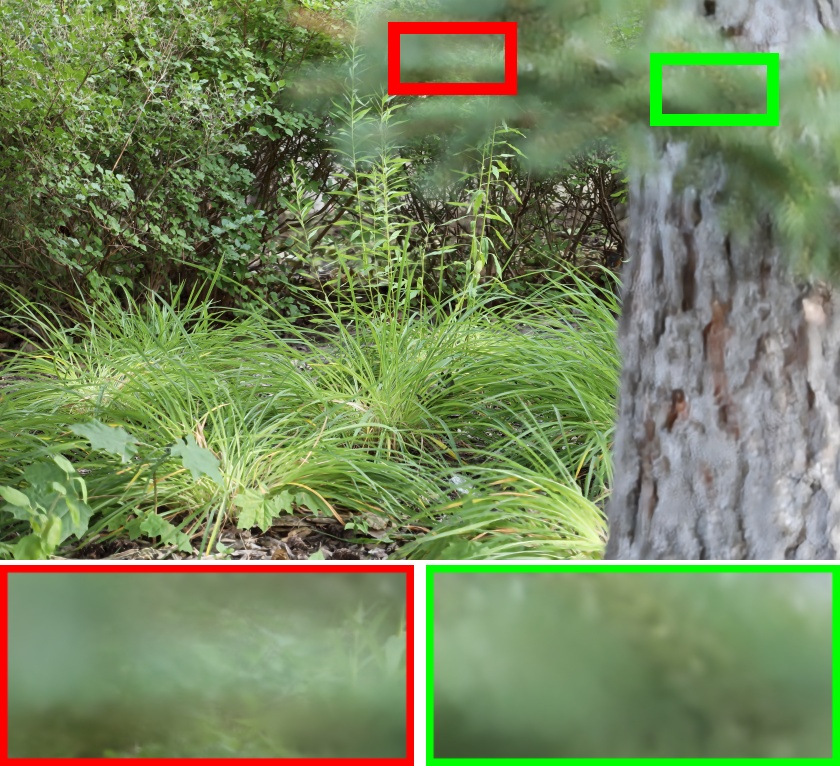}&\hspace{-4.2mm}
					\includegraphics[width=0.12\textwidth]{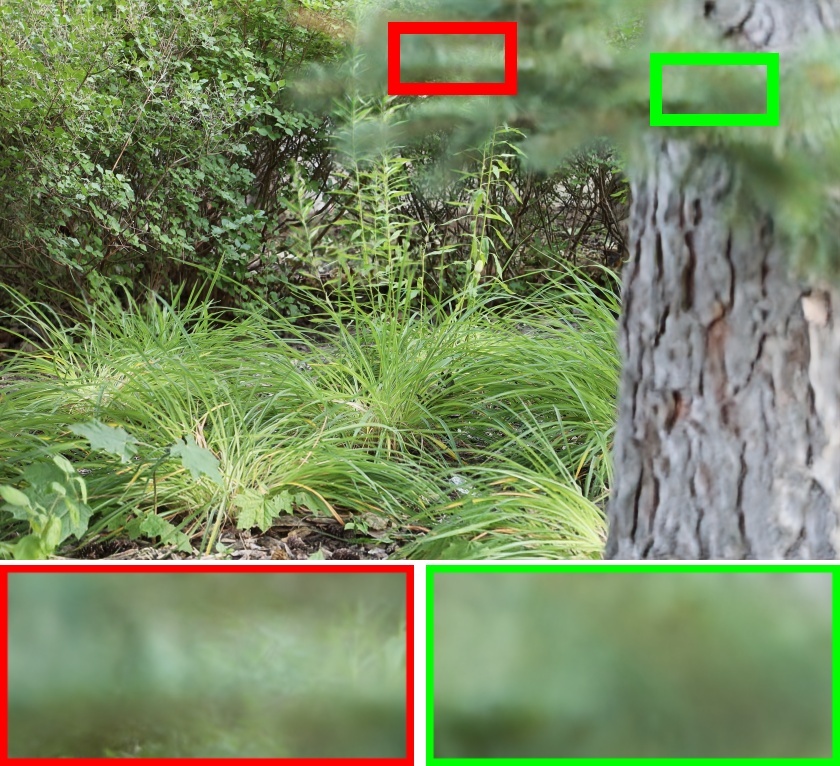}&\hspace{-4.2mm}
					\includegraphics[width=0.12\textwidth]{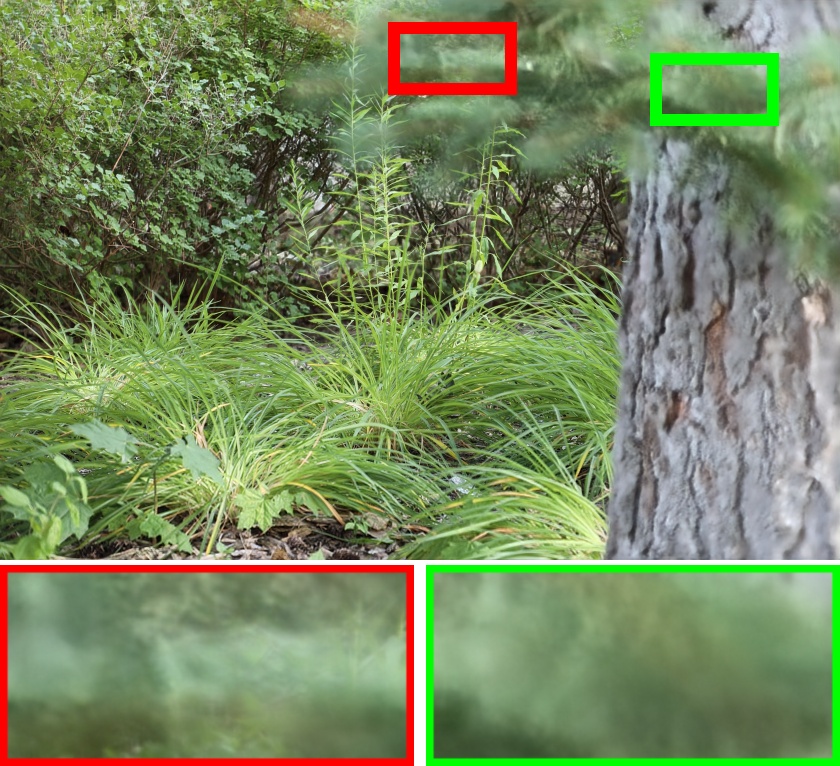}&\hspace{-4.2mm}
					\includegraphics[width=0.12\textwidth]{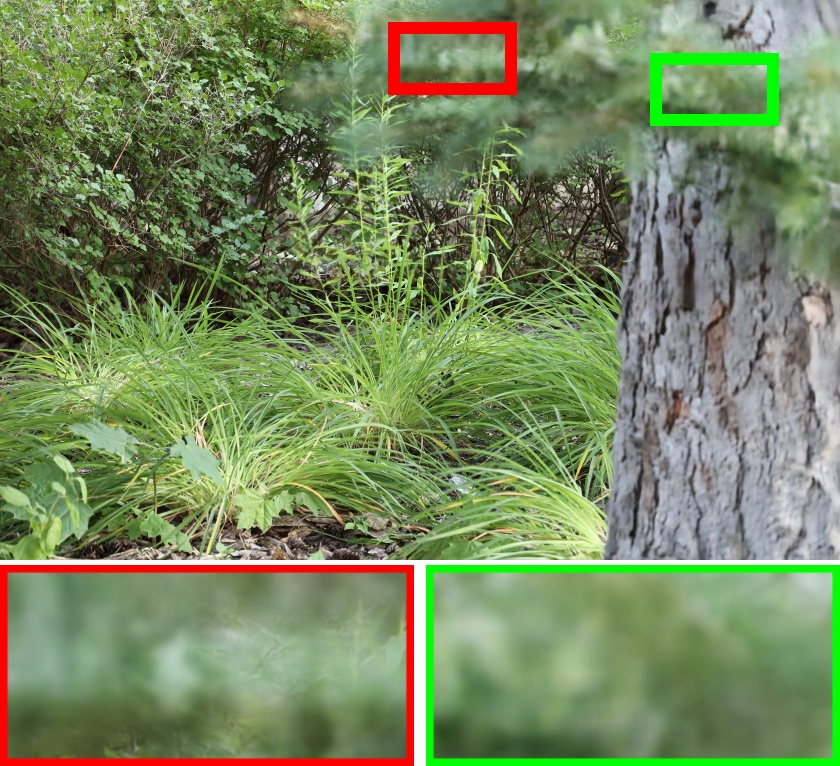}&\hspace{-4.2mm}
					\includegraphics[width=0.12\textwidth]{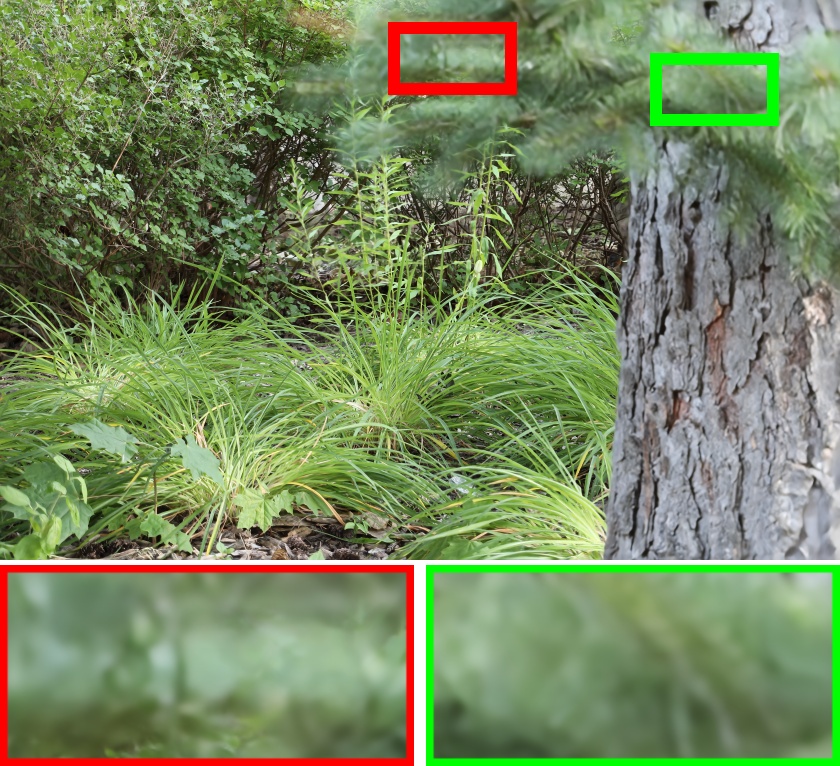}&\hspace{-4.2mm}
					\includegraphics[width=0.12\textwidth]{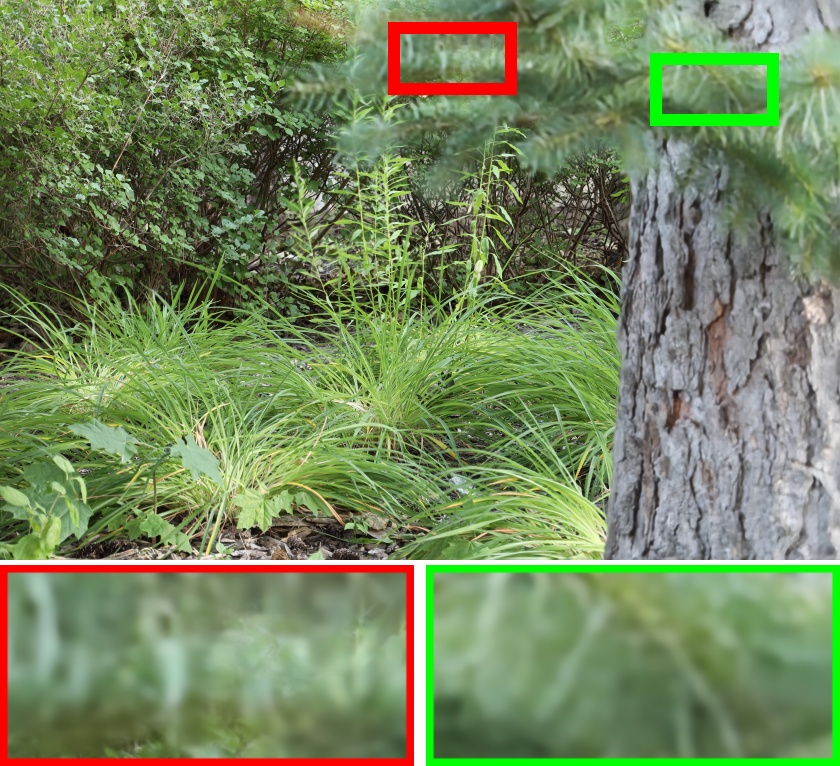}&\hspace{-4.2mm}
					\includegraphics[width=0.12\textwidth]{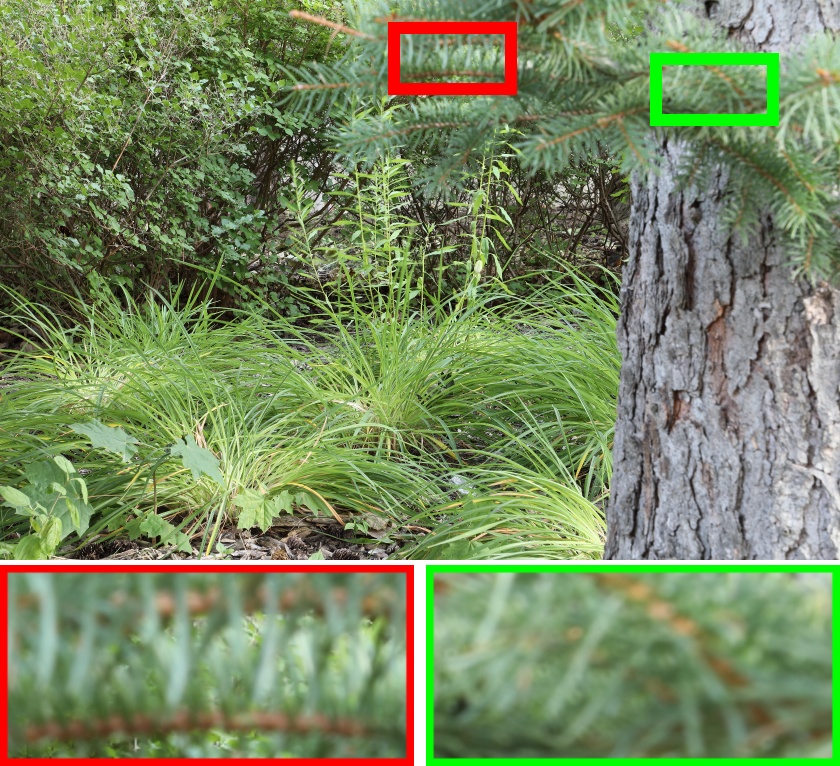}
					\\
					\includegraphics[width=0.12\textwidth]{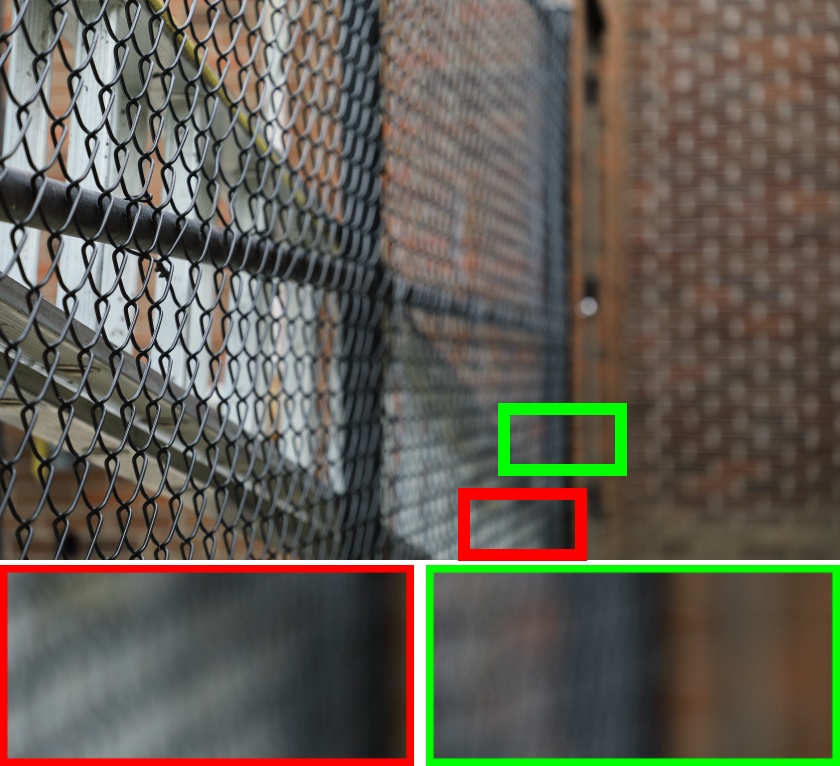}&\hspace{-4.2mm}					\includegraphics[width=0.12\textwidth]{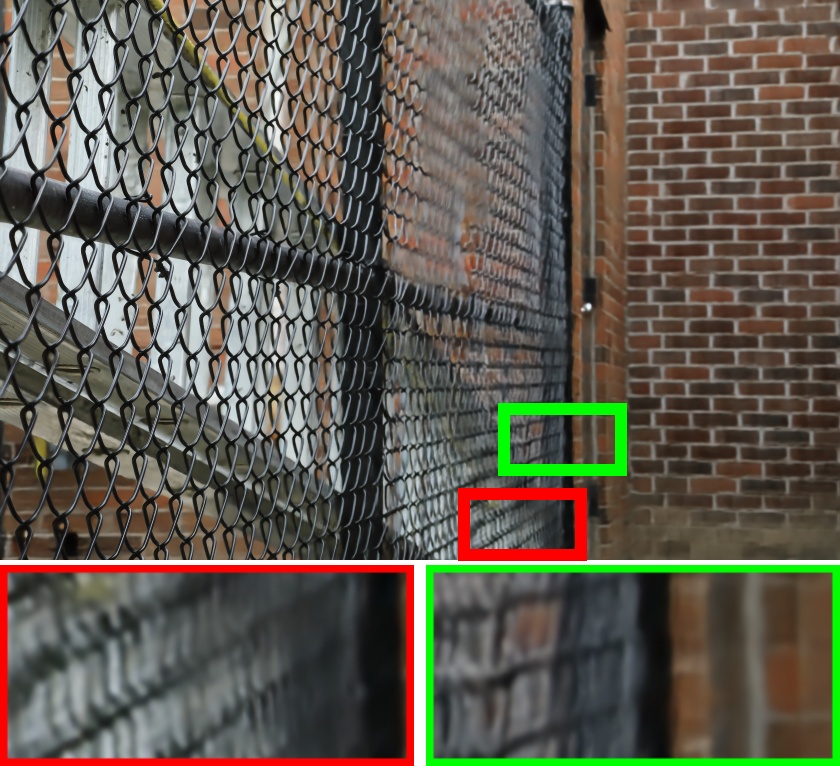}&\hspace{-4.2mm}
					\includegraphics[width=0.12\textwidth]{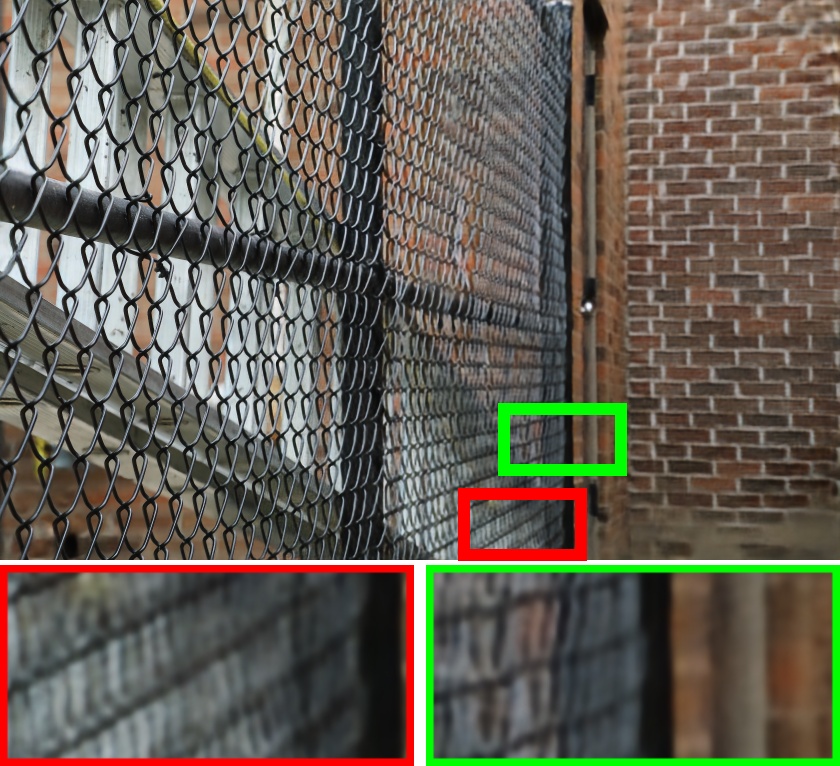}&\hspace{-4.2mm}
					\includegraphics[width=0.12\textwidth]{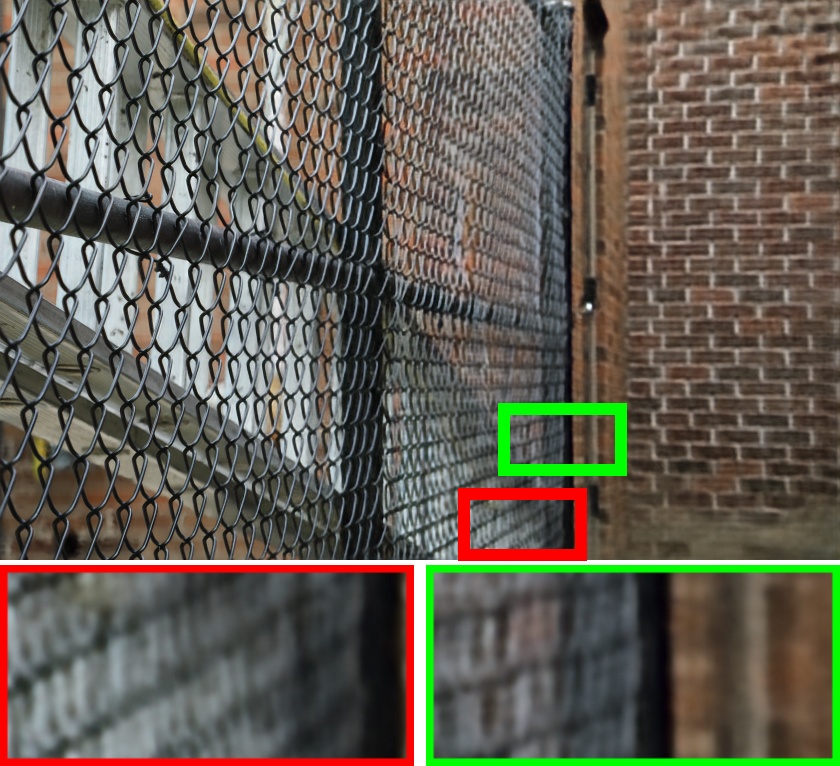}&\hspace{-4.2mm}
					\includegraphics[width=0.12\textwidth]{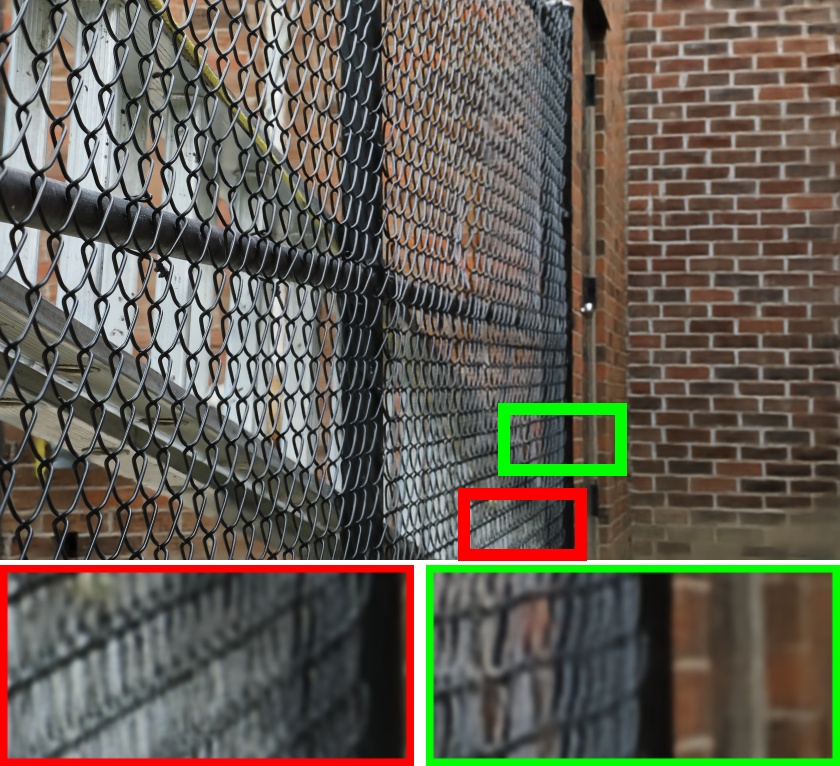}&\hspace{-4.2mm}
					\includegraphics[width=0.12\textwidth]{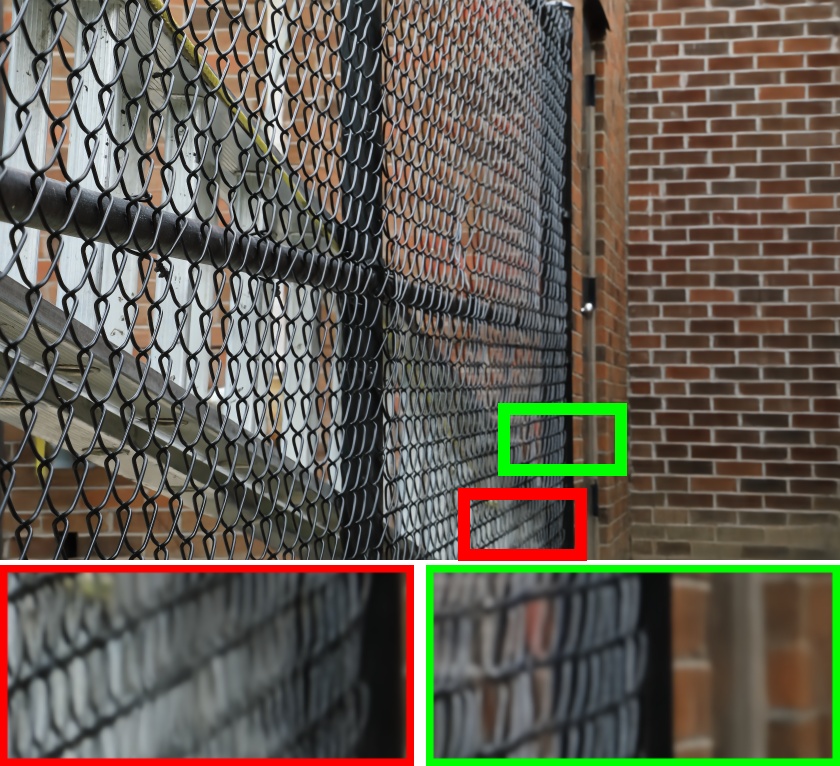}&\hspace{-4.2mm}
					\includegraphics[width=0.12\textwidth]{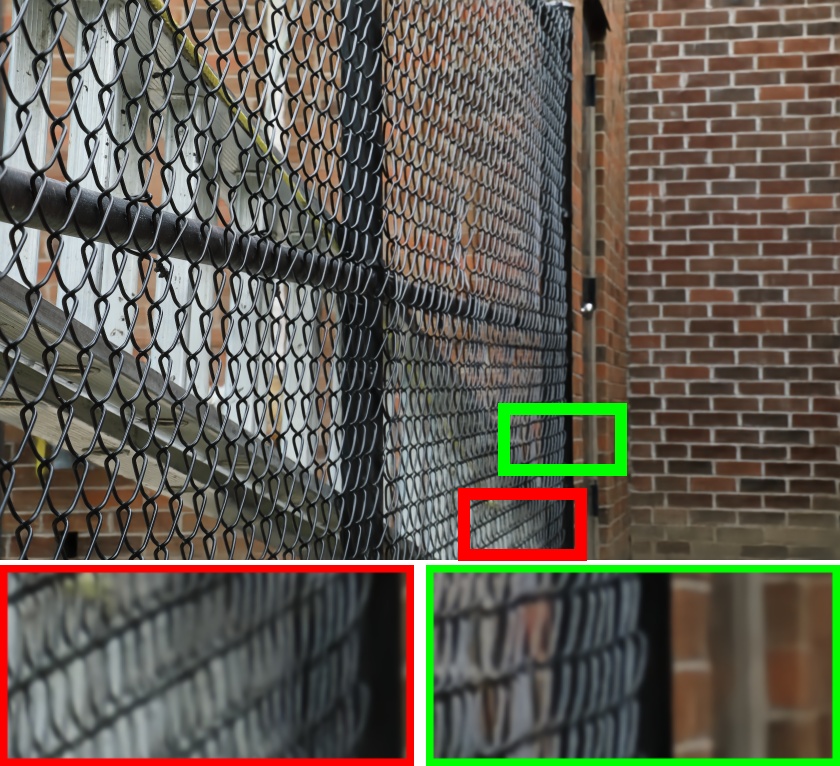}&\hspace{-4.2mm}
					\includegraphics[width=0.12\textwidth]{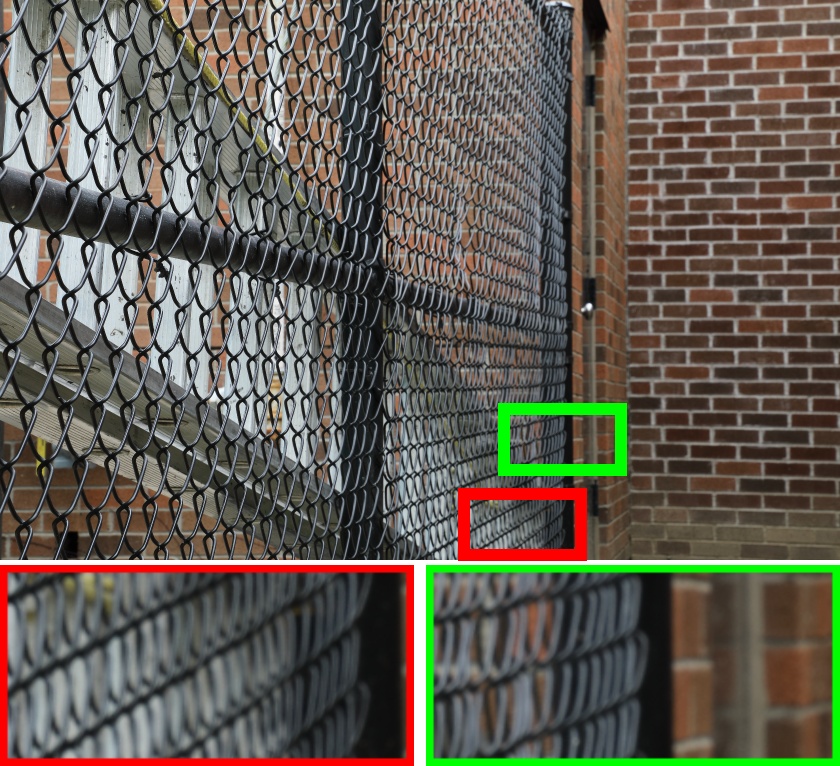}
					\\
					\includegraphics[width=0.12\textwidth]{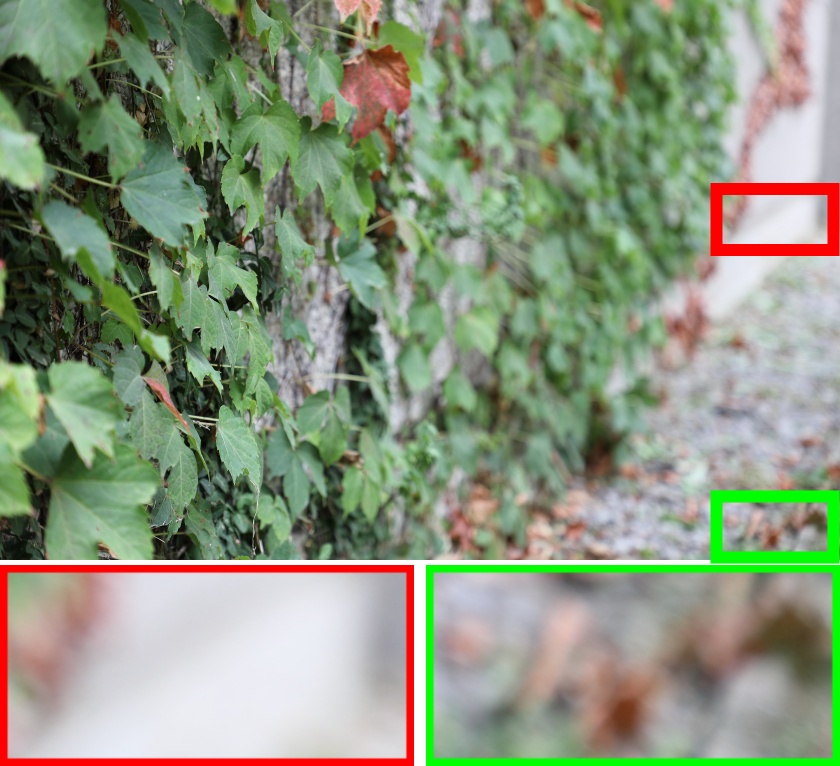}&\hspace{-4.2mm}
					\includegraphics[width=0.12\textwidth]{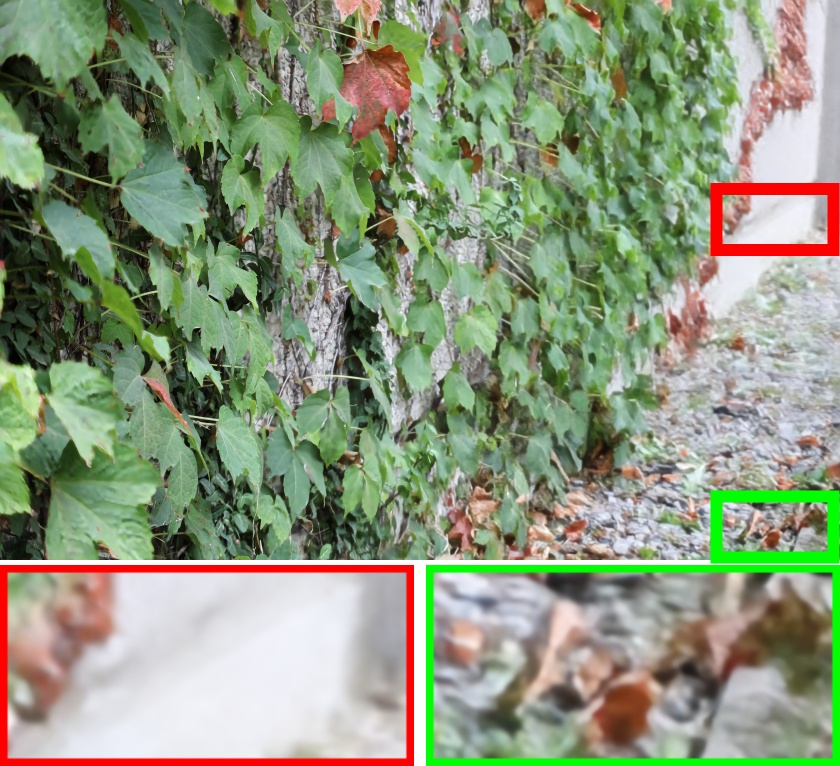}&\hspace{-4.2mm}
					\includegraphics[width=0.12\textwidth]{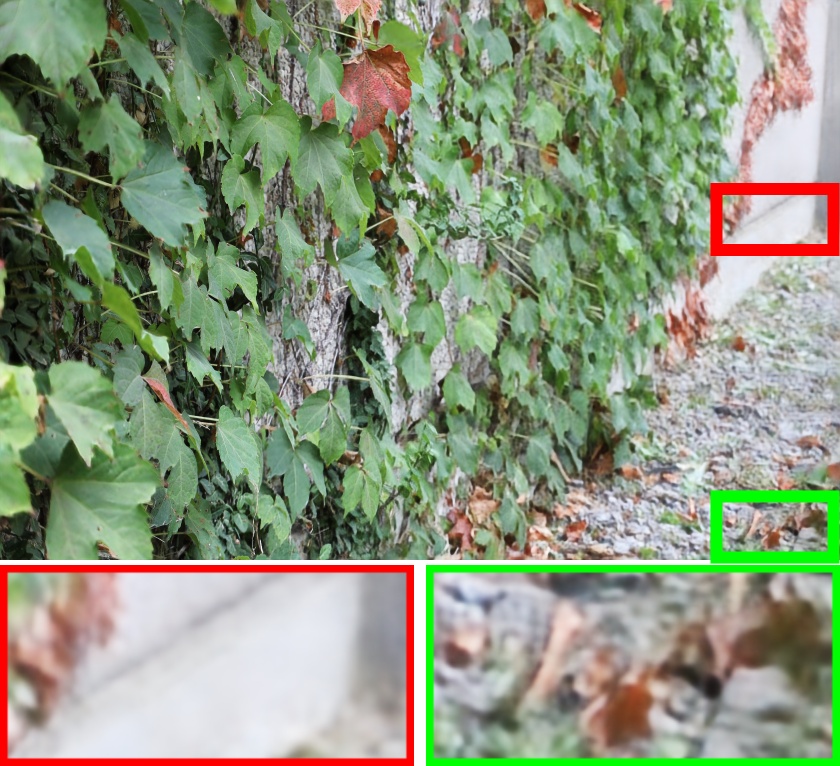}&\hspace{-4.2mm}
					\includegraphics[width=0.12\textwidth]{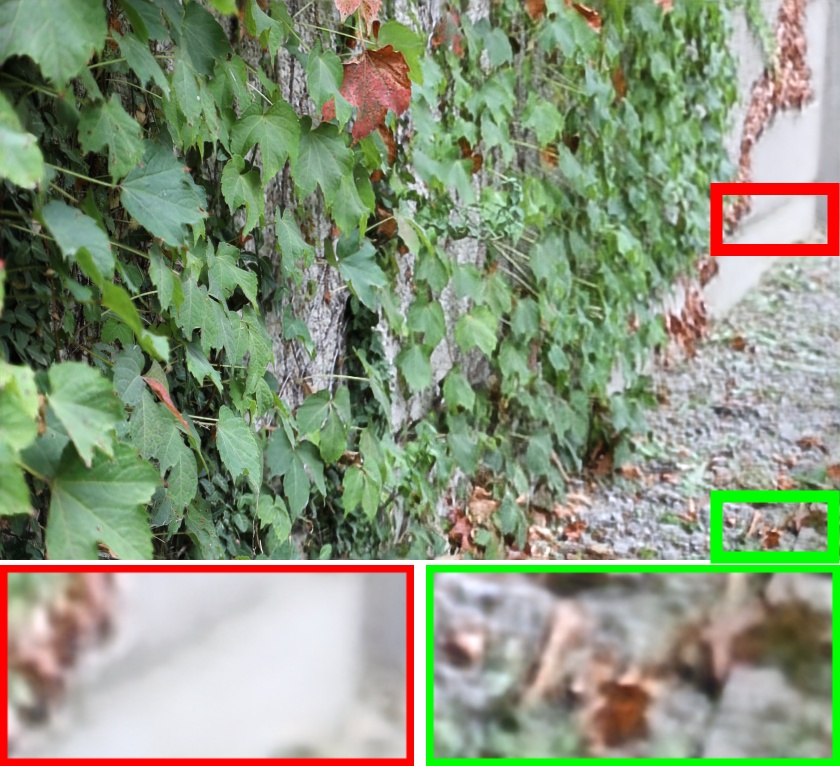}&\hspace{-4.2mm}
					\includegraphics[width=0.12\textwidth]{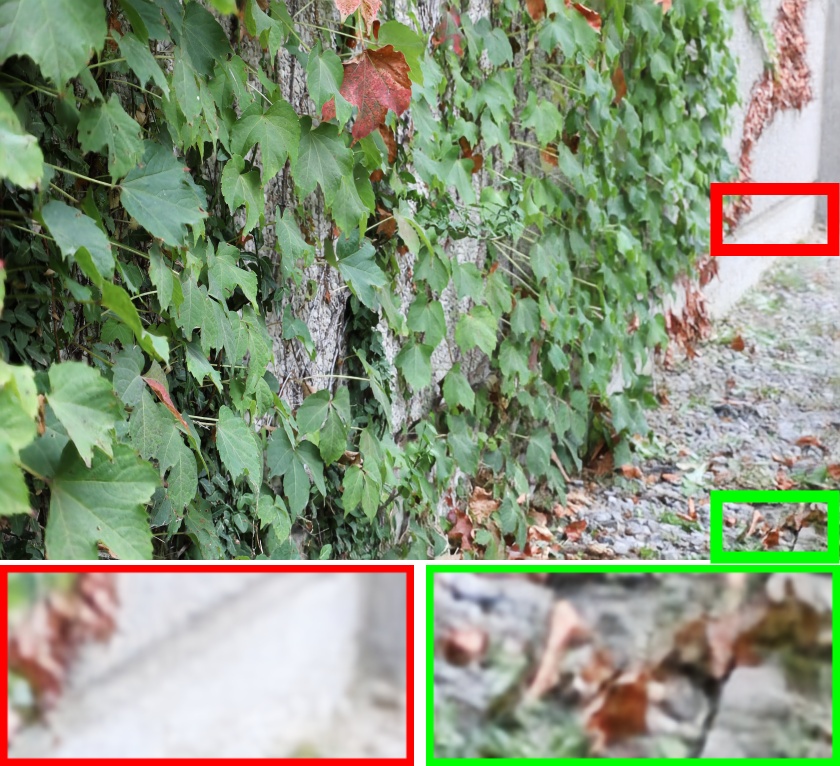}&\hspace{-4.2mm}
					\includegraphics[width=0.12\textwidth]{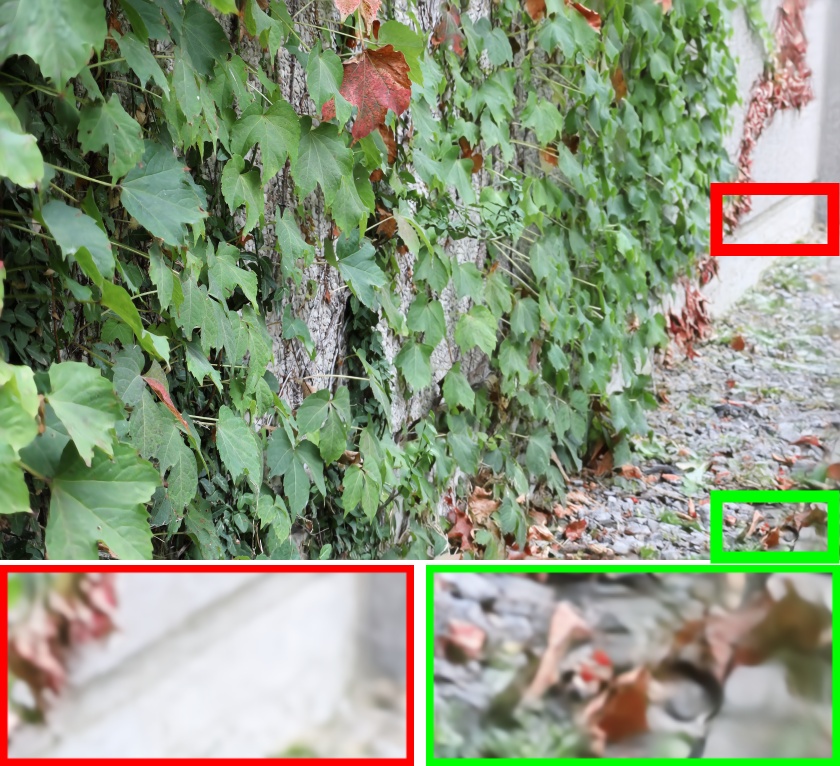}&\hspace{-4.2mm}
					\includegraphics[width=0.12\textwidth]{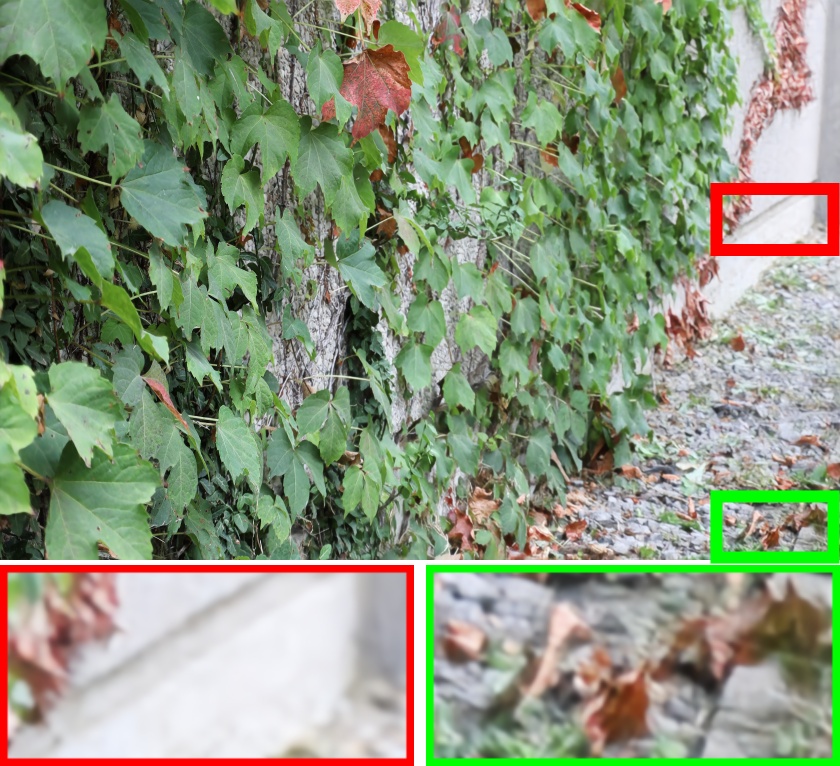}&\hspace{-4.2mm}
					\includegraphics[width=0.12\textwidth]{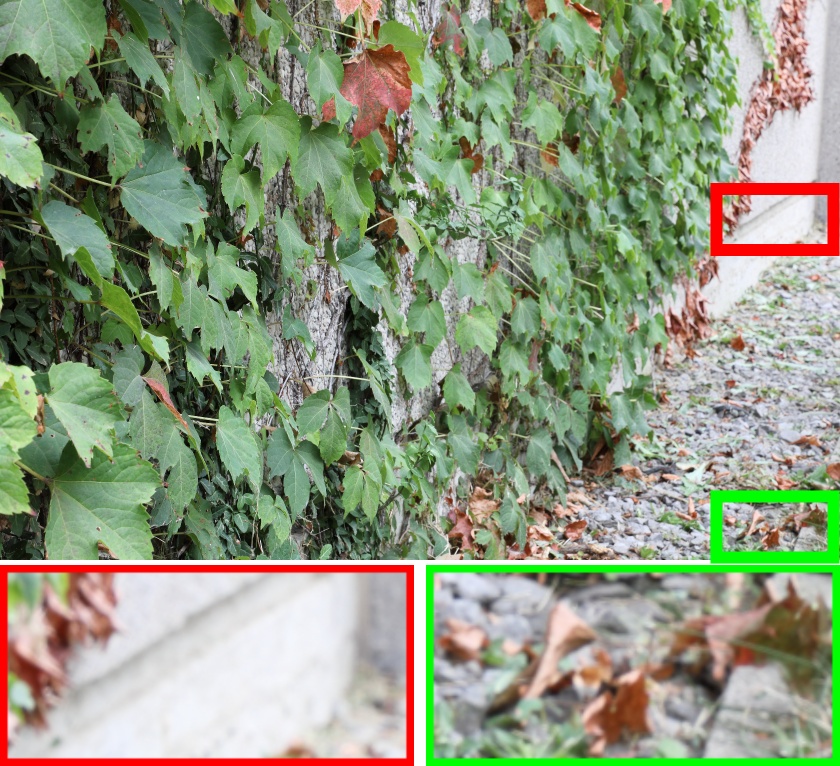}
					\\
					\includegraphics[width=0.12\textwidth]{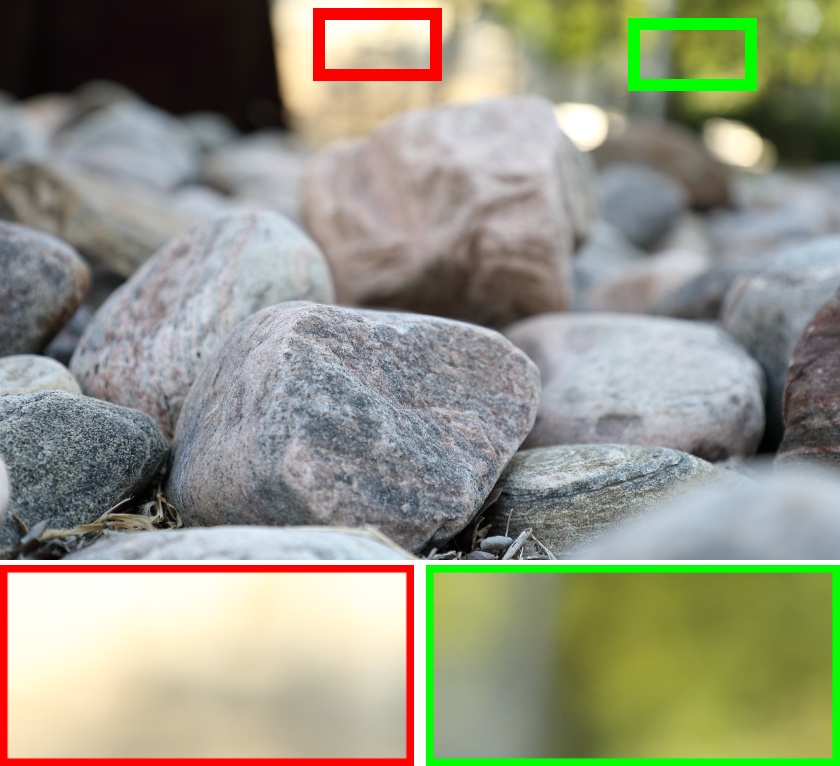}&\hspace{-4.2mm}
					\includegraphics[width=0.12\textwidth]{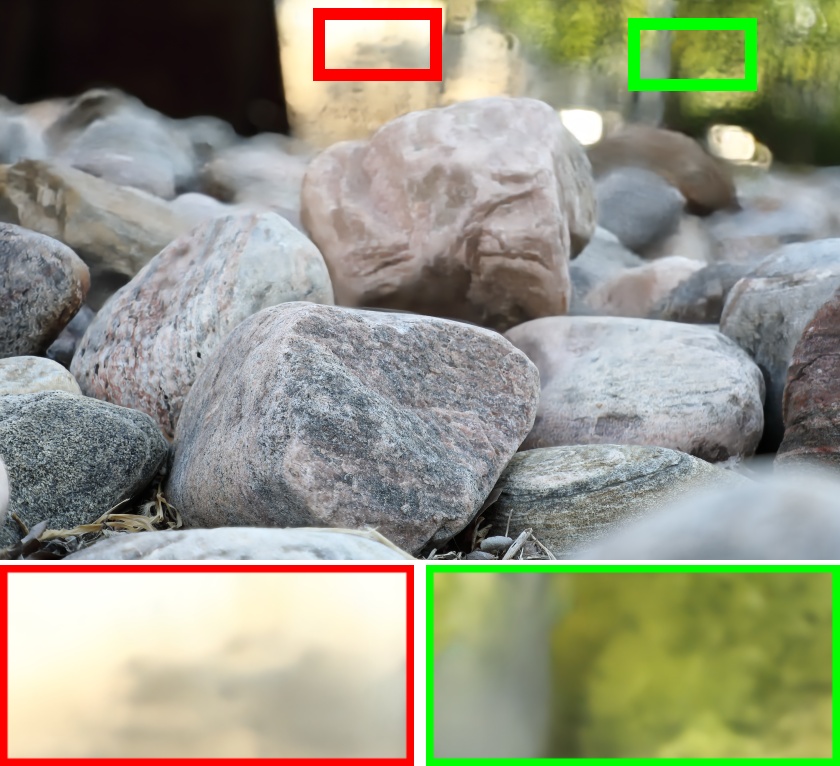}&\hspace{-4.2mm}
					\includegraphics[width=0.12\textwidth]{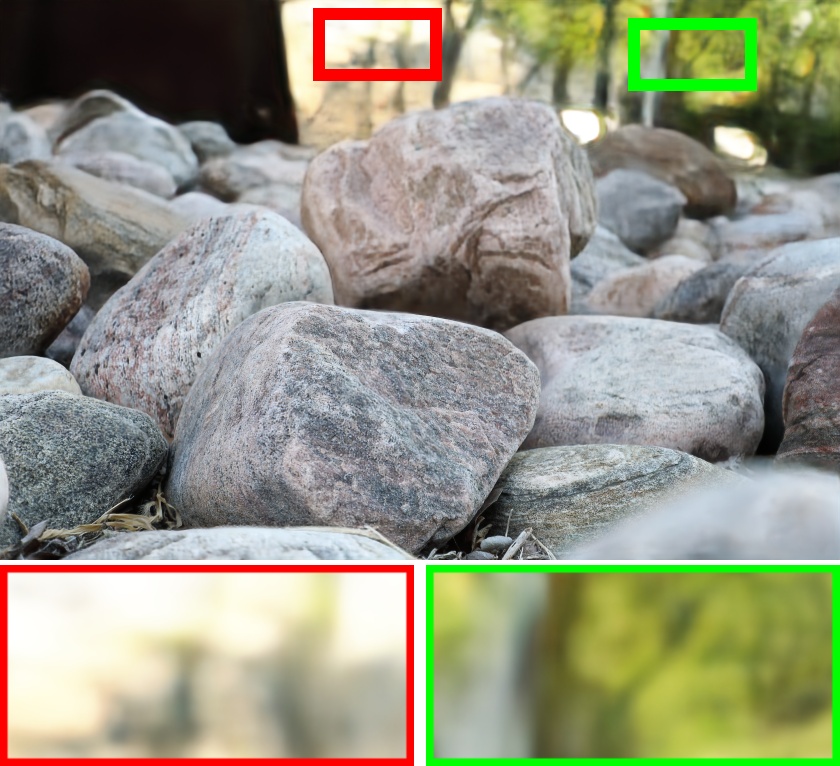}&\hspace{-4.2mm}
					\includegraphics[width=0.12\textwidth]{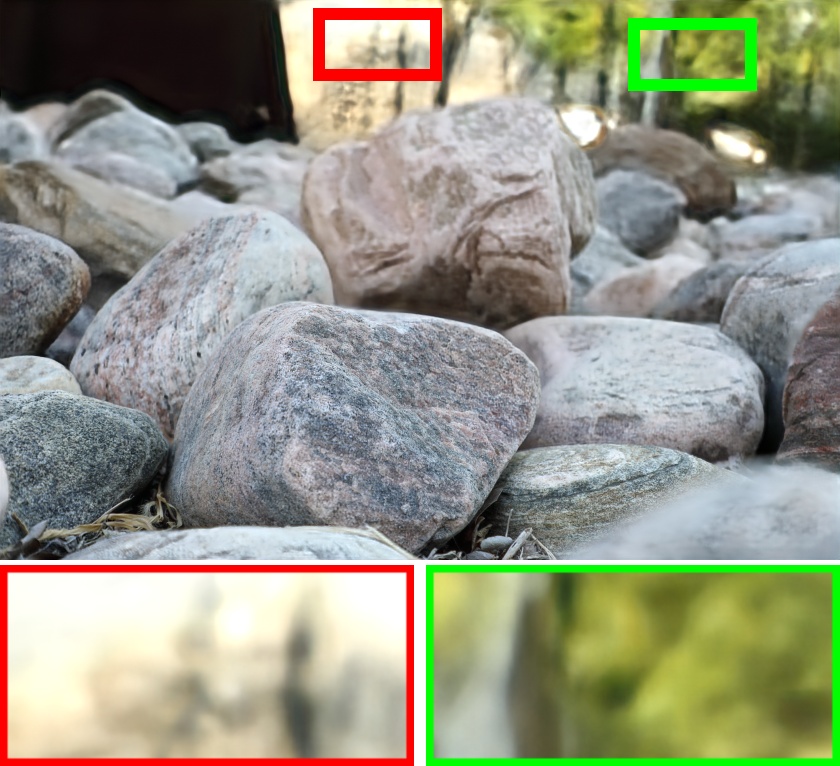}&\hspace{-4.2mm}
					\includegraphics[width=0.12\textwidth]{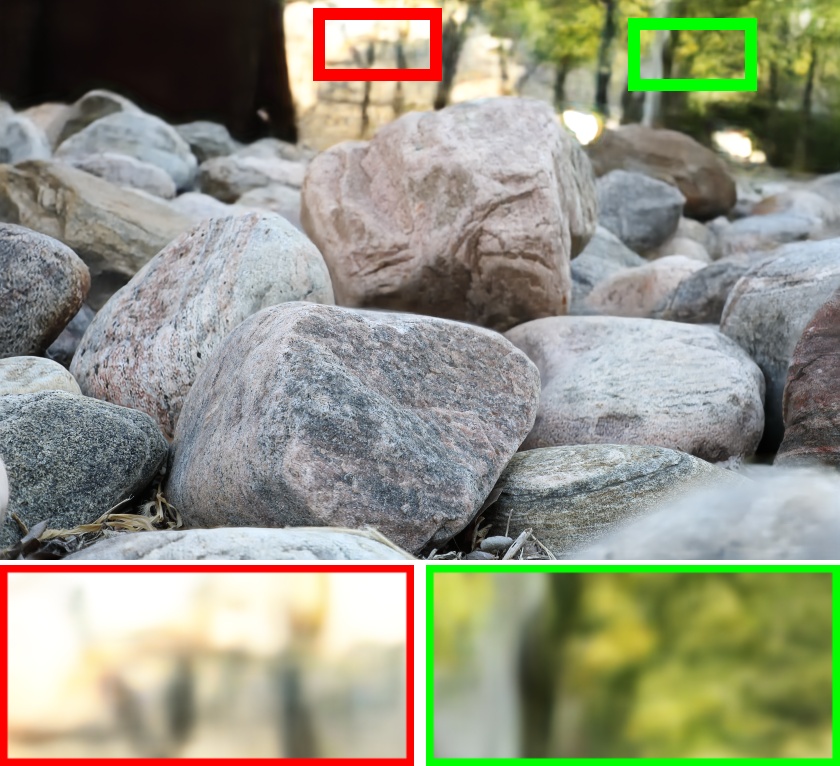}&\hspace{-4.2mm}
					\includegraphics[width=0.12\textwidth]{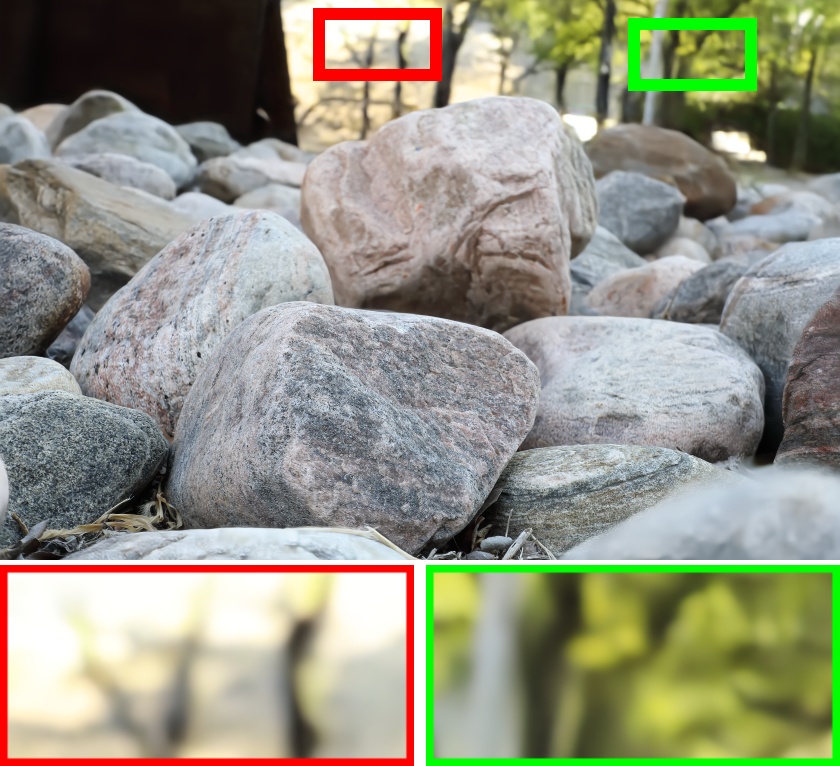}&\hspace{-4.2mm}
					\includegraphics[width=0.12\textwidth]{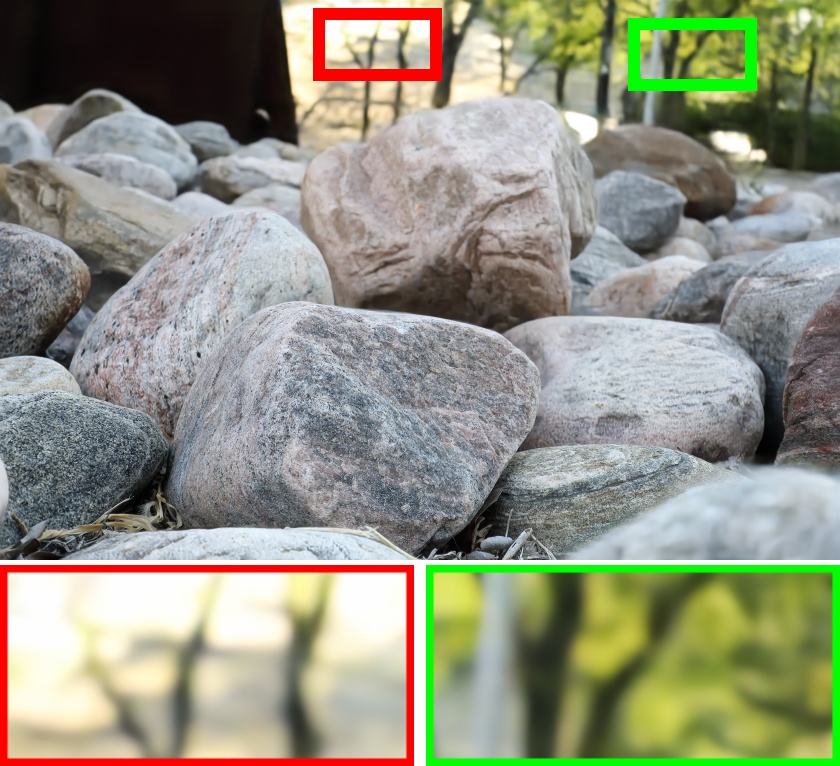}&\hspace{-4.2mm}
					\includegraphics[width=0.12\textwidth]{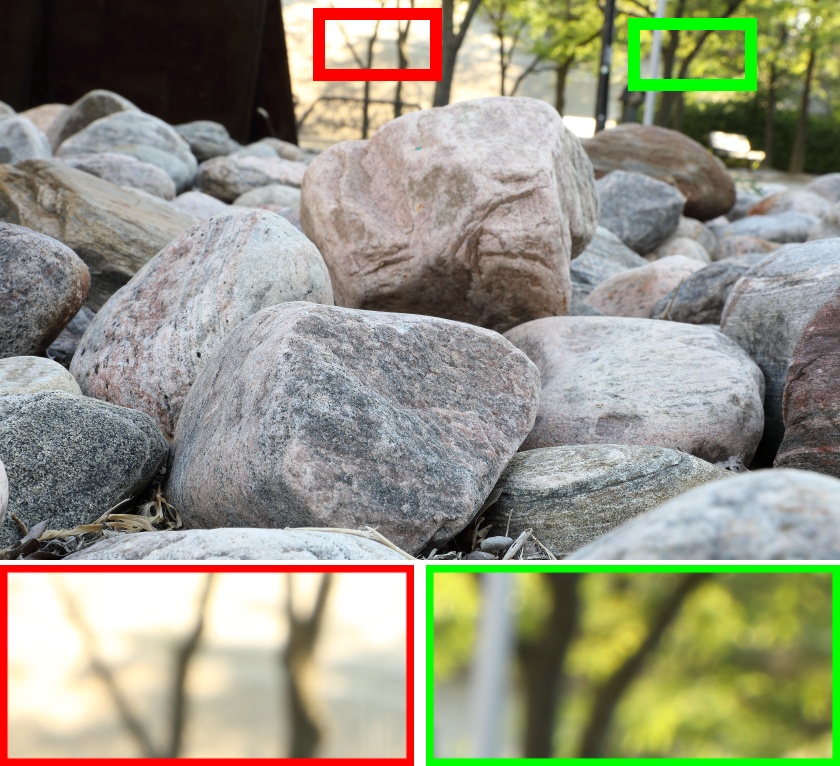}
					\\
					Blurry images&\hspace{-4.2mm}
					MPRNet~\cite{zamir2021multi}&\hspace{-4.2mm}
					DPDNet~\cite{abuolaim2020defocus}&\hspace{-4.2mm}
					RDPD+~\cite{abuolaim2020learning}&\hspace{-4.2mm}
					BaMBNet~\cite{abuolaim2020defocus}&\hspace{-4.2mm}
					Restormer~\cite{abuolaim2020learning}&\hspace{-4.2mm}
					DPANet& \hspace{-4.2mm}
					Ground-truth
					\\
				\end{tabular}
			\end{adjustbox}
			
		\end{tabular}
		\caption{Visual comparison of deblurring results of testing images from the DPDD dataset. }\label{fig:results dpdd}
	\end{figure*}

	\begin{figure*}[!htbp]
		\centering
		\begin{tabular}{cc}
			\footnotesize
			\hspace{-0.4cm}
			\begin{adjustbox}{valign=t}
				\begin{tabular}{ccccccc}
					\includegraphics[width=0.14\textwidth]{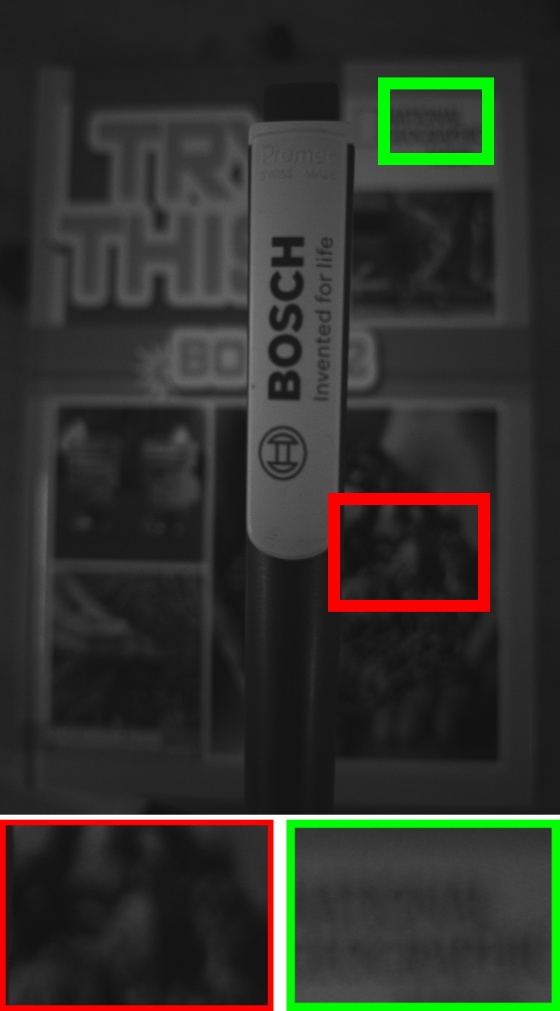}&\hspace{-4.2mm}
					\includegraphics[width=0.14\textwidth]{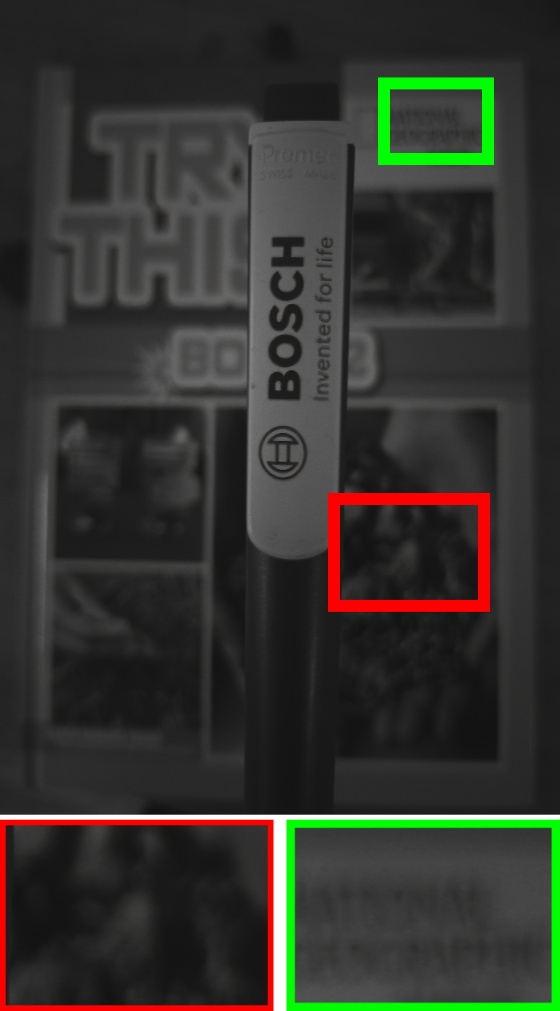}&\hspace{-4.2mm}
					\includegraphics[width=0.14\textwidth]{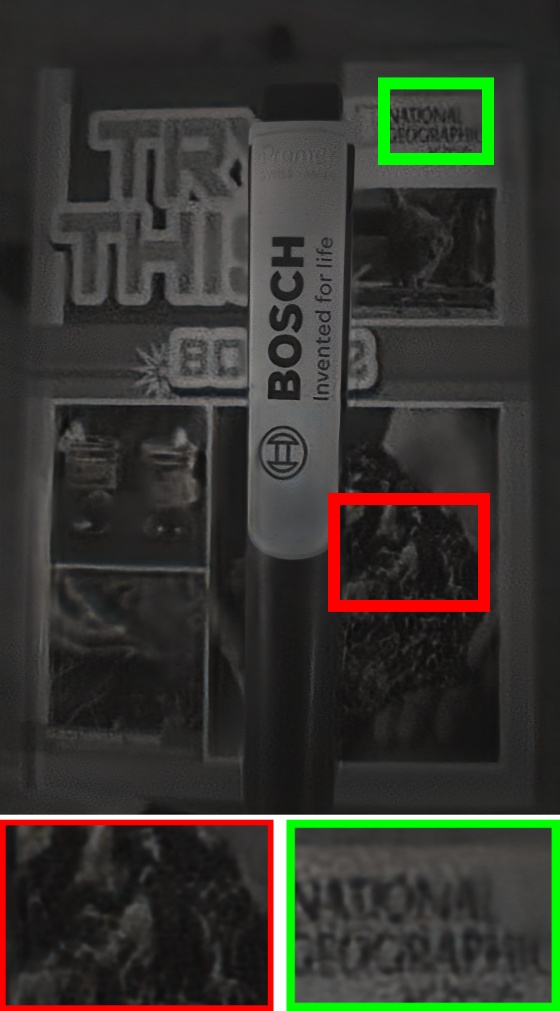}&\hspace{-4.2mm}
					\includegraphics[width=0.14\textwidth]{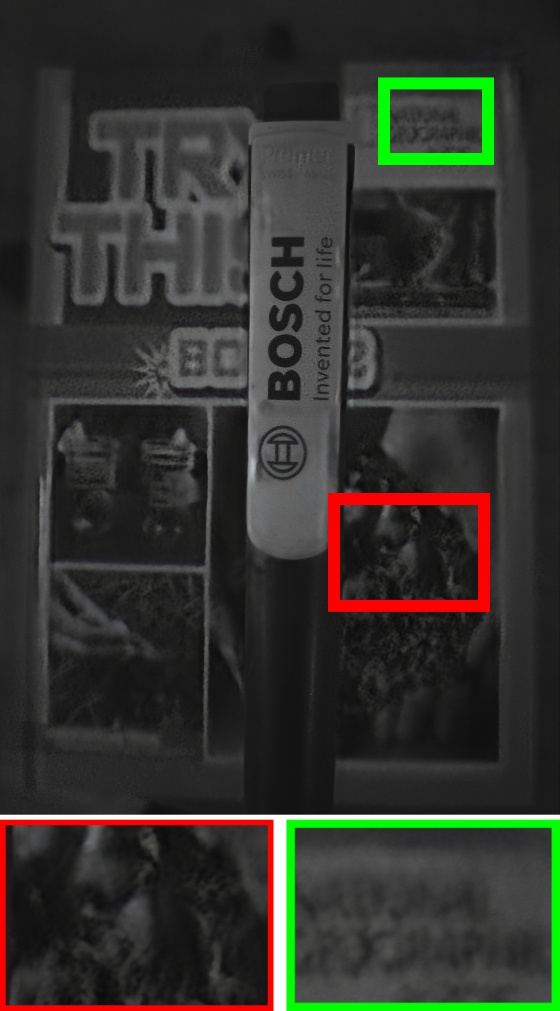}&\hspace{-4.2mm}
					\includegraphics[width=0.14\textwidth]{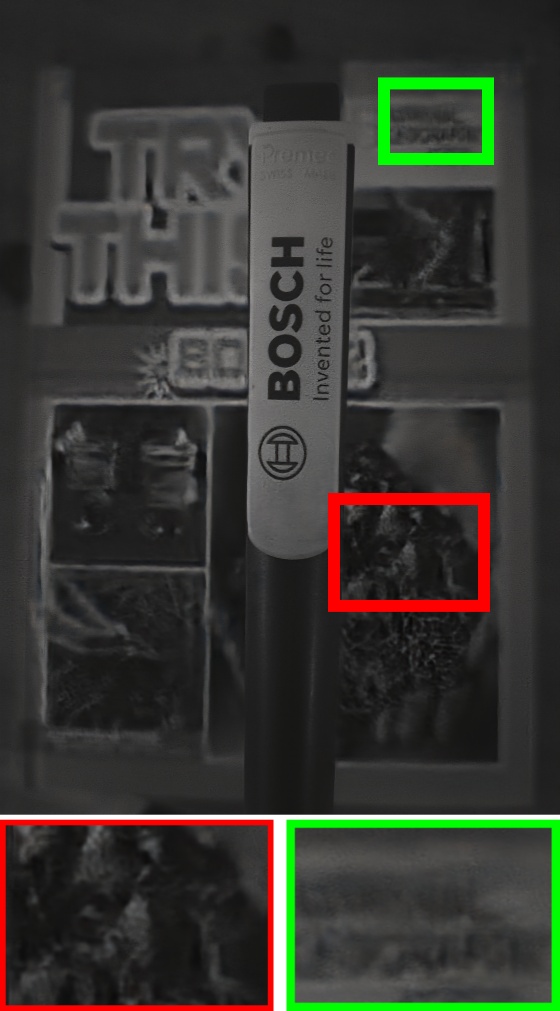}&\hspace{-4.2mm}
					\includegraphics[width=0.14\textwidth]{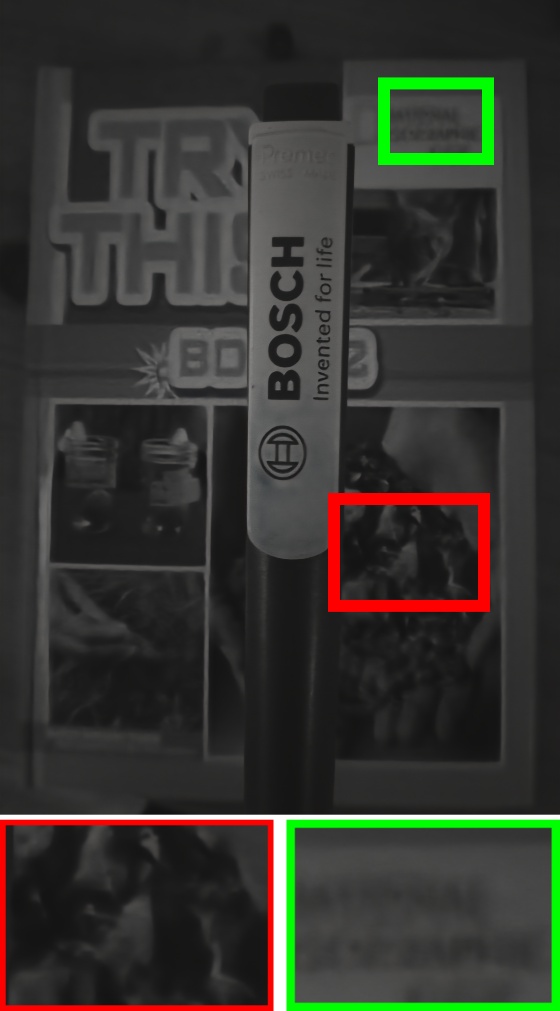}&\hspace{-4.2mm}
					\includegraphics[width=0.14\textwidth]{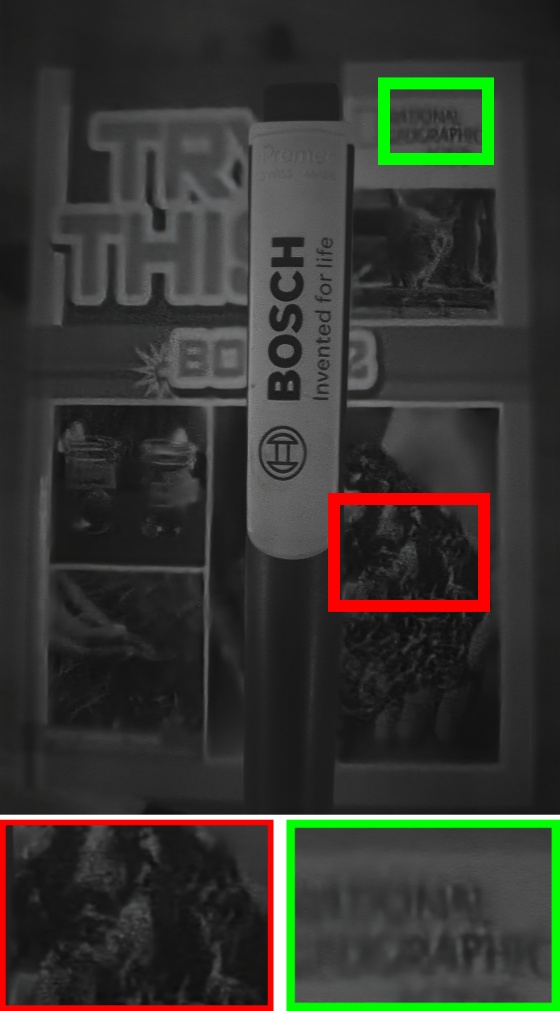}
					\\
					\includegraphics[width=0.14\textwidth]{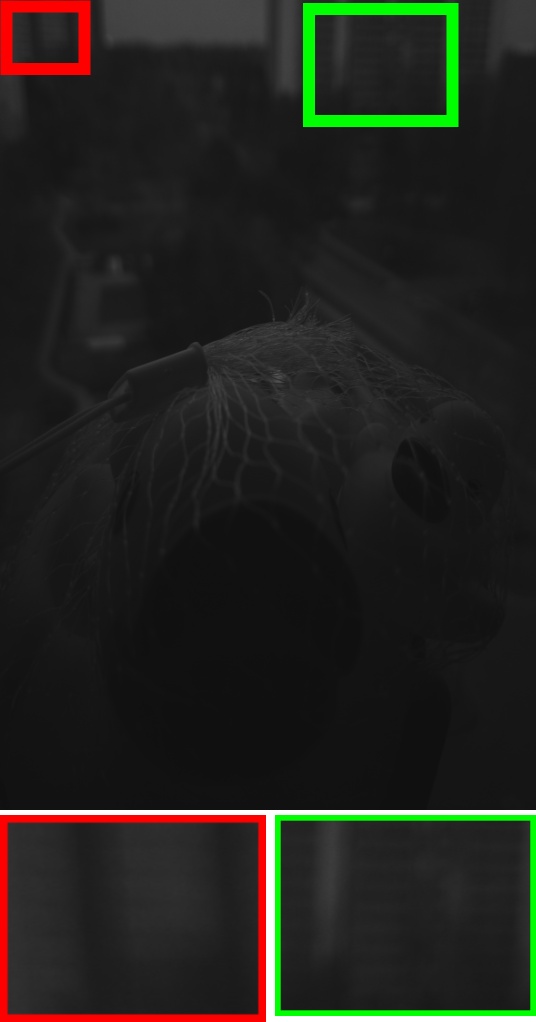}&\hspace{-4.2mm}
					\includegraphics[width=0.14\textwidth]{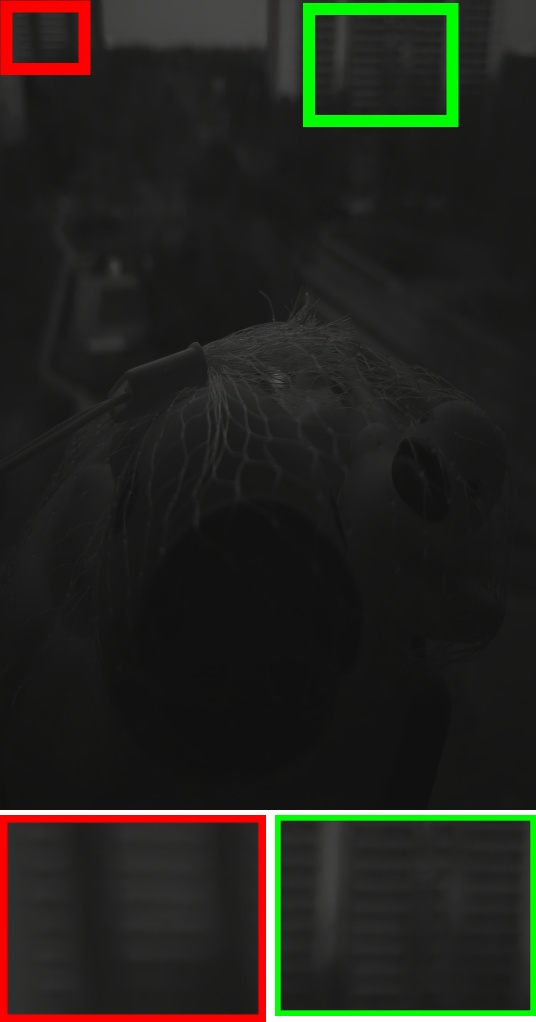}&\hspace{-4.2mm}
					\includegraphics[width=0.14\textwidth]{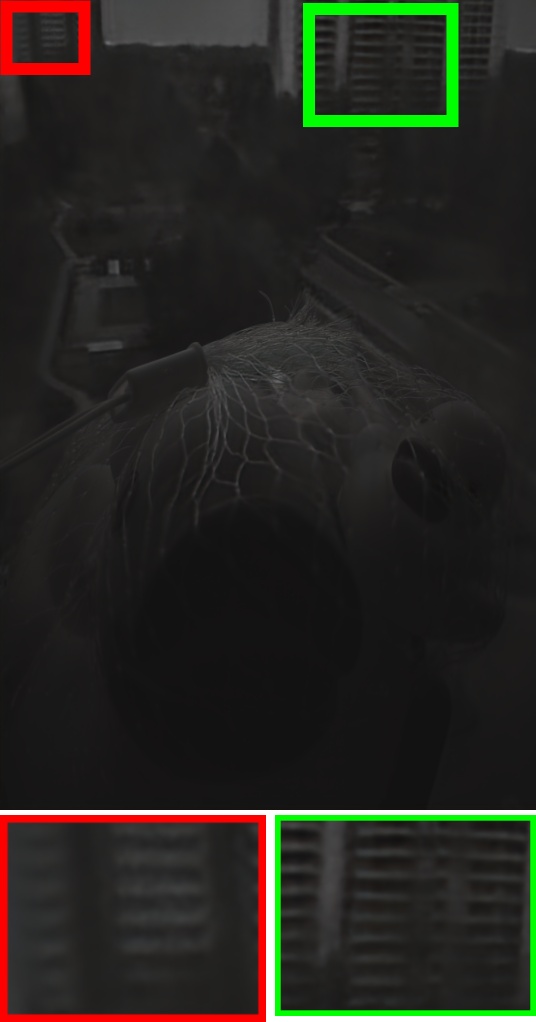}&\hspace{-4.2mm}
					\includegraphics[width=0.14\textwidth]{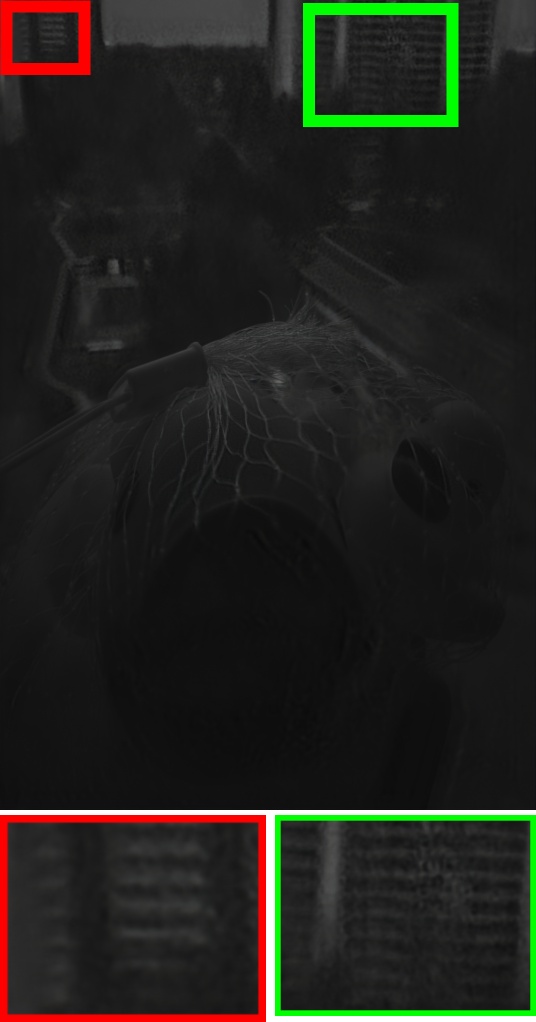}&\hspace{-4.2mm}
					\includegraphics[width=0.14\textwidth]{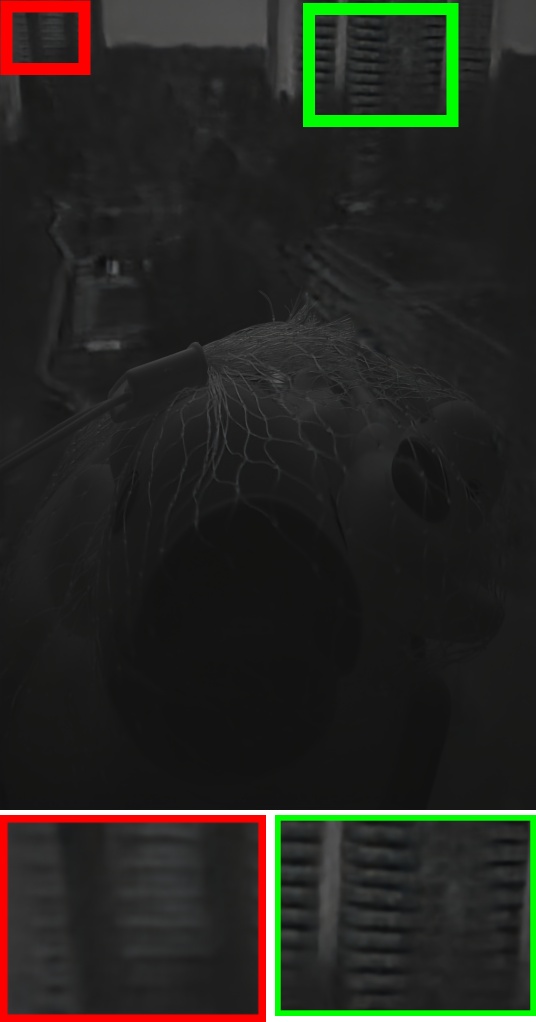}&\hspace{-4.2mm}
					\includegraphics[width=0.14\textwidth]{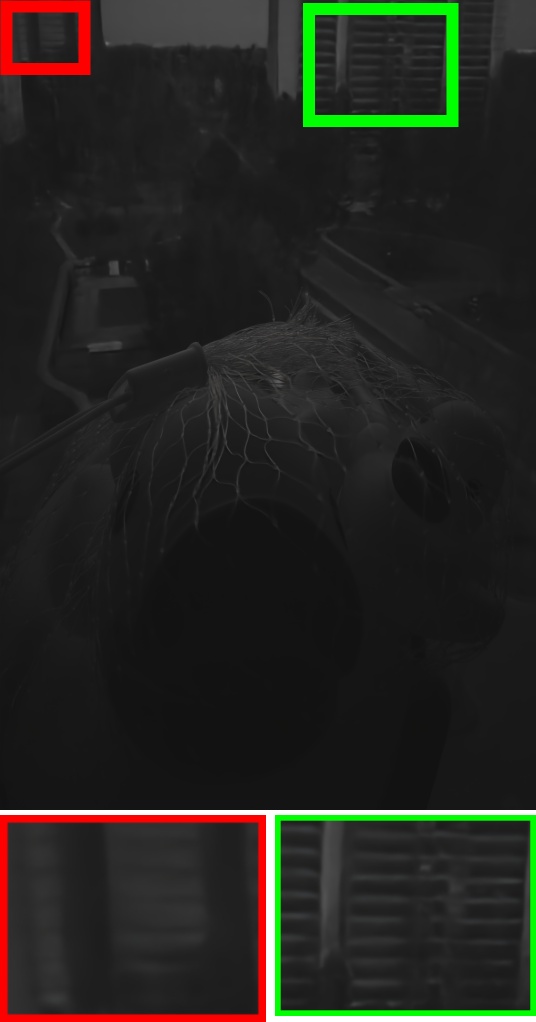}&\hspace{-4.2mm}
					\includegraphics[width=0.14\textwidth]{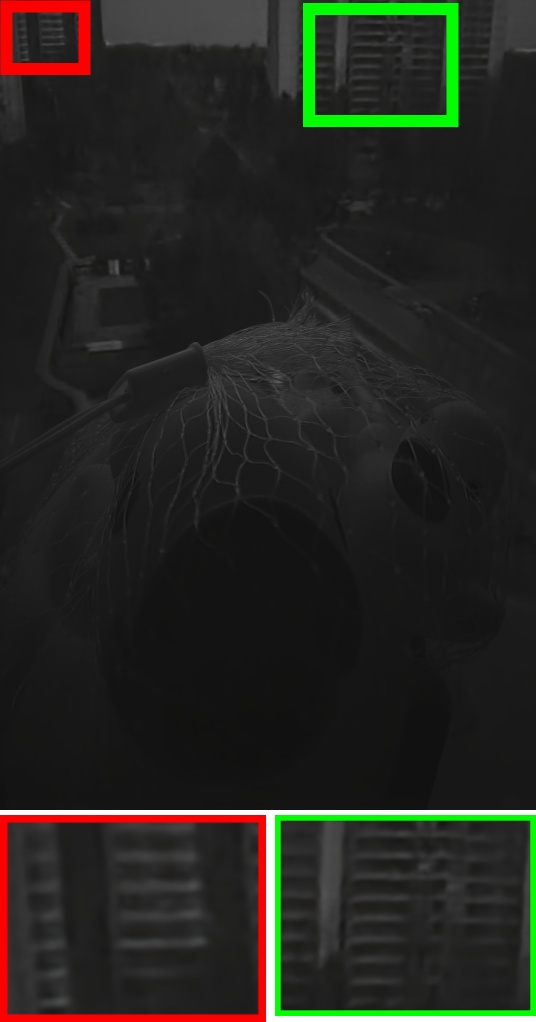}
					\\
					Blurry images&\hspace{-4.2mm}
					MPRNet~\cite{zamir2021multi}&\hspace{-4.2mm}
					DPDNet~\cite{abuolaim2020defocus}&\hspace{-4.2mm}
					RDPD+~\cite{abuolaim2020learning}&\hspace{-4.2mm}
					BaMBNet~\cite{abuolaim2020learning}&\hspace{-4.2mm}	
					Restormer~\cite{abuolaim2020learning}&\hspace{-4.2mm}
					DPANet
					\\
				\end{tabular}
			\end{adjustbox}
			
		\end{tabular}
		\caption{Visual Comparison on real-world defocus blurry images captured by the Google Pixel smartphone. 
			These methods are trained on the DPDD dataset captured by the Cannon camera. }
		\label{fig:results pixel}
	\end{figure*}

	\subsection{Comparison with State-of-the-Arts}
	\label{subsec:Method Comparison}

	\begin{figure}[htbp]
		\hspace{-0.2cm}
		\begin{overpic}[width=0.5\textwidth]{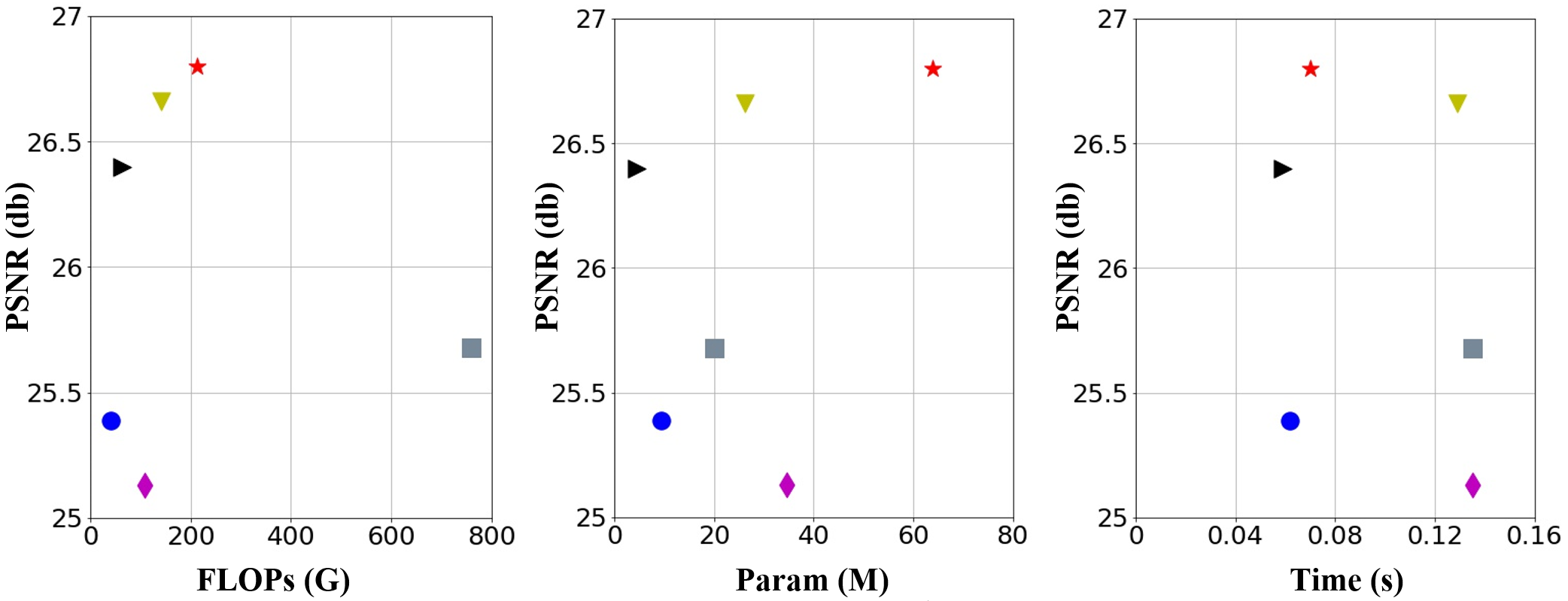}
			\put(10.3,7.2){\tiny DPDNet}
			\put(8.2,11.4){\tiny RDPD+}
			\put(23.2,17.2){\tiny MPRNet}
			\put(8.9,27){\tiny BaMBNet}
			\put(11.4,31.2){\tiny Restormer}
			\put(10,34.8){\tiny DPANet}
			
			\put(47.6,8.4){\tiny DPDNet}
			\put(40,12.3){\tiny RDPD+}
			\put(42.5,17.2){\tiny MPRNet}
			\put(41.7,27){\tiny BaMBNet}
			\put(43.7,32.7){\tiny Restormer}
			\put(57,34.8){\tiny DPANet}
			
			\put(90.5,8.4){\tiny DPDNet}
			\put(79.5,12.3){\tiny RDPD+}
			\put(90.5,17.2){\tiny MPRNet}
			\put(83,27){\tiny BaMBNet}
			\put(88.7,32.7){\tiny Restormer}
			\put(81,34.8){\tiny DPANet}
		\end{overpic}
		\caption{{{Comparison on PSNR \emph{v.s.} FLOPs / parameters / time between different methods.}}}
		\label{fig:compare-flops}
	\end{figure}
	
	\subsubsection{Quantitative Comparison}
	Our DPANet is compared with state-of-the-art DP defocus deblurring methods including DPDD \cite{abuolaim2020defocus}, RDPD+ \cite{abuolaim2020learning}, DDDNet \cite{pan2021dual}, {{BaMBNet~\cite{BaMBNet} and Restormer~\cite{Restormer}}}. 
	As for single image defocus deblurring methods, we take JNB~\cite{Shi_2015_CVPR}, EBDB~\cite{karaali2017edge} and DMENet~\cite{Lee_2019_CVPR} into comparison. 
	Since they are developed for estimating  defocus blur map, we then follow \cite{abuolaim2020defocus,abuolaim2020learning} to adopt non-blind deblurring method \cite{krishnan2009fast} for predicting final deblurring results. 
 {{
 In DPDD dataset, each scene comes along with four images in total: a middle-view blurry image, a middle-view sharp image, and dual pixel view blurry image pair. The middle-view blurry images are used as input for single image based methods in the comparison.
 }}
Moreover, we take two state-of-the-art motion deblurring methods MPRNet~\cite{zamir2021multi} and DMPHN~\cite{zhang2019deep} into comparison. 
For a fair comparison, MPRNet~\cite{zamir2021multi} and DMPHN~\cite{zhang2019deep} are re-trained on the DPDD dataset by keep consistent training protocol with our DPANet.  
	As for the quantitative evaluation, we use the test set of DPDD \cite{abuolaim2020defocus} with real-world defocus blur, which has 76 defocus blurry DP pairs of size $1680\times 1120$ along with corresponding ground-truth sharp images.

		\begin{table}[!t]
		\centering
		\small
		\caption{Comparison with DDDNet \cite{pan2021dual} on the DPDD dataset. In \cite{pan2021dual}, the quantitative results of DDDNet are computed by halving image resolution to $840\times560$. The average values of PSNR and SSIM on original resolution $1680\times 1120$ are provided by the authors of DDDNet \cite{pan2021dual}.}
		
		\label{tab:DDDNet}
		\setlength{\tabcolsep}{10pt}
		
		\begin{tabular}{c|cc|ccccc}
			\hline 
			
			\hline
			Image size&\multicolumn{2}{c|}{$1680\times 1120$}  & \multicolumn{2}{c}{$840\times 560$}\\
			\hline 
			Metric	& PSNR & SSIM & PSNR & SSIM \\
			\hline 
			DDDNet & 25.41& \textbf{0.839} & 26.76 & 0.842\\
			DPANet & \textbf{26.80}&0.835 & \textbf{27.63} & \textbf{0.868}\\
			\hline 
			
			\hline
		\end{tabular}
	\end{table}

	We adopt Peak Signal-to-Noise Ratio (PSNR), Structural Similiarity (SSIM)~\cite{wang2004image}, Mean Absolute Error (MAE), and Learned Perceptual Image Patch Similarity (LPIPS)~\cite{zhang2018unreasonable} as quantitative metrics to evaluate the deblurring performance of competing methods. 
	Table~\ref{tab:comparison} reports the quantitative results of all competing methods on the DPDD test set.
	One can easily find that JNB \cite{Shi_2015_CVPR}, EBDB \cite{karaali2017edge} and DMENet \cite{Lee_2019_CVPR} are much inferior to the other competing methods.  
	The reasons are two fold:
	First, defocus blur maps estimated by these methods are inevitable to suffer from estimation errors, which may be amplified when performing non-blind deblurring. 
	Second, based on the assumption of Gaussian blur kernel, the estimated defocus blur maps by these methods are used to reconstruct spatially variant Gaussian blur kernels. 
	However, real-world defocus blur may not satisfy the Gaussian kernel assumption. 
	Therefore, it is a more feasible solution for defocus deblurring to learn a deblurring mapping from blurry image to the latent sharp image.  
	As an evidence, DMPHN and MPRNet can achieve much better results than JNB, EBDB and DMENet, although DMPHN and MPRNet are developed for motion deblurring.  
	
	As for DP defocus deblurring methods, the deblurring results by DPDNet and RDPD+ are satisfactory, due to that they exploit the benefits of DP views. 
	However, both of them directly take DP pairs as the input of deblurring network, while neglecting the misalignment between left and right views. 
	During training, DPDNet and RDPD+ networks are forced to implicitly learn the alignment of DP views, introducing extra training difficulty. 
	We also note that the training of RDPD+ \cite{abuolaim2020learning} is not only based on the DPDD dataset but also exploits more realistic training data. 
	Even though the results of RDPD+ is only a little better than DPDNet, and is much inferior to our DPANet. 
	%
	%
 %
 {{
 Among all the compared methods indcluding BaMBNet~\cite{BaMBNet} and Restormer~\cite{Restormer} which are concurrent to our work, DPANet achieves the best performance in terms of PSNR and SSIM metrics. 
 }}

	One may notice that in the paper of DDDNet \cite{pan2021dual}, their quantitative results are higher than our DPANet. 
	However, DDDNet adopts different evaluation settings with the other competing methods, and their source code and trained models have not been fully released, making it infeasible to directly take DDDNet into comparison. 
	The quantitative metrics in \cite{pan2021dual} are computed by halving spatial resolution of testing images to be $840 \times 560$. 
	For a fair comparison, we have tested our DPANet by keeping the same setting with DDDNet.
	As shown in Table \ref{tab:DDDNet}, our DPANet achieves +0.87dB PSNR and +0.26 SSIM than DDDNet in image size $840 \times 560$. 
	We also report the quantitative results of DDDNet with resolution $1680\times 1120$\footnote{The average PSNR and SSIM values of DDDNet \cite{pan2021dual} on the DPDD dataset with original resolution $1680\times 1120$ are provided by the authors \url{https://github.com/panpanfei/Dual-Pixel-Exploration-Simultaneous-Depth-Estimation-and-Image-Restoration/issues/7}.}.
	As shown in Table \ref{tab:DDDNet}, our DPANet is better than DDDNet in terms of SSIM. 
	Considering that the source code DDDNet is not fully available, the quantitative comparison with DDDNet is reported for reference.

	\subsubsection{Qualitative Comparison}
	
	Fig. \ref{fig:results dpdd} shows the qualitative results along with the corresponding ground-truth images from the DPDD test set.
	%
	%
 {{We exclude from comparison the methods that rely on defocus map estimation, since they are much lower in PSNR and SSIM compared with the others.
 For motion deblurring methods, we only compare with MPRNet, which has the best quantitative performance of its kind. For DP methods, four SOTA methods DPDNet, RDPD+, BaMBNet and Restormer are compared. It can be seen that these methods can reduce blur in most regions. However, DPANet still yields better visual results in view of clear contours and texture recovery. As an example, the leaves and meshes in Fig. \ref{fig:results dpdd} are more clear in DPANet's output.
}}
	%

	Moreover, we evaluate these competing methods on other real-world defocus blurry images captured by Google PIXEL~\cite{abuolaim2020defocus}. 
	The PIXEL dataset consists of 13 real-world blurry images.
	Since a smartphone camera has shallow DoF, ground-truth sharp images are not provided in the PIXEL dataset.  
	These competing methods are trained on the DPDD dataset captured by a Cannon camera, and thus the evaluation on the PIXEL dataset can also provide some evidence of the generalization ability.  
	%
	%

 {{As shown in Fig. \ref{fig:results pixel}, the output images of the motion-based method MPRNet have blurry contours compared with those generated by DP-based methods. DPDNet and RDPD+ still suffer from visual artifacts, \emph{e.g.}, distortions and noises, which may originate from the misalignment between left and right views. 
BaMBNet and Restormer present better visual effects but the generated contours and textures are still not clear enough.
In comparison, the deblurring results yielded by DPANet are more visually satisfactory.}}

{{A comparison on PSNR v.s. FLOPS, parameter number and running time is presented in Fig.~\ref{fig:compare-flops}. All the statistical results are obtained by feeding the networks with 256$\times$256 image patches. Among the compared methods, DPDNet and RDPD+ are implemented in Tensorflow and all the others are implemented in Pytorch. From Fig.~\ref{fig:compare-flops}, it can be seen that DPANet has the highest PSNR compared with BaMBNet and Restormer. Although DPANet has more FLOPs and parameters, it is highly computationally efficient, such that the time consumption of DPANet is less than Restormer and marginally more than BaMBNet.}}


		\begin{figure*}[!htbp]
	\small
	
	\centering
	\begin{tabular}{cc}
		\footnotesize
		\hspace{-0.4cm}
		\begin{adjustbox}{valign=t}
			\begin{tabular}{cccccc}			
				\includegraphics[width=0.16\textwidth]{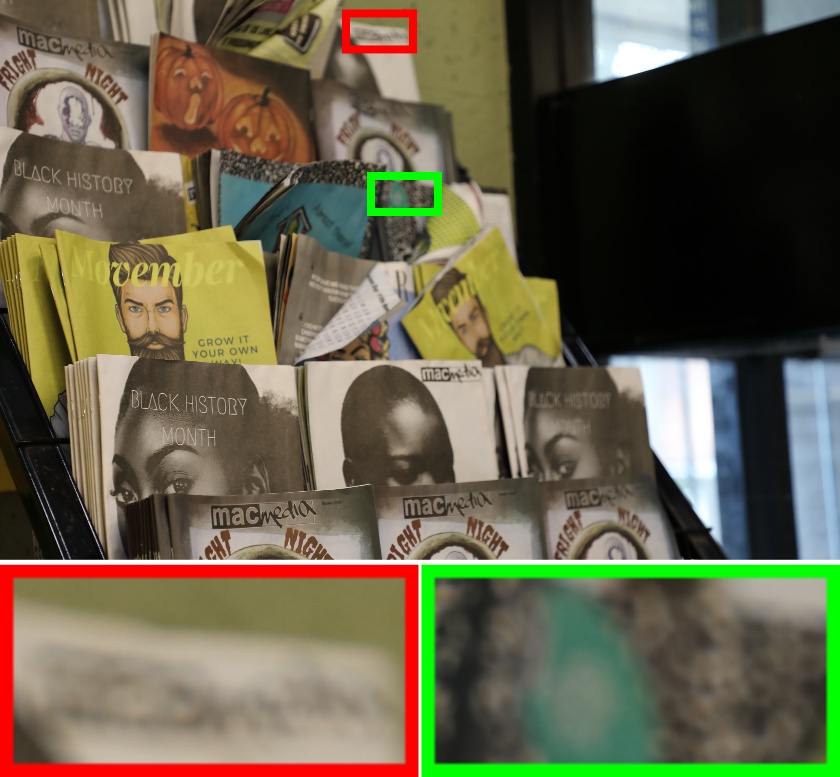}&\hspace{-4.2mm}
				\includegraphics[width=0.16\textwidth]{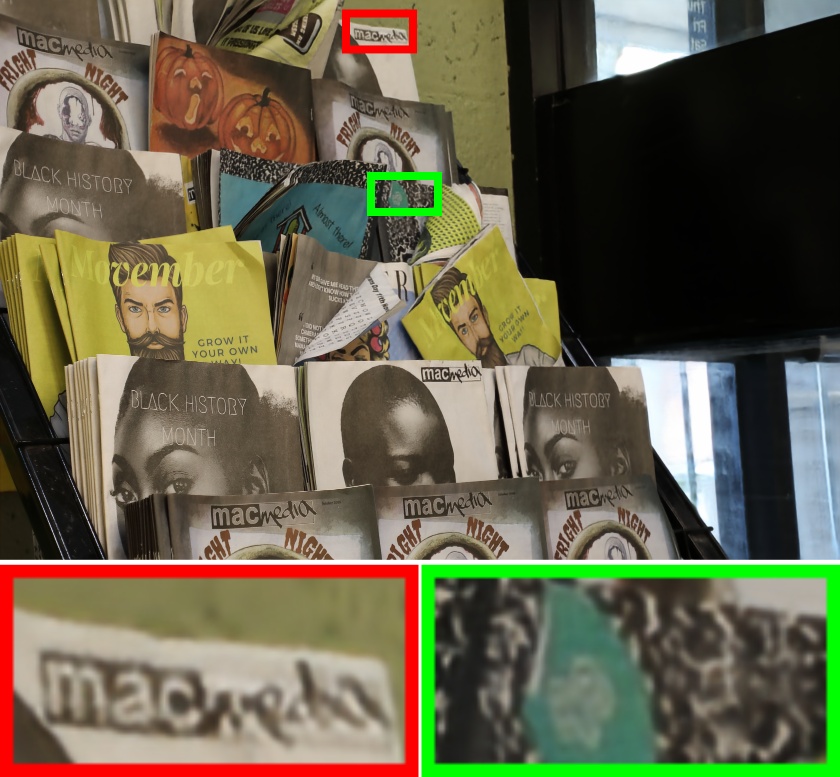}&\hspace{-4.2mm}
				\includegraphics[width=0.16\textwidth]{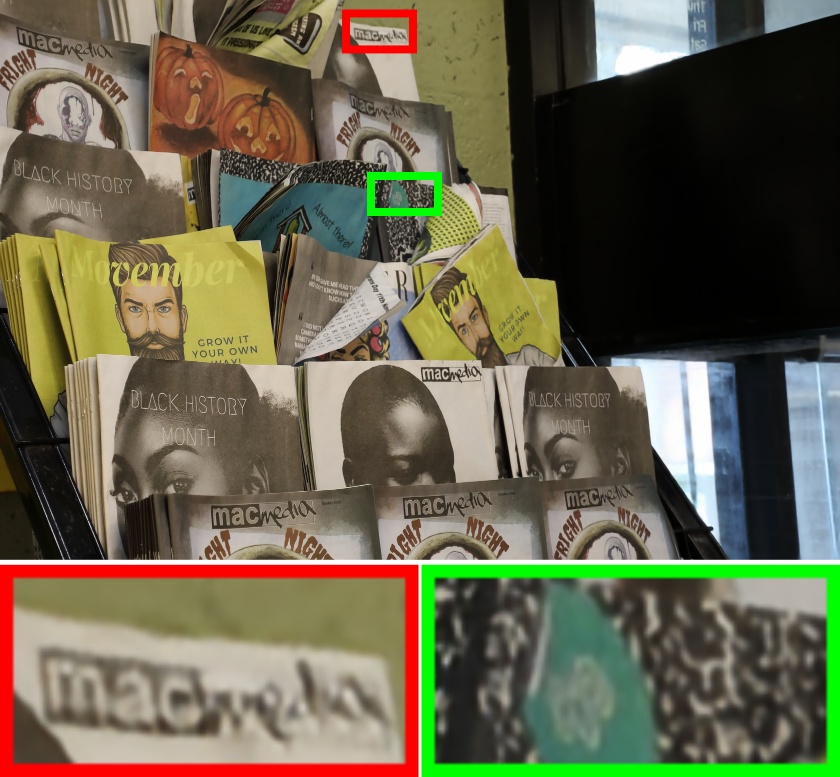}&\hspace{-4.2mm}
				\includegraphics[width=0.16\textwidth]{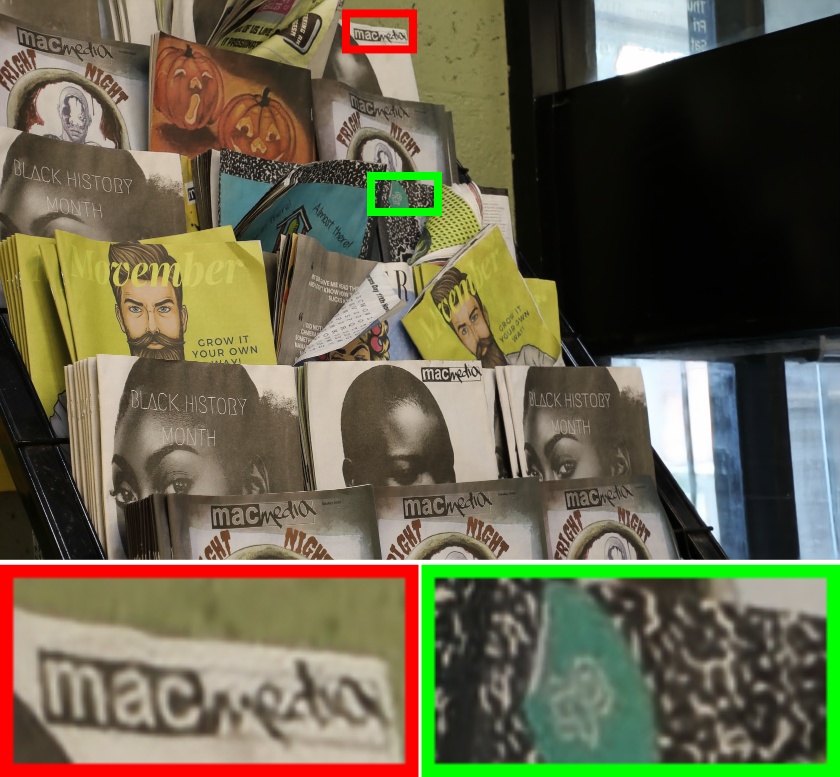}&\hspace{-4.2mm}
				\includegraphics[width=0.16\textwidth]{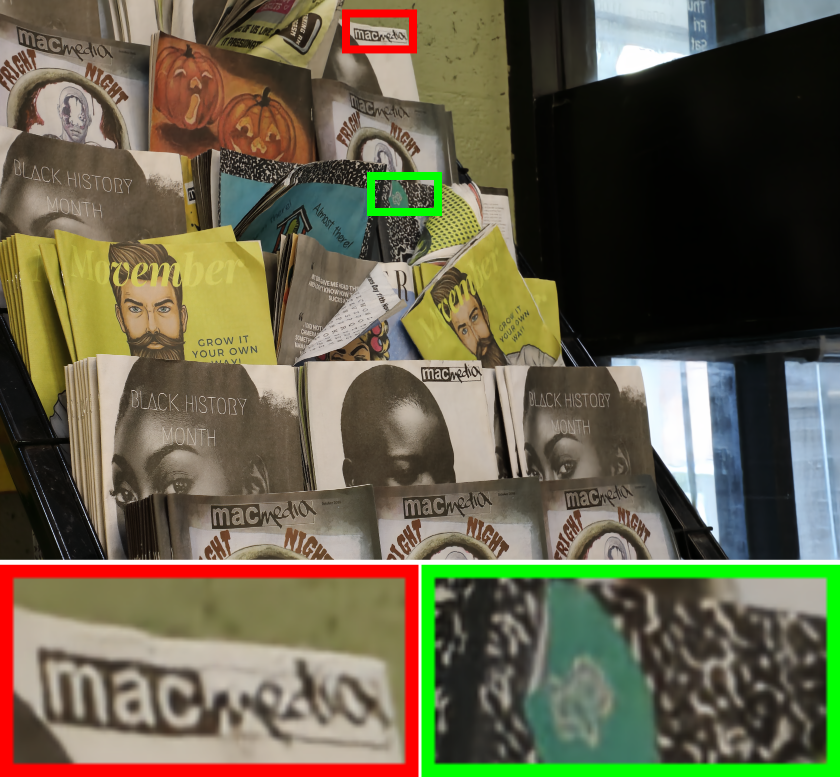}&\hspace{-4.2mm}
				\includegraphics[width=0.16\textwidth]{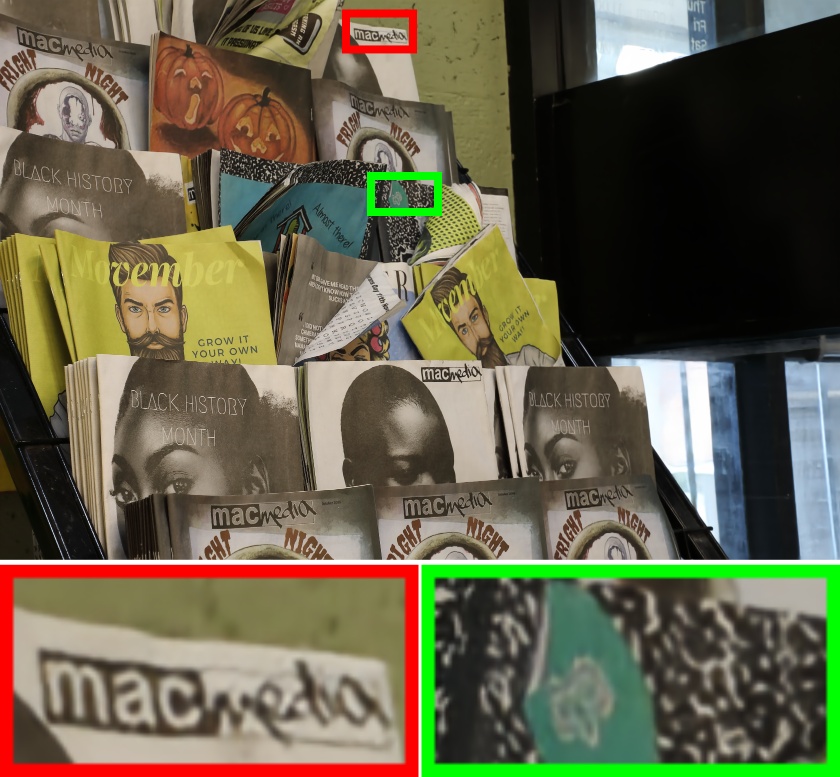}
				\\
				\includegraphics[width=0.16\textwidth]{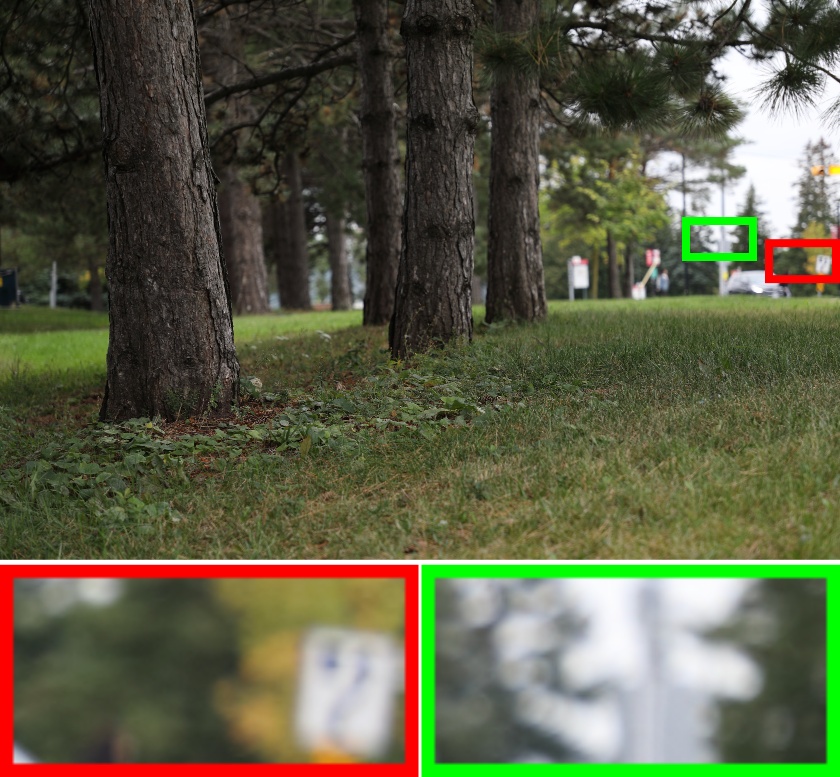}&\hspace{-4.2mm}
				\includegraphics[width=0.16\textwidth]{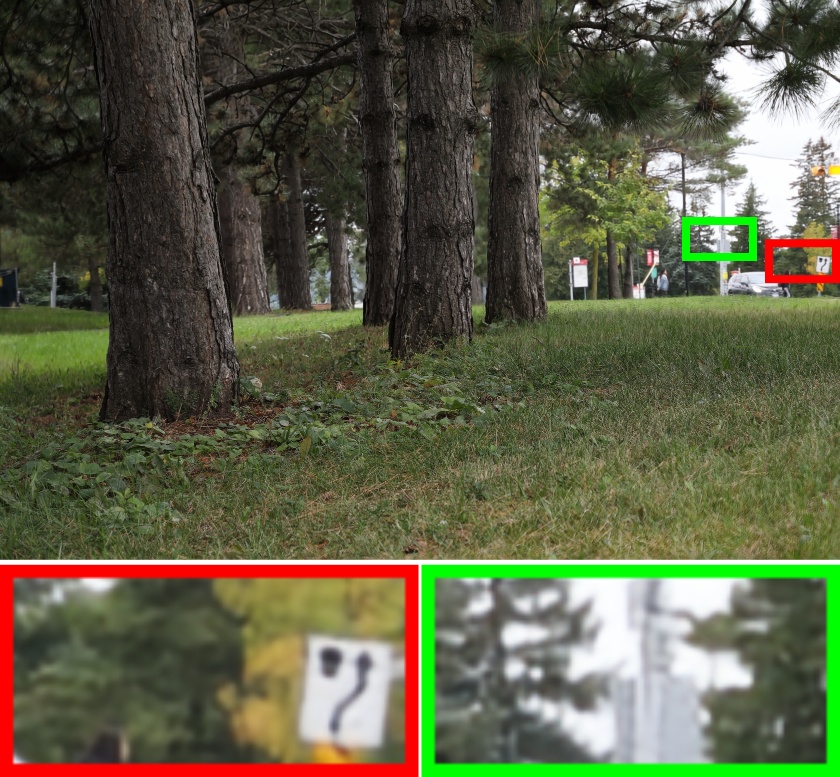}&\hspace{-4.2mm}
				\includegraphics[width=0.16\textwidth]{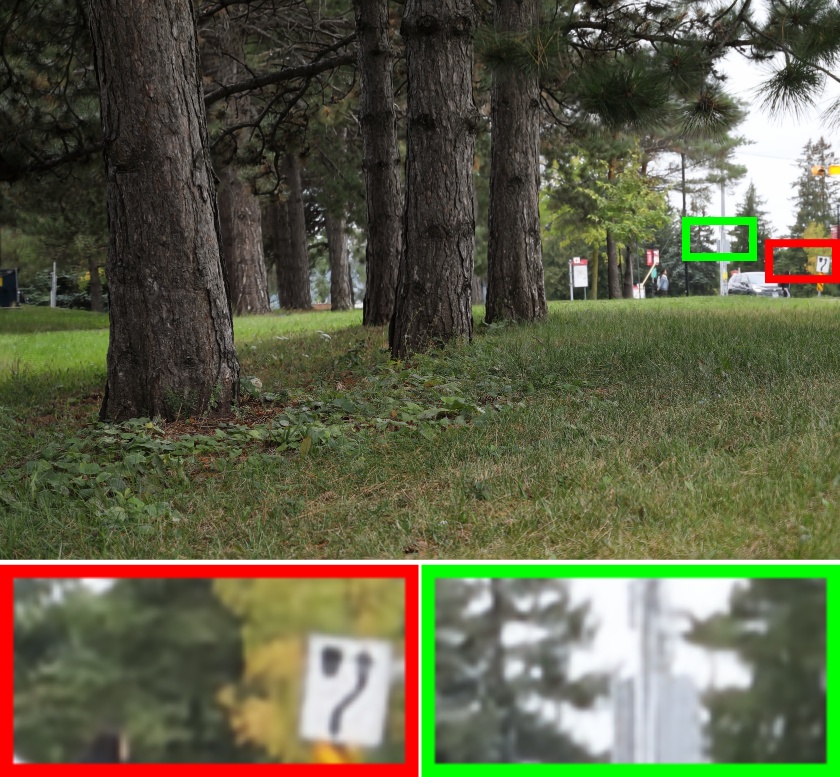}&\hspace{-4.2mm}
				\includegraphics[width=0.16\textwidth]{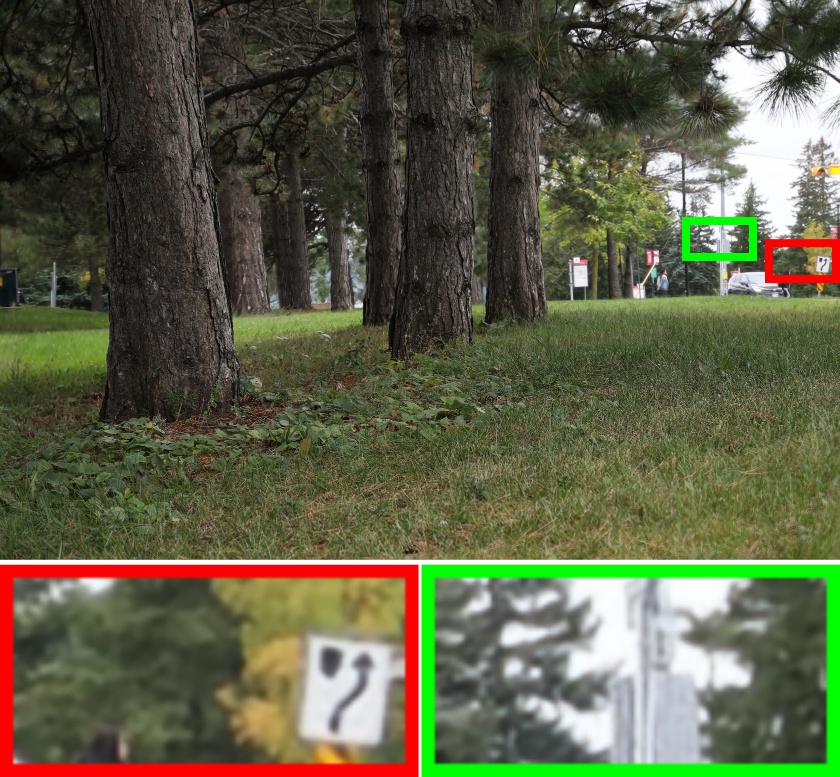}&\hspace{-4.2mm}
				\includegraphics[width=0.16\textwidth]{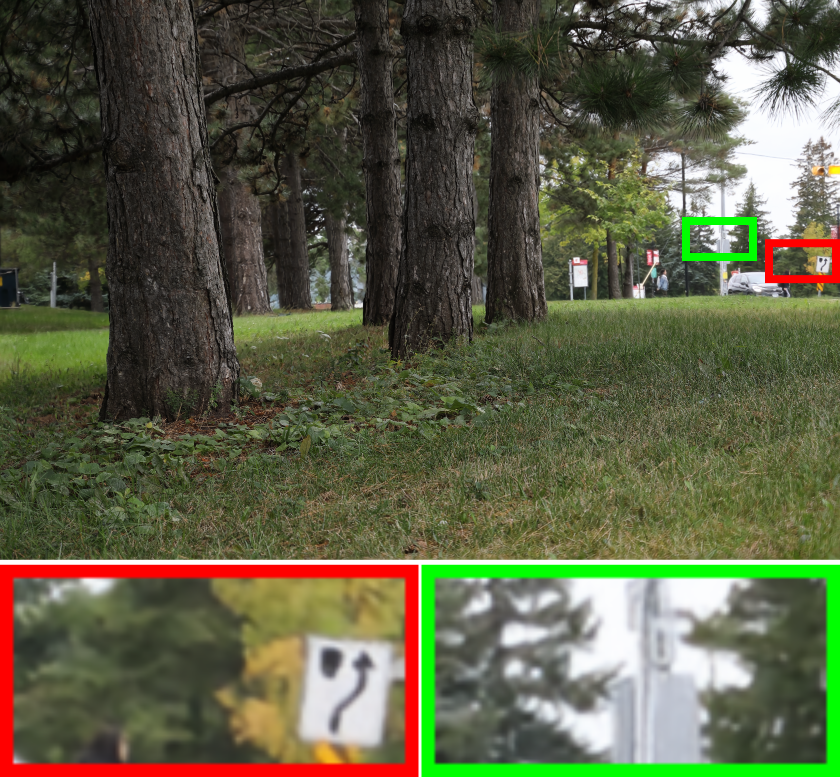}&\hspace{-4.2mm}
				\includegraphics[width=0.16\textwidth]{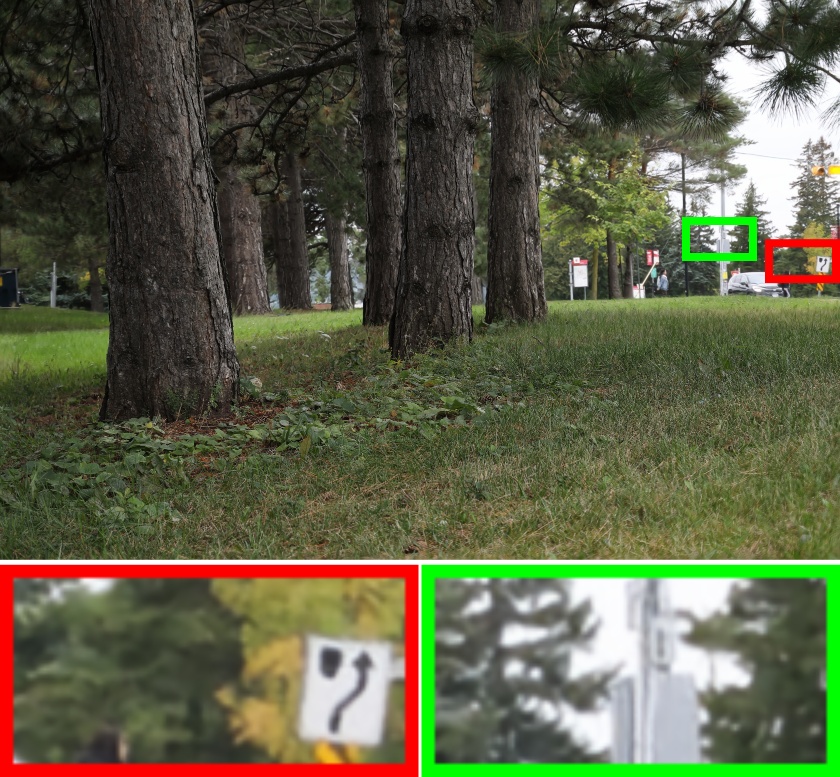}
				\\
				Blurry images&\hspace{-4.2mm}
				\#1&\hspace{-4.2mm}
				\#2& \hspace{-4.2mm}
				\#3& \hspace{-4.2mm}
				\#4& \hspace{-4.2mm}
				DPANet
				\\
			\end{tabular}
		\end{adjustbox}
		
	\end{tabular}
	\caption{Visual results of ablation studies on network backbone and training loss. The images are from the DPDD dataset.
		Two variants in terms of network backbone are \#1: two branches in the encoder are not shared, and 
		\#2: the initial feature extractor $\mathcal{F}_E^0$ is removed. 
		One variant for loss function is \#3: MSE is adopted as the training loss. {{\#4: Three convolution layers are adopted in the side branch of DCNv2.}}}
	\label{Ab_structure}
\end{figure*}

\begin{figure*}[t]
	\small
	\scriptsize
	
	\centering
	\begin{tabular}{cc}
		\footnotesize
		\hspace{-0.4cm}
		\begin{adjustbox}{valign=t}
			\begin{tabular}{ccccccc}			
				\includegraphics[width=0.138\textwidth]{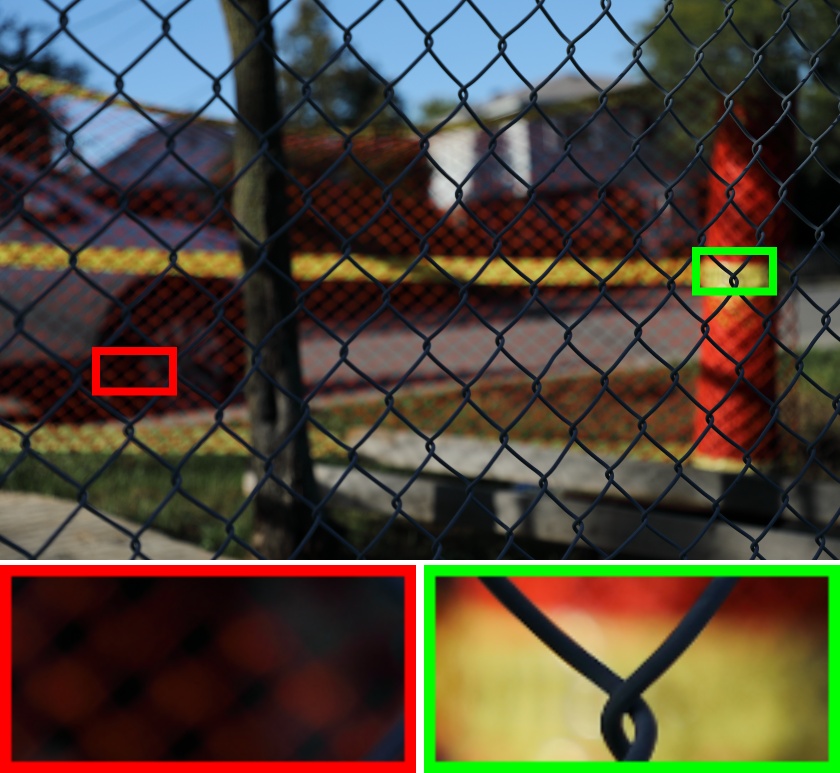}&\hspace{-4.2mm}
				\includegraphics[width=0.138\textwidth]{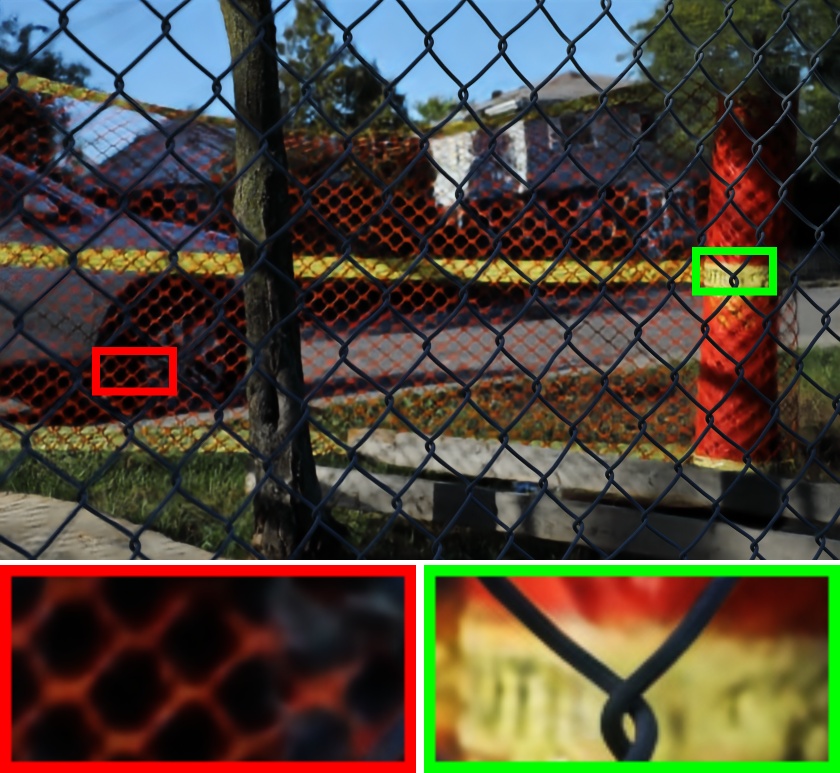}&\hspace{-4.2mm}
				\includegraphics[width=0.138\textwidth]{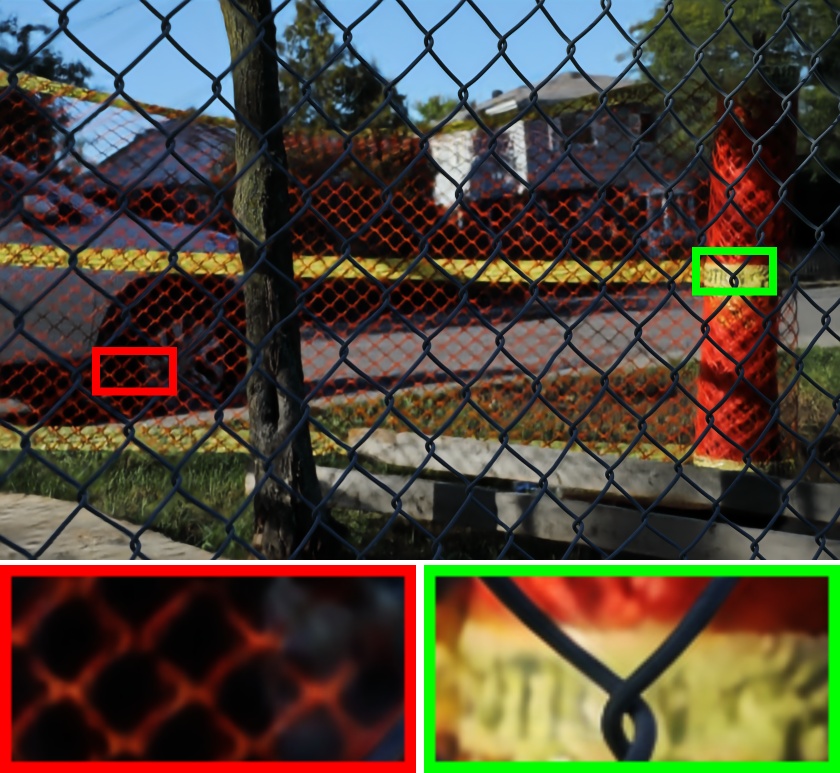}&\hspace{-4.2mm}
				\includegraphics[width=0.138\textwidth]{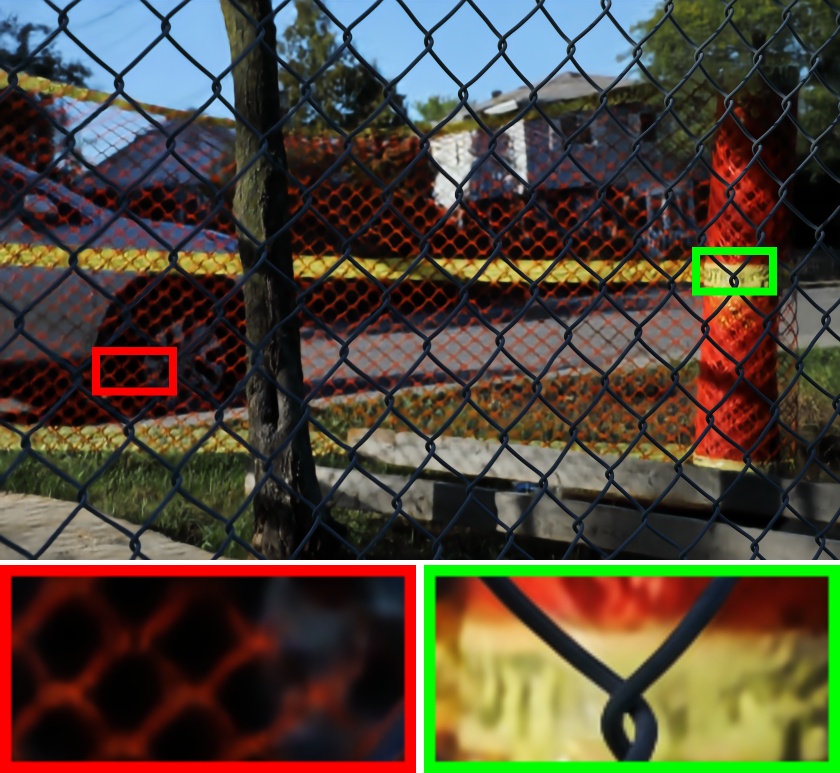}&\hspace{-4.2mm}
				\includegraphics[width=0.138\textwidth]{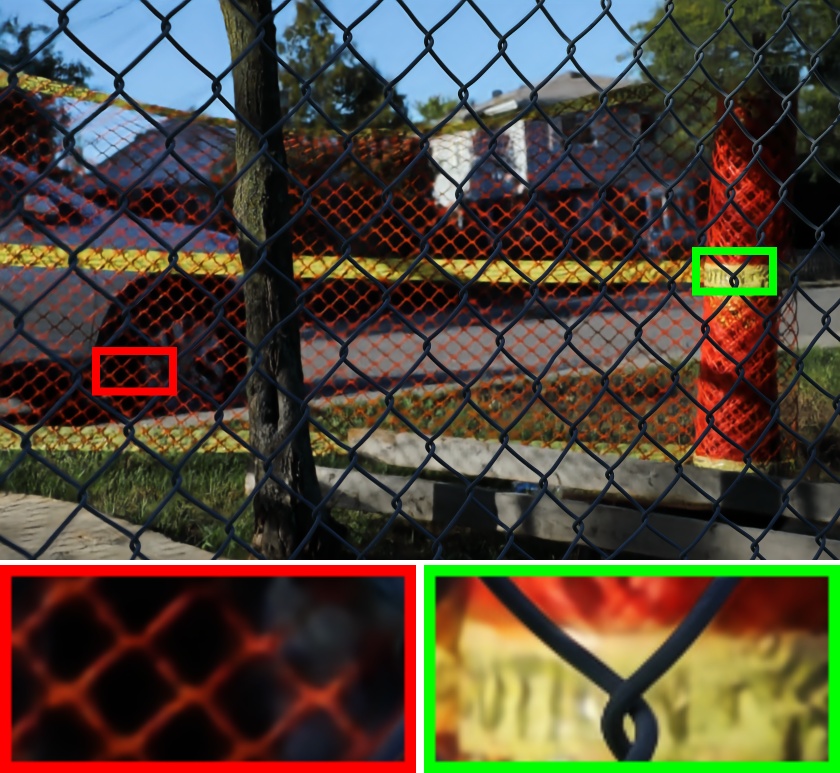}&\hspace{-4.2mm}
				\includegraphics[width=0.138\textwidth]{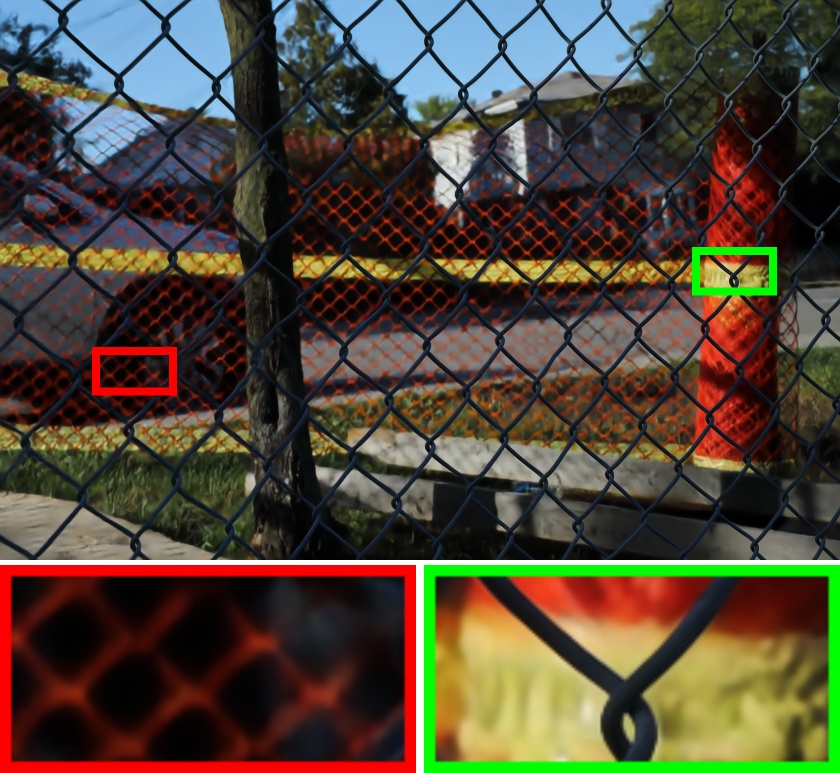}&\hspace{-4.2mm}
				\includegraphics[width=0.138\textwidth]{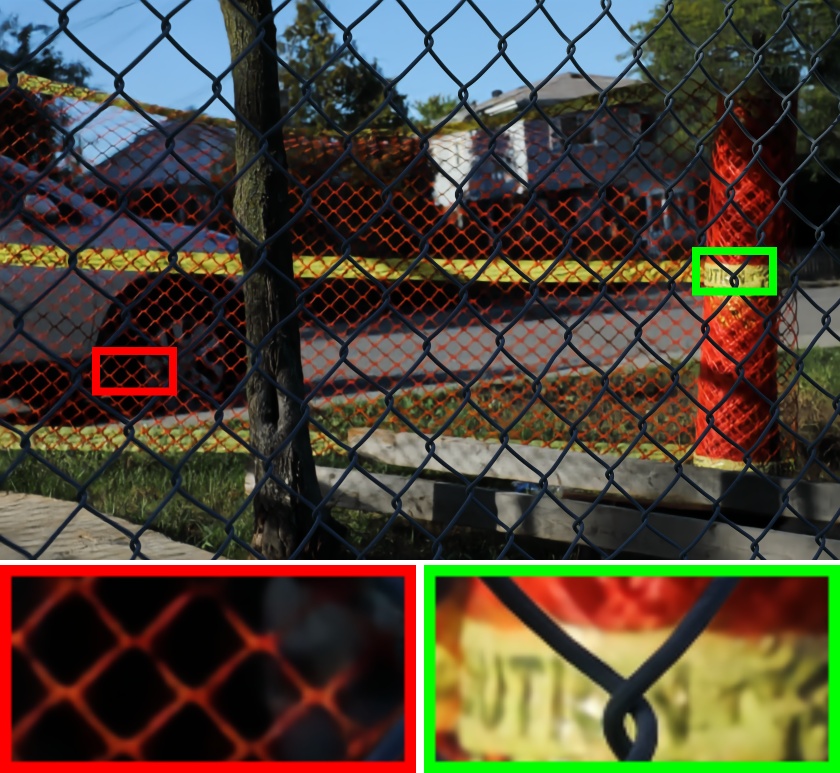}
				\\
				\includegraphics[width=0.138\textwidth]{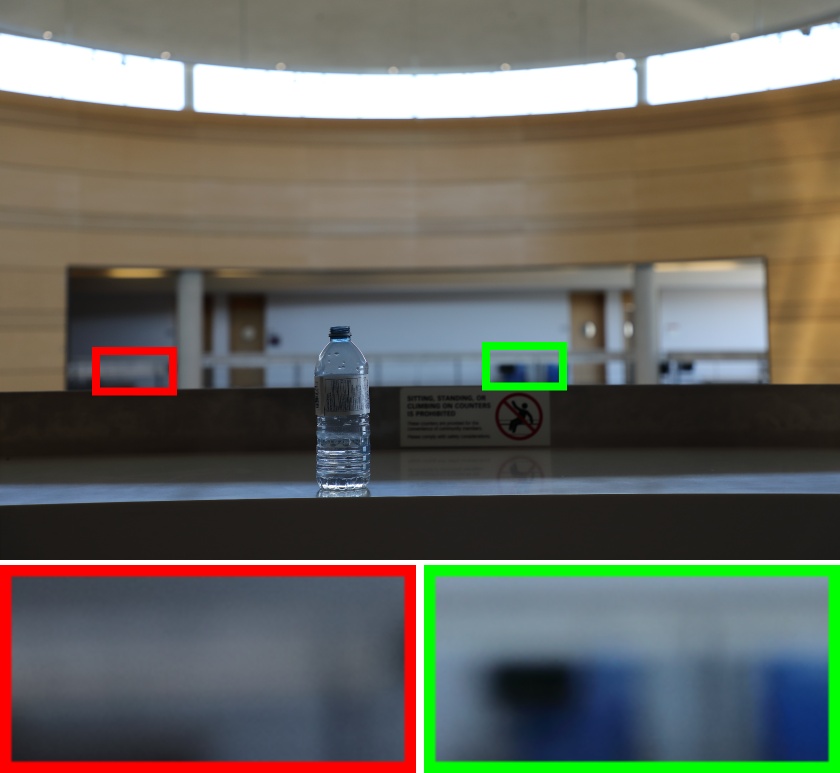}&\hspace{-4.2mm}
				\includegraphics[width=0.138\textwidth]{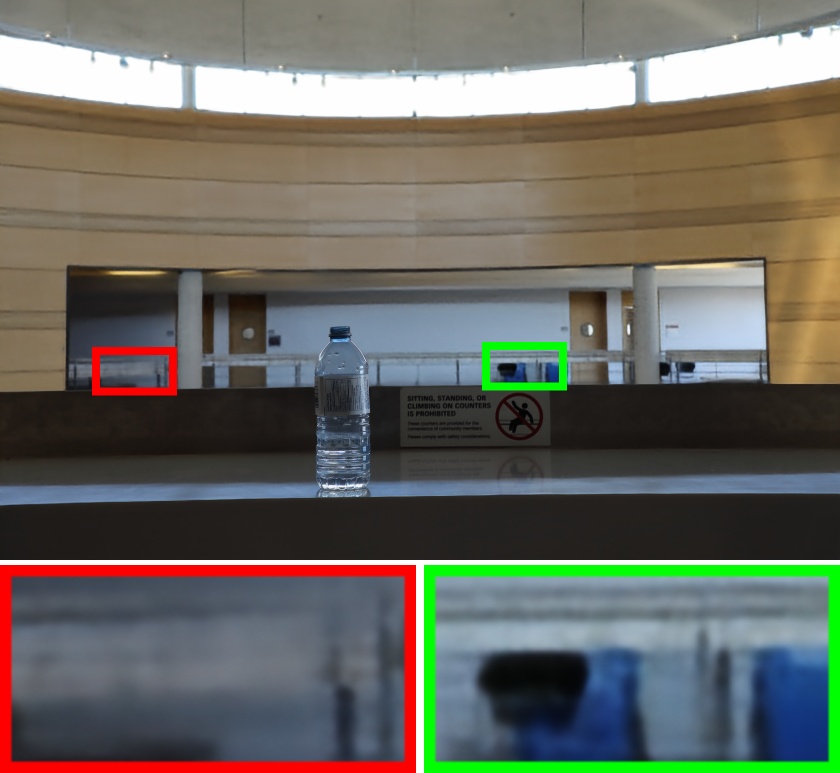}&\hspace{-4.2mm}
				\includegraphics[width=0.138\textwidth]{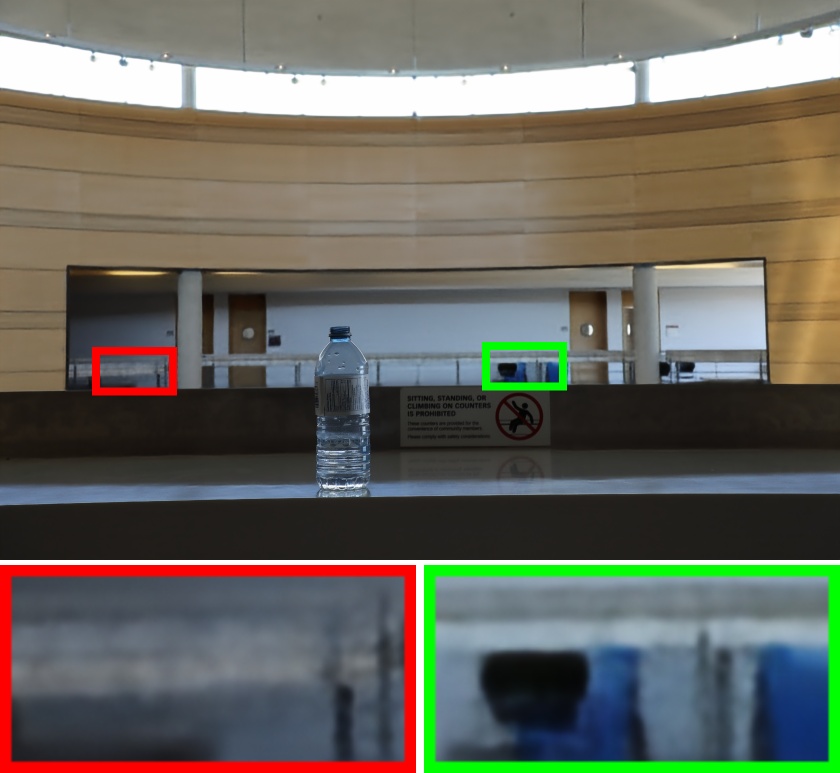}&\hspace{-4.2mm}
				\includegraphics[width=0.138\textwidth]{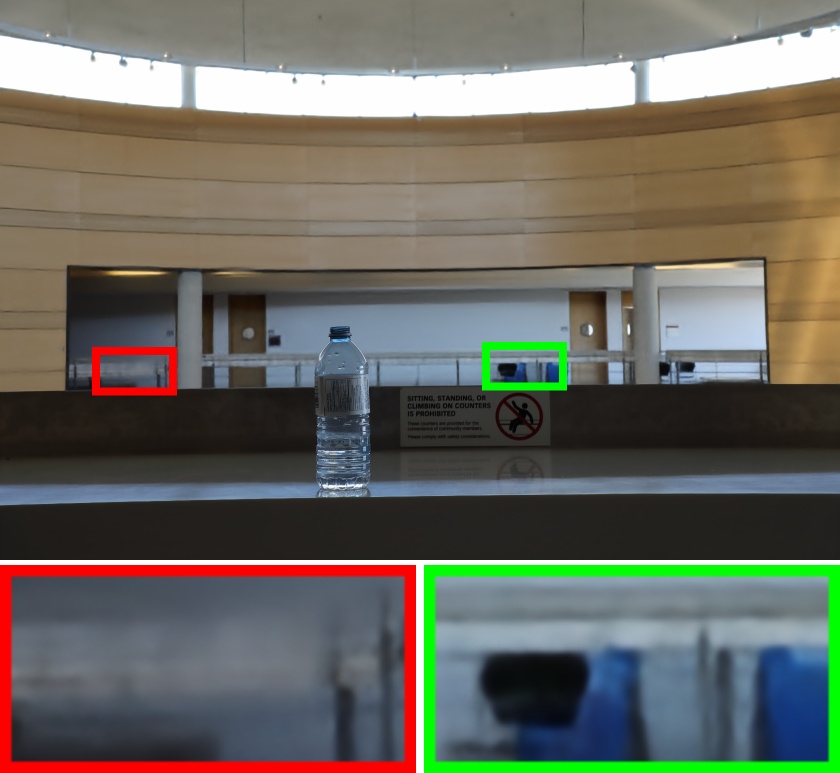}&\hspace{-4.2mm}
				\includegraphics[width=0.138\textwidth]{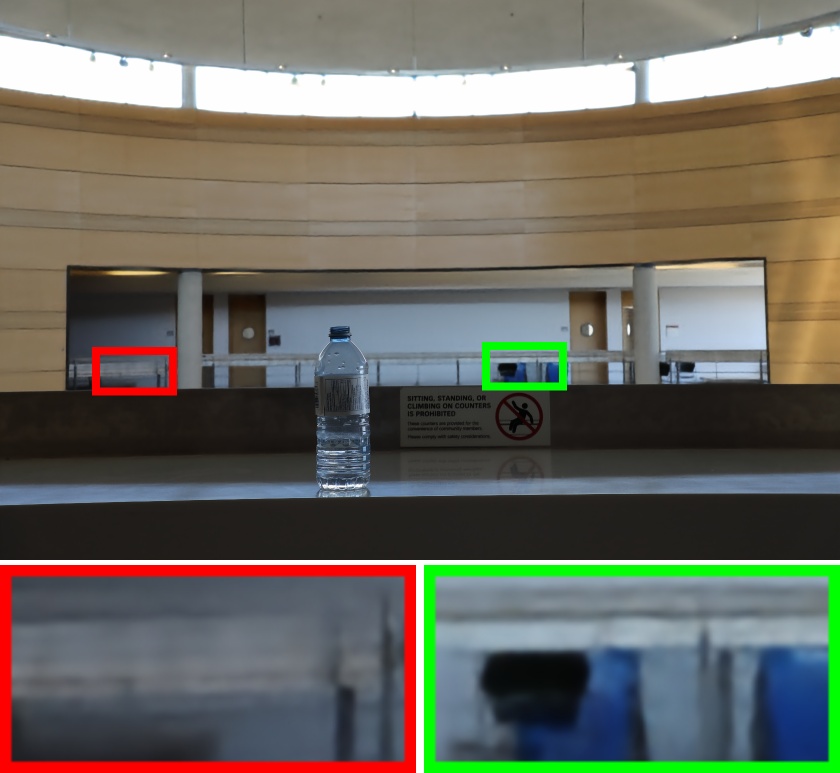}&\hspace{-4.2mm}
				\includegraphics[width=0.138\textwidth]{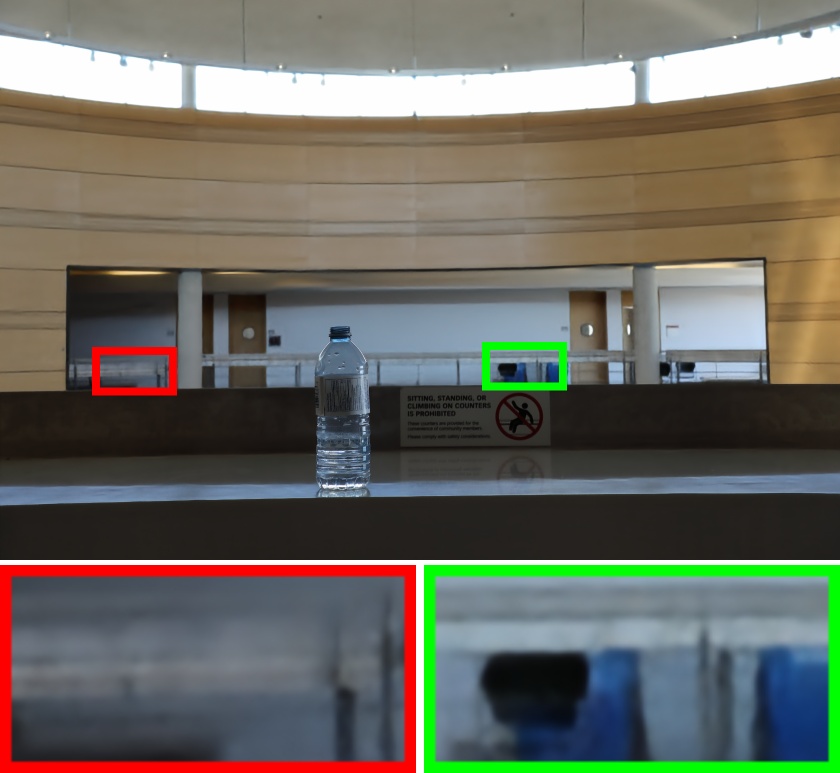}&\hspace{-4.2mm}
				\includegraphics[width=0.138\textwidth]{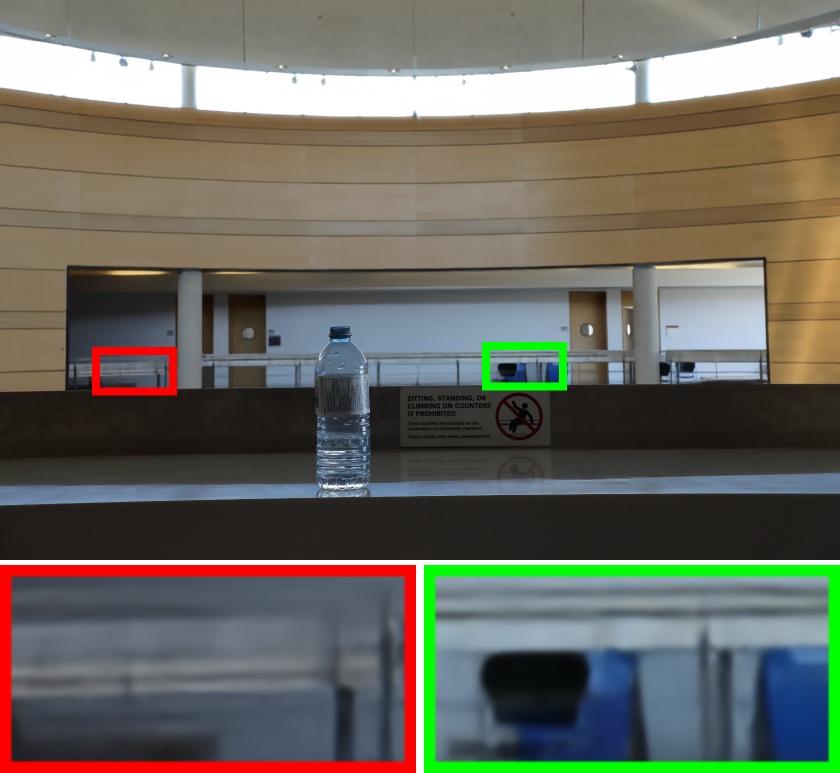}
				\\
				Blurry images&\hspace{-4.2mm}
				Baseline&\hspace{-4.2mm}
				\#5&\hspace{-4.2mm}
				\#6& \hspace{-4.2mm}
				\#7& \hspace{-4.2mm}
				\#8& \hspace{-4.2mm}
				DPANet
				\\
			\end{tabular}
		\end{adjustbox}
		
	\end{tabular}
	\caption{Visual results of ablation studies on EAM and DAM. The images are from the DPDD dataset.
		{{Baseline: DPANet without EAM and DAM}}. \#5: all EAMs removed.
		\#6: the correlation layer in EAM removed.
		\#7: the offsets and modulation scalars only estimated from the cost volume.
		\#8: all DAMs removed.}
	\label{fig:ablation1}
\end{figure*}

	
	\subsection{Ablation Study}
	\label{subsec:Ablation Studies}
	
	We analyze DPANet in terms of network backbone, training loss and alignment modules, respectively. 
	The quantitative metrics in ablation study are all computed on the DPDD test set.
	
			\begin{table}[t]
		\centering
		\small
		\caption{Ablation studies on network backbone and training loss. 
		}
		\label{tab:Ab_structure}
		\setlength{\tabcolsep}{9pt}
		\begin{tabular}{l|cccc}
			\hline
			
			\hline
			Experiment             &  PSNR & SSIM & MAE &LPIPS\\
			\hline
			\#1 {{w/o weight-share}} & 26.46 & 0.823 & 0.036 & 0.182 \\
			\#2 {{w/o $\mathcal{F}_E^0$}}& 26.52 & 0.826 & 0.036 & 0.174 \\
			\#3 {{MSE loss}}& 26.41 & 0.821 & 0.036 & 0.178 \\
			{{\#4 3 convs in DCN}}& 26.66 & 0.832 & 0.036 & 0.178 \\
			\hline
			DPANet & \textbf{26.80} & \textbf{0.835} & \textbf{0.035} & \textbf{0.162} \\			
			\hline
			
			\hline
		\end{tabular}
	\end{table}
	
	\subsubsection{Network Backbone and Training Loss}
	We provide four DPANet variants to verify the network backbone and training losses. 
	\#1: The parameters of two branches are not shared, \emph{i.e.}, two individual encoders are adopted to extract features from DP views. 
	\#2: The initial feature of encoder $\bm{E}_L^0$ and $\bm{E}_R^0$ can be directly set as input DP views $\bm{I}_L$ and $\bm{I}_R$. 
	\#3: The training loss is substituted as MSE.
    {{\#4: All the deformable convolution layers adopted in the network are designed to have three convolutional layers in the side branch for generating offsets and modulation scalars.}}

	The quantitative results are reported in Table \ref{tab:Ab_structure}, and the visual comparison is provided in Fig. \ref{Ab_structure}. 
	(i) It is interesting to see that model \#1 is inferior to DPANet, although it has more parameters by adopting two individual branches in the encoder. 
	The reason may be attributed to the DP alignment in DPANet. 
	By employing EAM in the encoder, deep features of left and right views can be well aligned, and then features from DP views are actually very similar. 
	Individual branches in the encoder actually introduce uncertainty during training, while shared branches can better handle well aligned deep features from left and right views. 
	(ii) By adopting the pyramid feature extractor as $\mathcal{F}_E^0$, multi-scale features can be extracted and are beneficial to deblurring task. 
	(iii) MSE loss function usually leads to over-smoothing texture details, while $\ell_1$-norm loss performs better in preserving sharp structure and texture details.  
    {{(iv) More convolution layers in the side branch of DCN do not contribute to network performance. Considering the computational complexity, the original one-layer design~\cite{Zhu_2019_CVPR} is kept in DPANet.}}

	\subsubsection{Alignment Modules}
	We analyze the alignment effects of EAM and DAM.  {{To start with, we build a baseline model by removing all the EAMs and DAMs. The other settings are preserved, including the encoder-decoder architecture and the two-branch design of encoder.
    From Table \ref{tab:Ab_cam_dam} and Fig.~\ref{fig:ablation1}, it can be seen that EAM and DAM improve the network performance by a large margin.
    Besides, the performance gain brought by EAMs and DAMs does not come at the cost of much computational overhead (\textit{i.e.}, 63.98M parameters and 213.77G FLOPs for DPANet v.s. 60.76M parameters and 190.82G FLOPs for the baseline model).
 }}
	
		\begin{table}[t]
		\centering
		\small
		\caption{Results of ablation studies on EAM and DAM.}
		\label{tab:Ab_cam_dam}
		\setlength{\tabcolsep}{10pt}
		\begin{tabular}{l|cccc}
			\hline 
			
			\hline
			Experiment             &  PSNR & SSIM & MAE &LPIPS\\
			\hline
            {{Baseline}}& 25.90 & 0.803 & 0.038& 0.227 \\
            \hline
			\#5 {{w/o EAM}}&  26.33 & 0.818 & 0.036 & 0.184 \\
			\#6 {{w/o $\bm{V}$}}& 26.44 & 0.821 & 0.036 & 0.176 \\
			\#7 {{offset from $\bm{V}$}}&  26.46 & 0.824 & 0.036 & 0.181 \\
			\hline
			\#8 {{w/o DAM}}& 26.45 & 0.822 & 0.036 & 0.177  \\
			\hline
			DPANet & \textbf{26.80} & \textbf{0.835} & \textbf{0.035} & \textbf{0.159} \\			
			\hline
			
			\hline
		\end{tabular}
	\end{table}
	
	\textbf{EAM:}
	We conduct three ablative experiments to verify the effectiveness of EAM.
	\#5: The alignment modules EAM are all removed.
	\#6: The correlation layer in EAM is removed, \emph{i.e.}, the offset fields for deformable convolution are estimated from the concatenation of deep features from DP views. 
	\#7: The offset fields are estimated only from the cost volume $\bm{V}$.
	The quantitative results are reported in Table \ref{tab:Ab_cam_dam}.
	From Table \ref{tab:Ab_cam_dam}, one can see that:
	(i) The deblurring performance of model \#5 is significantly inferior to DPANet, since deep features in the encoder cannot be aligned without EAM. 
	(ii) Considering the results of  model \#6 and model \#7, one can see that the correlation layer in EAM plays a crucial role in measuring the misalignment between left and right views. 
	Since model \#7 is better than model \#6, the correlation layer is more effective than direct concatenation of deep features of DP views. 
	The final DPANet leads to notable gains over these variants. 
	From the visual comparison in Fig. \ref{fig:ablation1}, ``\emph{Meshes}" are still blurry in the results of ablation variants, while the result of final DPANet is very close to the ground-truth sharp image.

	\textbf{DAM:}
	In order to verify the contribution of DAM, a variant of DPANet is designed by removing all the four DAMs. 
	This experimental setup is denoted as \#8 in Table~\ref{tab:Ab_cam_dam}, from which it can be seen that the performance is worse than the final DPANet. 
	Fig.~\ref{fig:ablation1} gives a visual comparison with or without DAM. 
	One can see that the variant without DAM fails to fully restore image details, while the final DPANet with DAM leads to more sharp and richer textures.
	The reason for the performance gain can be attributed to the alignment effects of DAM, by which skip-connected features from the encoder can be well aligned to features in the decoder, especially for the first several encoder blocks.

	\section{Conclusion}
	\label{sec:conclusion}
	In this paper, we proposed a dual-pixel alignment network for defocus deblurring, in which DP alignment and defocus deblurring can be simultaneously learned. 
	DPANet is generally an encoder-decoder, where two shared branches in the encoder are used to extract and align features from DP views and one decoder is employed to predict deblurring results.
	To relieve the adverse effects of misalignment between DP views, EAM and DAM are proposed to align deep features in encoder and decoder, respectively. 
	In DPANet, deep features from DP views can be progressively aligned to the latent sharp image, making the defocus deblurring network easier to be trained. 
	Experimental results on the DPDD dataset validated that our DPANet is superior to existing state-of-the-art deblurring methods quantitatively and qualitatively. 
	When applying to PIXEL dataset, DPANet is also verified to have better generalization ability.


	\ifCLASSOPTIONcaptionsoff
	\newpage
	\fi
	
	
\end{document}